\documentclass[11pt,a4paper]{article}
\usepackage[T1]{fontenc}
\usepackage[utf8]{inputenc}
\usepackage[margin=23mm,top=19mm,bottom=21mm,headheight=14pt,headsep=7mm]{geometry}
\usepackage{newtxtext,newtxmath}
\usepackage{microtype,inconsolata}
\usepackage{graphicx,adjustbox,booktabs,multirow,amsmath}
\usepackage[table]{xcolor}
\usepackage{natbib}
\usepackage{titlesec,fancyhdr,enumitem,caption,float}
\usepackage[section]{placeins}
\usepackage{needspace}
\usepackage[most]{tcolorbox}
\usepackage{hyperref}
\usepackage{bookmark}
\definecolor{Navy}{HTML}{154677}
\definecolor{Sand}{HTML}{E5C687}
\definecolor{Mist}{HTML}{F1F5F8}
\definecolor{Rule}{HTML}{C5D1DD}
\hypersetup{colorlinks=true,linkcolor=Navy,citecolor=Navy,urlcolor=Navy,
 pdftitle={MInfoVisQA: A Multilingual Benchmark for Information-Rich Visual Question Answering},
 pdfauthor={Changjiang Jiang, Jiang Zhou, Xinbin Yuan, Fan Zhang, Tao Feng, Zhuohan Xie}}
\newcommand{\benchmark}{MInfoVisQA}
\newcommand{\todo}[1]{}
\setlist{itemsep=3pt,topsep=5pt,leftmargin=1.5em}
\titleformat{\section}{\large\bfseries\color{Navy}}{\thesection}{0.65em}{}
\titleformat{\subsection}{\normalsize\bfseries\color{Navy}}{\thesubsection}{0.65em}{}
\titlespacing*{\section}{0pt}{15pt plus 3pt}{7pt}
\titlespacing*{\subsection}{0pt}{11pt plus 2pt}{5pt}

\fancypagestyle{cover}{\fancyhf{}\fancyfoot[R]{\small\thepage}}
\newcommand{\resource}[3]{\href{#1}{\tcbox[on line,colback=white,colframe=Rule,boxrule=0.5pt,arc=2pt,left=8pt,right=8pt,top=4pt,bottom=4pt,boxsep=0pt]{\raisebox{-2.5pt}{\includegraphics[height=12pt]{assets/#2.pdf}}\hspace{6pt}\small\textbf{#3}}}}
\renewenvironment{abstract}{\par\smallskip\noindent\ignorespaces}{\par}
\begin{document}
\thispagestyle{cover}
\noindent\raisebox{-0.5\height}{\includegraphics[width=37mm]{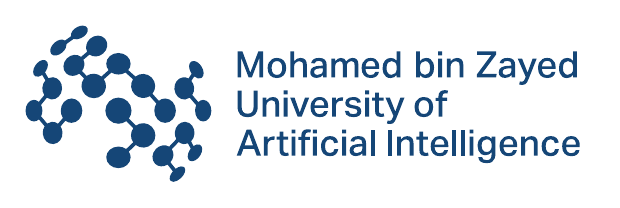}}\hfill
\raisebox{-0.5\height}{\includegraphics[width=36mm]{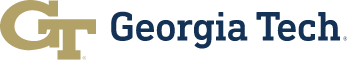}}\hfill
\raisebox{-0.5\height}{\includegraphics[width=31mm]{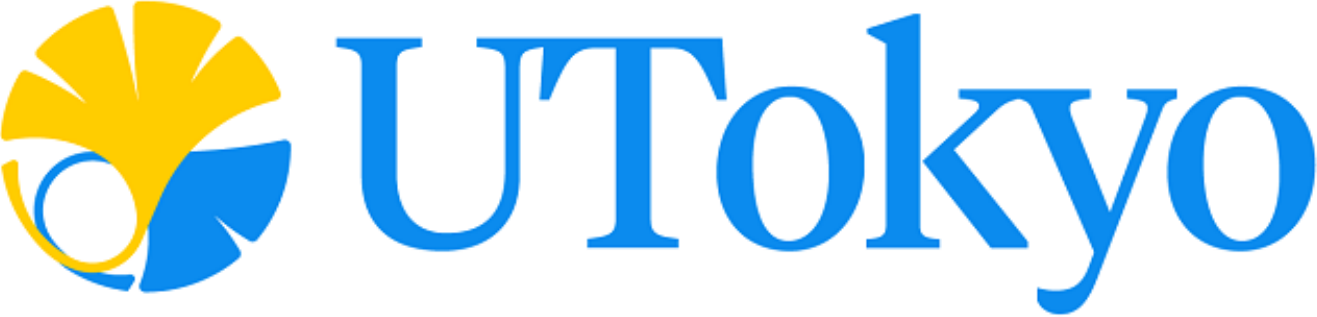}}\hfill
\raisebox{-0.5\height}{\includegraphics[width=44mm]{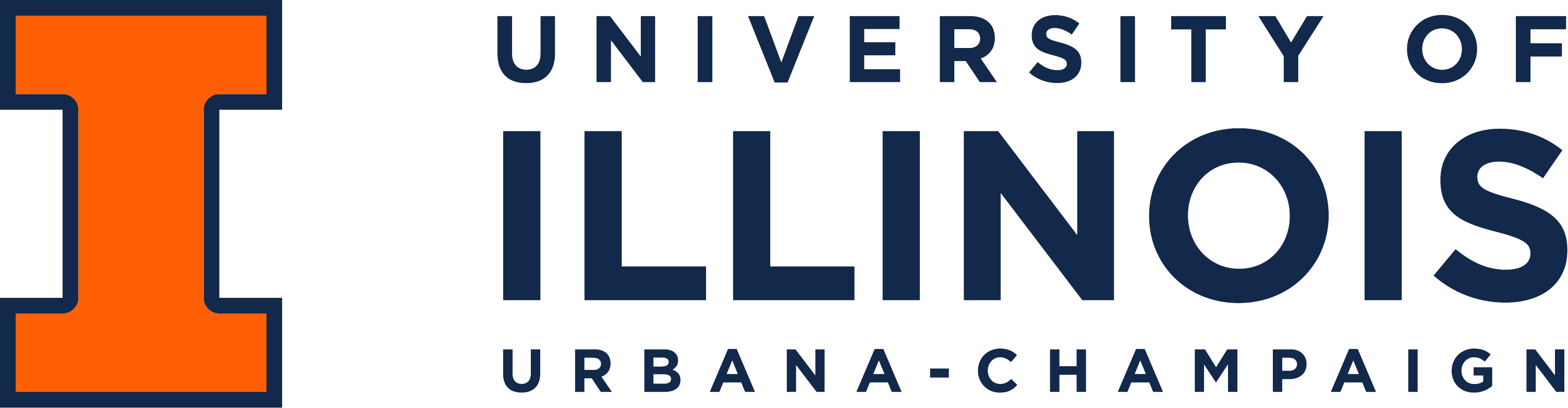}}
\par\vspace{5pt}{\color{Rule}\hrule}\vspace{9pt}
\begin{tcolorbox}[colback=Mist,colframe=Mist,boxrule=0pt,arc=3pt,left=10pt,right=10pt,top=10pt,bottom=10pt]
{\fontsize{21}{23}\selectfont\bfseries\color{Navy}MInfoVisQA: A Multilingual Benchmark\\
for Information-Rich Visual Question Answering\par}
\vspace{8pt}
{\fontsize{10.5}{13}\selectfont\bfseries
Changjiang Jiang\textsuperscript{1}, Jiang Zhou\textsuperscript{2}, Xinbin Yuan\textsuperscript{1}, Fan Zhang\textsuperscript{1,3},\\
Tao Feng\textsuperscript{4}, Zhuohan Xie\textsuperscript{1,*}\par}
\vspace{4pt}
{\fontsize{9}{11.5}\selectfont
\textsuperscript{1}Mohamed bin Zayed University of Artificial Intelligence\quad
\textsuperscript{2}Georgia Institute of Technology\\
\textsuperscript{3}The University of Tokyo\quad
\textsuperscript{4}University of Illinois at Urbana-Champaign\\[3pt]
\textsuperscript{*}\textbf{Corresponding author:} \href{mailto:zhuohan.xie18@gmail.com}{zhuohan.xie@mbzuai.ac.ae}\par}
\vspace{7pt}
\resource{https://github.com/arnodjiang/MInfoVisQA}{github}{GitHub}\hspace{5pt}
\resource{https://arnodjiang.github.io/MInfoVisQA/}{project}{Project Page}\hspace{5pt}
\resource{https://huggingface.co/datasets/arnodjiang/MInfoVisQA}{huggingface}{Benchmark}
\vspace{5pt}
{\fontsize{10.5}{13}\selectfont\begin{abstract}
Multimodal large language models (MLLMs) excel in vision-language tasks; yet, whether they maintain consistent understanding of information-rich visuals (InfoVis) when the surface language changes remains unclear. To quantify this capability, we introduce \emph{MInfoVisQA}, a parallel benchmark with 8,960 QA configurations derived from 128 seed instances across 24 languages. We construct seed-matched, language-parallel visuals and QA pairs by disentangling numerical-structural data from text and programmatically synchronizing translated labels. We evaluate 25 MLLMs under two settings: Language-Aligned QA (LQA, matching visual-text languages) and Cross-Language QA (XQA, fixed QA language with varying visual languages). The results show that aggregate accuracy masks substantial robustness gaps: two models with comparable mean LQA accuracy (80.4\%) exhibit performance spreads of 7.8 and 20.3 percentage points across languages. Moreover, all 25 models score lower with non-English visuals under fixed English QA, underscoring the need for cross-language sensitivity metrics.
\end{abstract}

}
\end{tcolorbox}
\vspace{3pt}
\begingroup\centering
\includegraphics[width=\linewidth]{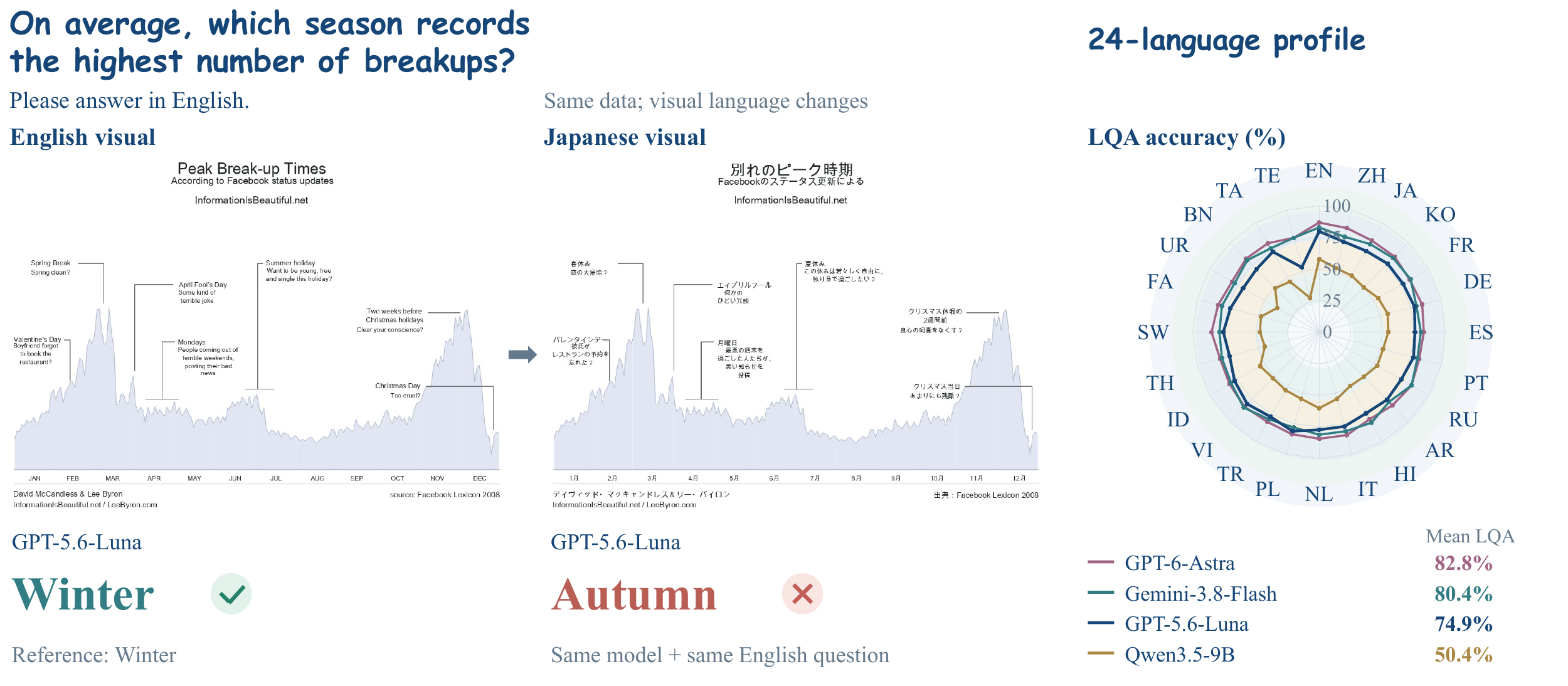}
\captionof{figure}{\textbf{MInfoVisQA overview.} With the English question and model fixed, GPT-5.6-Luna answers ``Winter'' on the English visual and ``Autumn'' on its Japanese counterpart (reference: ``Winter''). This recorded example uses identical numeric data. Right: LQA accuracy across 24 languages for four selected models, with their language-averaged accuracies in the legend (Table~\ref{tab:minfovisqa-main}).}
\label{fig:fig1_mot}
\endgroup
\clearpage
\section{Introduction}

Multimodal Large Language Models (MLLMs) have demonstrated capabilities in visual perception~\citep{team2026polyocr,cite_segui} and multilingual text understanding~\citep{zhou2026incentivizing}. Charts and tables in scientific publications and financial reports condense quantitative information into visual representations~\citep{xie-etal-2026-finchain}. Understanding these information-rich visuals (InfoVis) requires models to connect textual labels with numerical relationships and spatial
structure~\citep{charxiv,masry-etal-2025-chartqapro,jiang2026think}. Yet, multilingual text understanding alone does not establish whether models can interpret such visuals consistently across languages. Can an MLLM preserve its factual and numerical understanding when the language changes but the underlying information remains the same?

Figure~\ref{fig:fig1_mot} illustrates language-parallel visuals and the performance differences that can persist across languages, including when the question and answer language is fixed to English. Recent work has extended multilingual visual evaluation~\citep{tang-etal-2025-mtvqa,geigle-etal-2025-centurio,polychartqa}. However, comparisons across unmatched subsets can confound language related difficulties with differences in visual content and task complexity. We compare answer correctness on shared seeds across languages, while recognizing that translation and rendering can affect these comparisons. Moreover, information visuals are inherently heterogeneous: charts require separating visual rendering code from plotted parameters, whereas tables range from structured formats (HTML/Markdown) to flat raster images with implicit topology~\citep{xing2025mmtu,kim2024tablevqa}.

To address these challenges, we introduce an agent-based data synthesis framework that decouples structural-numerical content from textual semantics across distinct visual modalities. For charts, our framework extracts executable plotting programs, isolates textual elements for contextual translation, and re-executes the code to produce natively rendered target-language visuals. For tables, it parses structured cells directly or employs topology recovery for image-only instances before performing localized cell translation. Accompanying QA pairs are synchronized with the translated visual labels under rigorous structural and numerical sanity checks.

Leveraging this framework, we construct \benchmark{}, a multilingual benchmark spanning 24 languages and 8,960 QA configurations from 128 seeds across two complementary protocols: Language-Aligned QA (LQA, visual, query, and answer in the same language) to assess aligned multilingual performance, and Cross-Language QA (XQA, mismatched visual and query languages) to assess accuracy with a fixed QA language. 

Our contributions can be summarized as follows:

\begin{itemize}
    \item We design an agent-based synthesis pipeline that unifies chart code reconstruction and table cell-structure recovery, supporting multilingual rendering with structural checks and human review.
	\item We release \benchmark{}, a seed-matched benchmark covering charts and tables in 24 languages under LQA and XQA settings, extending language and visual-modality coverage.
	\item We evaluate 25 MLLMs, documenting language-associated accuracy differences and correct-coverage patterns for ten models and subgroup patterns for nine models on the shared seeds.
\end{itemize}

\section{Related Work}

\subsection{Information-Rich Visual QA}

With the development of MLLMs, multimodal understanding has expanded to encompass tasks such as chart understanding~\citep{jiang2026think} and table question answering~\citep{jiang2025tabdsr}. Chart question answering requires models to interpret visual representations and answer the query. ChartQA~\citep{masry-etal-2022-chartqa} combines visual and logical reasoning over real-world charts, whereas ChartQAPro~\citep{masry-etal-2025-chartqapro} broadens chart sources and question formats to increase evaluation difficulty. CharXiv~\citep{charxiv} focuses on scientific charts, pairing descriptive questions with reasoning tasks. ChartMuseum~\citep{chartmuseum} emphasizes expert-authored questions that require reasoning over both visual and textual evidence. These benchmarks assess complementary aspects of chart understanding, primarily in English.

\begin{table}[h]
\centering
\resizebox{0.72\linewidth}{!}{
\begin{tabular}{@{}lrrrrr@{}}
\toprule
\textbf{Dataset} & \multicolumn{2}{c}{\textbf{Test QA}} &
\multicolumn{2}{c}{\textbf{Lang.}} & \textbf{Image Types} \\
\cmidrule(lr){2-3}\cmidrule(lr){4-5}
& Chart & Table & Q & V & \\
\midrule
MMTU~\citep{xing2025mmtu} & 0 & 8,921 & 1 & 1 & -- \\
TableVQA-Bench~\citep{kim2024tablevqa} & 0 & 1,500 & 1 & 1 & -- \\
ChartQA~\citep{masry-etal-2022-chartqa} & 2,500 & 0 & 1 & 1 & 3 \\
ChartQAPro~\citep{masry-etal-2025-chartqapro} & 1,948 & 0 & 1 & 1 & 9 \\
CharXiv~\citep{charxiv} & 6,615 & 0 & 1 & 1 & -- \\
ChartMuseum~\citep{chartmuseum} & 1,000 & 0 & 1 & 1 & -- \\
\midrule
MTVQA~\citep{tang-etal-2025-mtvqa} & -- & -- & 9 & 10 & 20 \\
SMPQA~\citep{geigle-etal-2025-centurio} & 14,300 & 0 & 1 & 11 & 2 \\
PolyChartQA~\citep{polychartqa} & 26,151 & 0 & 10 & 10 & 16 \\
\midrule
\rowcolor{Sand!35}\textbf{\benchmark (ours)} & 6,090 & 2,870 & \textbf{24} & \textbf{24} & 32 \\
\bottomrule
\end{tabular}
}
\caption{Comparison of relevant benchmarks.}
\label{tab:benchmark-comparison}
\end{table}

Question answering over table images further requires resolving cell contents and their row and column relationships. TableVQA-Bench~\citep{kim2024tablevqa} integrates table images and question--answer annotations from multiple datasets and compares visual and textual input formats. Visual-TableQA~\citep{visual-tableqa} uses synthetic table images and generated questions to assess reasoning over visually structured tabular data. Beyond image-based QA, MMTU~\citep{xing2025mmtu} evaluates a broader range of table understanding and reasoning tasks. Our work extends chart and table QA to aligned multilingual evaluation, using chart program reconstruction and cell-level table processing to translate visual text and render corresponding language variants. Table~\ref{tab:benchmark-comparison} compares existing benchmarks by chart and table coverage, question and visual-text languages, and visual categories. In addition to scaling up language coverage to 24 languages, our benchmark encompasses a broader spectrum of InfoVis types with finer-grained visual table categorizations. Although more compact in sample size compared to PolyChartQA~\citep{polychartqa}, our benchmark offers substantially richer diversity in both language and visual modalities.

\subsection{Multilingual Visual Understanding}

InfoVisQA is an active direction in adapting MLLMs to information-visual Reasoning tasks~\citep{masry-etal-2022-chartqa}, requiring models to extract values, interpret visual relationships, and reason over quantitative evidence. Recent chart-specialized approaches explore complementary strategies. Chart-R1~\citep{chartr1} combines programmatically synthesized reasoning data with chain-of-thought supervision and numerically sensitive reinforcement fine-tuning. ChartVerse~\citep{chartverse} scales training through complexity-aware chart synthesis and answer-first QA generation grounded in executable chart code. ChartSketcher~\citep{Chartsketcher} introduces visual feedback into reasoning by annotating charts with intermediate sketches and revisiting these annotations to refine subsequent steps. Multilingual multimodal research examines how training data and model design affect visual understanding across languages, such as Centurio~\citep{geigle-etal-2025-centurio} and MTVQA~\citep{tang-etal-2025-mtvqa}. Such data capture realistic language use, but independently collected language subsets need not contain equivalent visual content or reasoning tasks. Comparisons across these subsets can therefore mix language effects with differences in content and difficulty.

Controlled and cross-language evaluations address complementary parts of this problem. SMPQA~\citep{geigle-etal-2025-centurio} uses synthetic plots with matched configurations across visual languages and English question templates to assess text-in-image understanding. XT-VQA~\citep{xtvqa} evaluates mismatches between the question language and the language of text within the image, reporting lower performance under cross-language inputs. PolyChartQA~\citep{polychartqa} combines coordinated translation with code-based rendering to construct multilingual charts and question--answer pairs, and examines cross-language performance by varying chart and question languages. These studies motivate evaluating visual-language alignment explicitly rather than inferring it from multilingual text capabilities. We extend coverage to 24 languages and tables, emphasizing cell-structure recovery and cell translation. PolyChartQA covers charts in ten languages. Appendix~\ref{app:contribution} discusses these contributions and their scope. We provide a detailed review of research in this area in Appendix~\ref{app:research-review}.

\section{MInfoVisQA}

\begin{figure}[tbp]
  \includegraphics[width=0.95\linewidth]{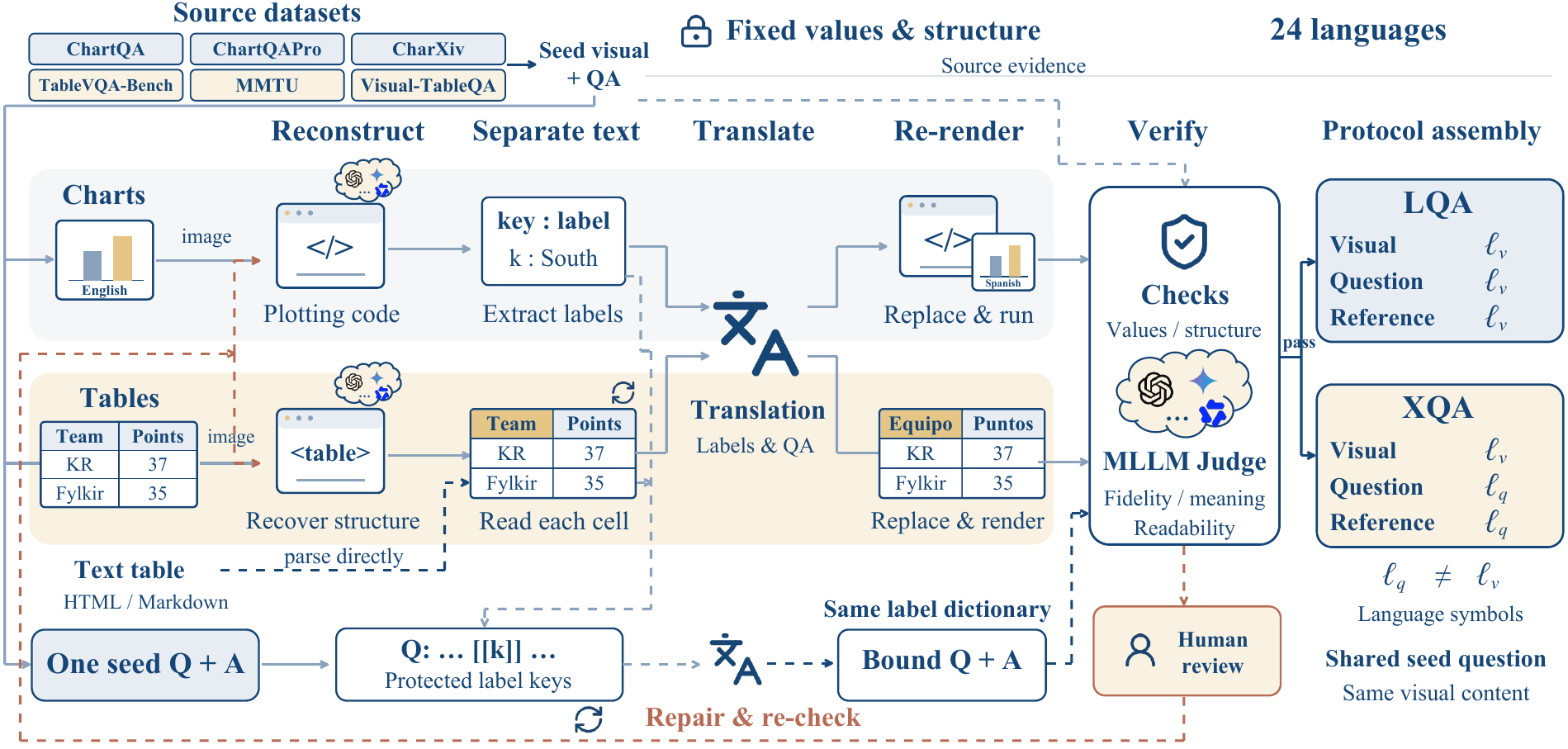}
    \caption{Construction of parallel multilingual visuals and QA for \benchmark{}; $\ell_v$ and $\ell_q$ denote visual and query languages, respectively. The prompts can be seen in Appendix~\ref{app:prompts}.}
  \label{fig:fig2_pipeline}
\end{figure}

We present \benchmark{}, a benchmark for InfoVisQA across 24 languages. LQA uses the same language for the visual, question, and answer. XQA uses Chinese or English for the question and answer, with a different visual language. Figure~\ref{fig:fig2_pipeline} summarizes construction. Examples and translation-cost estimates appear in Appendix~\ref{app:benchmark-examples} and Appendix~\ref{ap:api_cost}.

\subsection{Seeds and Coverage}

\begin{figure}[tbp]
  \centering
  \includegraphics[width=0.48\textwidth]{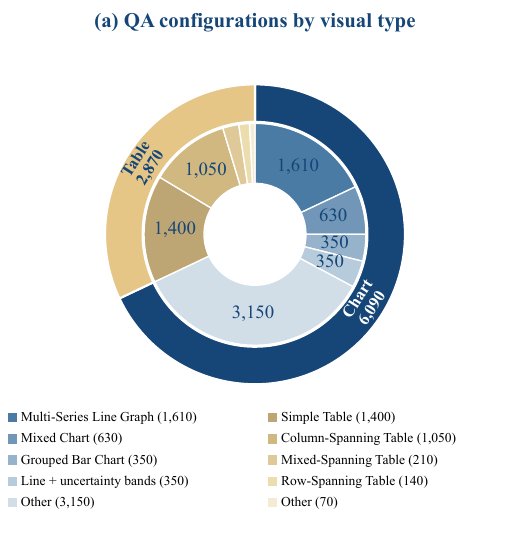}
  \hfill
  \includegraphics[width=0.48\textwidth]{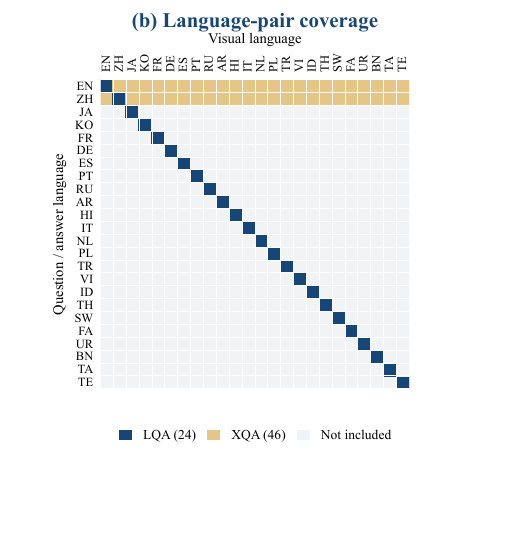}
  \caption{Composition and language coverage of \benchmark{}.
  (a) The outer ring shows chart and table counts; the inner ring
  shows the four most frequent types within each family.
  (b) Language pairs included in LQA and XQA.}
  \label{fig:benchmark-coverage}
\end{figure}

Seed instances were drawn from CharXiv~\citep{charxiv}, ChartQAPro~\citep{masry-etal-2025-chartqapro}, ChartQA~\citep{masry-etal-2022-chartqa}, TableVQA-Bench~\citep{kim2024tablevqa}, Visual-TableQA~\citep{visual-tableqa}, and MMTU~\citep{xing2025mmtu}, covering charts and tables with varied layouts and reasoning requirements. We represent a seed as $s_i=(V_i,Q_i,A_i,C_i)$, where $V_i$ denotes the source visual evidence, $Q_i$ the question, $A_i$ the reference answer, and $C_i$ optional supporting prose. Selection was restricted to source-linked, single-turn QA. It excluded the VTabFact subset of TableVQA-Bench and used the test subset of Visual-TableQA. Known erroneous and unsupported records were removed before construction. Available source-group identifiers prevented repeated selection of the same figure or table within the sampling pool. Reconstruction and translation artifacts were bound to this identity and the source-image hashes to prevent mismatches between visuals and QA. We retained 128 seeds: 87 charts and 41 tables. Figure~\ref{fig:benchmark-coverage} summarizes visual-family composition and language-pair coverage. Selection details can be seen in Appendix~\ref{app:selection}.

\subsection{Multilingual Data Construction}

\paragraph{Visual Reconstruction.}

Each seed instance was represented by a language independent numerical and structural specification $D_i$, a rendering procedure $R_i$, and a dictionary $L_i^{\mathrm{en}}$ mapping stable label keys to English text. These components define the English reconstruction, $\widetilde V_i^{\mathrm{en}}=R_i(D_i,L_i^{\mathrm{en}})$. A shared specification ensures consistency across variants, but source fidelity requires separate verification.

\paragraph{Charts and Tables.}
For charts, the model received the source image without QA. Following Chart-R1~\citep{chartr1} and PolyChart~\citep{polychartqa}, it reconstructed executable Python code with separately stored text strings. Translated labels were substituted before re-rendering. For tables, we parsed available HTML or Markdown directly, otherwise recovering cells from the image. Cell text was translated and re-rendered while retaining table structure. Both branches underwent the checks below.

\paragraph{Translation and QA alignment.}
Google Translate provided the initial translations of visual labels, questions, answers, and required context. Stable label keys linked mentions in question and answer templates to the same translated dictionary used for rendering. Numerical content and protected references were retained across languages, while textual labels and unit names were translated. Query refinement used answer blind copyediting followed by a separate semantic equivalence review of proposed edits. This review checked that editing preserved the requested quantity, entities, comparisons, negation, and temporal scope; substantive ambiguities were flagged rather than silently resolved. Model assessments informed subsequent human inspection rather than establishing translation correctness on their own.

\paragraph{Source context.}
Questions requiring prose retained the ChartQAPro paragraph or MMTU/FinQA pre and post table text, translated into the query language. Numeric placeholders and paragraph order were checked. Reference answers, rationales, and gold evidence selectors were excluded. In XQA, context and label mentions use the query language. Construction and evaluation prompts can be seen in Appendix~\ref{app:prompts}.

\subsection{Translation Quality and Human Review}
\label{sec:quality-control}

\paragraph{Programmatic and model checks.}
Programmatic safeguards bound reconstructed visuals and QA to source identifiers and image hashes, and checked protected numeric placeholders and their multiplicity. These checks detect structural inconsistencies; they do not establish semantic equivalence or linguistic naturalness. LLM-based review separately compared reconstructed visuals with source evidence and assessed answer preservation, translation meaning, label correspondence, and rendered readability. The review prompts addressed numbers, units, conditions, category distinctions, and complex-script shaping. Source fidelity and translation quality were evaluated separately, since correct translation cannot repair an inaccurate reconstruction. Generation and review could use the same model, so separate calls did not constitute independent model validation. This visual construction review was also distinct from the text-only answer judge used in evaluation.

\paragraph{Human judgments and tool assistance.}
Annotators inspected the source visual, reconstructed image, and associated QA pairs for visual completeness, numerical and structural fidelity, and readability. For translation review, annotators verified semantic equivalence and label alignment between visuals and text: translated categories had to point unambiguously to the identical visual evidence, while questions and answers had to preserve their original intent. Google Translate and the Gemini interface served as auxiliary tools for multilingual verification. Additionally, we report demographic and background details of the annotators, including their native languages and domain expertise in Appendix~\ref{ap:rback}.

\paragraph{Correction and revalidation.}
Annotators adjudicated automatic flags against source evidence, dismissing false positives and identifying confirmed errors. We revised the relevant construction prompts and regenerated affected instances through the same framework, rather than manually patching final images. Confirmed errors were repaired, with further regeneration when needed, and rechecked before inclusion. Human diagnosis informed reproducible construction constraints. A documented repair restored question graphical evidence omitted from a table image. The revised prompt required the complete visual and its label associations, and regenerated versions were reinspected for answer support. This example demonstrates reconstruction repair rather than a measured gain in translation accuracy. The source, repaired renderings, and case details can be seen in Appendix~\ref{app:erroranalysis}.

\paragraph{Recorded outcomes and limits.}

The initial judgments classified 68 items as passes, 48 as failures, and 12 as uncertain. Automatic screening accepted 1,631 of 8,960 configurations (18.2\%) and flagged the remaining 7,329 (81.8\%) for review. The reviewers reported inspecting all 8,960 configurations and ultimately accepting them all. Screening flags indicate potential issues rather than confirmed errors, while final acceptance reflects the review outcome and does not provide an independent estimate of correctness. We did not retain correction counts for each review round and therefore could not quantify how often corrections were made or how much they improved the results. Appendix~\ref{app:humancheck} provides details of the review.

\Needspace{7\baselineskip}
\section{Experiments}

\subsection{Experimental Setup}

\paragraph{Models.}
We evaluated 25 MLLMs: five proprietary, 16 general-purpose, and four chart-specialized models, as listed in the main table.
\textit{Proprietary MLLMs} comprise GPT-6-Astra~\citep{gpt6}, GPT-5.6-Sol~\citep{gpt56}, GPT-5.6-Luna~\citep{gpt56}, GPT-5.5~\citep{gpt55} and Gemini-3.8-Flash~\citep{gemini38}.
\textit{General-purpose open-source MLLMs} comprise Kimi-K3~\citep{team2026kimi}, MiniMax-M3~\citep{minimaxm3}, DeepSeek-V4.1-Flash~\citep{xu2026deepseek}, Qwen series~\citep{qwen35,qwen38}, Vision-OPD~\citep{yuan2026vision}, and ZwZ~\citep{wei2026zooming}.
\textit{Chart-specialized MLLMs} comprise Chart-R1~\citep{chartr1}, ChartVerse~\citep{chartverse} and ChartSketcher~\citep{Chartsketcher}.

\paragraph{Evaluation Setup.}
The maximum output length was 10,240 tokens. The temperature was set to 0.4 for all evaluated models. Deterministic answer matching was followed by a text-only GPT-6-Astra judge for non-matches, assessing semantic equivalence rather than exact wording. Judge sensitivity is reported in Appendix~\ref{ap:judge_ab}.

\subsection{Evaluation Metrics}
\label{sec:evaluation-metrics}

Let $y_{i,v,q}\in\{0,1\}$ denote the final answer correctness for
seed $i$, visual language $v$, and QA language $q$.
Exact matches or answers judged semantically equivalent receive
credit; unresolved failures count as incorrect.
For a seed subset $\mathcal{S}$, the configuration accuracy is defined as:
\begin{equation}
 A_{\mathcal{S}}(v,q)
 = \frac{100}{|\mathcal{S}|}
   \sum_{i\in\mathcal{S}} y_{i,v,q}.
\end{equation}

We omit the subscript when using all 128 seeds.
With $\mathcal{L}$ denoting the 24 languages, the main-table
metrics are
\begin{align}
 \mathrm{LQA}_{\ell} &= A(\ell,\ell), \\
 \mathrm{XQA}_{q} &= \frac{1}{23}
   \sum_{v\in\mathcal{L}\setminus\{q\}} A(v,q), \\
 \mathrm{AVG} &= \frac{1}{70}\Biggl[
   \sum_{\ell\in\mathcal{L}}\mathrm{LQA}_{\ell}
   \notag\\
 &\qquad {}+23\!
   \sum_{q\in\{\mathrm{EN},\mathrm{ZH}\}}
   \mathrm{XQA}_{q}\Biggr].
\end{align}

Thus, AVG weight all 70 distinct configurations equally.
The LQA heatmap reports $\mathrm{LQA}_{\ell}$, while subgroup
LQA scores average $A_{\mathcal{S}}(\ell,\ell)$ over 24 languages.

For XQA diagnostics, we measure the change relative to the
same-language visual baseline:
\begin{equation}
 \Delta_{\mathcal{S},q}
 = \frac{1}{23}\sum_{v\ne q}A_{\mathcal{S}}(v,q)
   - A_{\mathcal{S}}(q,q).
\end{equation}

Subgroup forest plots report differences between these
$\Delta$ values: Language-bearing minus Translation-invariant,
and Spanning minus Simple, in percentage points.

Correct coverage measures the proportion of seeds answered
correctly in at least $k$ visual languages:
\begin{equation}
\begin{aligned}
 C_q(k)
 &= \frac{100}{128}\sum_{i=1}^{128}
 \mathbf{1}\!\left[
 \sum_{v\in\mathcal{L}}y_{i,v,q}\geq k
 \right],\\
 &\hspace{2em} k=1,\ldots,24.
\end{aligned}
\end{equation}

In particular, $C_q(24)$ measures the correctness across all visual
languages. We estimate 95\% confidence intervals using
10,000 seed-level bootstrap samples with random seed 20261002,
retaining all language variants of each sampled seed.
For subgroup contrasts, the two groups are resampled independently.

\begin{table*}[h]
  \centering
  \begingroup
  \resizebox{0.95\linewidth}{!}{
  \begin{tabular}
    {@{\extracolsep{\fill}}l*{27}{c}@{}}
    \toprule
    \multirow{2}{*}{\textbf{Model}}
      & \multicolumn{2}{c}{\textbf{XQA}}
      & \multicolumn{24}{c}{\textbf{LQA}}
      & \multirow{2}{*}{\textbf{AVG}} \\
    \cmidrule(lr){2-3}\cmidrule(lr){4-27}
      & ZH & EN
      & EN & ZH & JA & KO & FR & DE & ES & PT
      & RU & AR & HI & IT & NL & PL & TR & VI
      & ID & TH & SW & FA & UR & BN & TA & TE
      & \\
    \midrule

    \multicolumn{28}{l}{\textit{Proprietary MLLMs}} \\

    GPT-6-Astra
      & 80.3 & 82.5
      & 86.7 & 85.2 & 83.6 & 84.4
      & 82.8 & 84.4 & 82.8 & 82.0
      & 84.4 & 82.0 & 79.7 & 84.4
      & 84.4 & 83.6 & 82.0 & 83.6
      & 82.0 & 81.2 & 85.2 & 82.8
      & 79.7 & 82.0 & 81.2 & 77.3
      & 81.9 \\

    Gemini-3.8-Flash
      & 78.1 & 81.3
      & 82.8 & 78.1 & 80.5 & 82.8
      & 83.6 & 79.7 & 80.5 & 80.5
      & 84.4 & 78.1 & 82.8 & 81.2
      & 81.2 & 78.1 & 79.7 & 84.4
      & 80.5 & 79.7 & 78.9 & 79.7
      & 77.3 & 80.5 & 76.6 & 77.3
      & 79.9 \\

    GPT-5.6-Sol
      & 77.5 & 78.7
      & 81.2 & 79.7 & 81.2 & 78.1
      & 81.2 & 80.5 & 80.5 & 78.1
      & 78.9 & 78.1 & 80.5 & 78.1
      & 80.5 & 80.5 & 79.7 & 78.9
      & 78.9 & 77.3 & 78.9 & 78.1
      & 77.3 & 78.1 & 74.2 & 70.3
      & 78.3 \\

    GPT-5.6-Luna
      & 73.7 & 73.3
      & 79.7 & 74.2 & 74.2 & 76.6
      & 76.6 & 78.1 & 75.8 & 77.3
      & 75.0 & 75.8 & 74.2 & 77.3
      & 77.3 & 81.2 & 77.3 & 80.5
      & 77.3 & 73.4 & 75.8 & 72.7
      & 69.5 & 70.3 & 73.4 & 53.1
      & 74.0 \\
GPT-5.5
      & 71.8 & 75.5
      & 79.7 & 73.4 & 77.3 & 79.7
      & 78.9 & 75.0 & 74.2 & 76.6
      & 78.1 & 75.0 & 74.2 & 78.9
      & 74.2 & 77.3 & 75.8 & 80.5
      & 75.0 & 71.1 & 74.2 & 73.4
      & 70.3 & 70.3 & 68.8 & 61.7
      & 74.0 \\
    \midrule
    \multicolumn{28}{l}
      {\textit{General-purpose Open-source MLLMs}} \\

    Kimi-K3
      & 77.2 & 79.9
      & 83.6 & 81.2 & 82.8 & 82.8
      & 78.9 & 78.9 & 78.1 & 79.7
      & 82.8 & 83.6 & 81.2 & 82.0
      & 80.5 & 82.8 & 81.2 & 84.4
      & 81.2 & 82.0 & 84.4 & 81.2
      & 78.9 & 78.1 & 75.8 & 64.1
      & 79.2 \\

    Vision-OPD-9B
      & 66.8 & 64.7
      & 78.1 & 75.8 & 68.0 & 68.0
      & 64.8 & 75.0 & 69.5 & 73.4
      & 68.0 & 64.1 & 65.6 & 71.1
      & 71.9 & 71.1 & 71.9 & 72.7
      & 73.4 & 68.8 & 66.4 & 64.1
      & 58.6 & 66.4 & 57.0 & 40.6
      & 66.4 \\

Qwen3.8-Flash-Next
& 59.5 & 64.3
& 74.2 & 71.9 & 70.3 & 68.8
& 71.1 & 70.3 & 73.4 & 67.2
& 74.2 & 63.3 & 66.4 & 74.2
& 71.9 & 64.1 & 73.4 & 77.3
& 71.9 & 62.5 & 68.8 & 60.2
& 59.4 & 66.4 & 51.6 & 46.9
& 63.8 \\

    MiniMax-M3
      & 61.5 & 62.5
      & 77.3 & 67.2 & 64.1 & 63.3
      & 68.0 & 71.1 & 71.9 & 69.5
      & 69.5 & 60.9 & 57.8 & 71.1
      & 68.8 & 72.7 & 70.3 & 69.5
      & 71.9 & 59.4 & 66.4 & 59.4
      & 57.0 & 52.3 & 51.6 & 46.9
      & 63.0 \\

    Vision-OPD-4B
      & 57.9 & 62.4
      & 73.4 & 75.0 & 61.7 & 65.6
      & 64.1 & 71.1 & 71.1 & 69.5
      & 66.4 & 58.6 & 61.7 & 72.7
      & 68.8 & 62.5 & 63.3 & 69.5
      & 71.9 & 59.4 & 52.3 & 55.5
      & 51.6 & 50.0 & 45.3 & 39.1
      & 61.0 \\
      Qwen3.5-397B-A17B
& 55.1 & 54.2 & 67.2 & 66.4 & 68.0 & 64.1 & 66.4 & 64.1 & 66.4 & 64.8 & 68.8 & 64.1 & 65.6 & 64.8 & 66.4 & 60.9 & 64.8 & 64.8 & 64.1 & 64.1 & 64.1 & 64.1 & 64.8 & 66.4 & 57.0 & 46.9 & 57.9 \\
Qwen3.5-122B-A10B
& 49.3 & 50.4 & 65.6 & 63.3 & 59.4 & 54.7 & 60.9 & 60.9 & 62.5 & 63.3 & 62.5 & 57.8 & 57.8 & 64.1 & 61.7 & 60.9 & 59.4 & 59.4 & 60.9 & 59.4 & 57.8 & 53.1 & 46.1 & 59.4 & 50.8 & 39.8 & 52.8 \\

    DeepSeek-V4.1-Flash
      & 48.1 & 52.9
      & 59.4 & 51.6 & 52.3 & 52.3
      & 59.4 & 59.4 & 62.5 & 62.5
      & 54.7 & 52.3 & 56.2 & 62.5
      & 57.0 & 57.8 & 56.2 & 59.4
      & 57.8 & 48.4 & 50.8 & 51.6
      & 44.5 & 44.5 & 46.1 & 44.5
      & 51.8 \\

    ZwZ-8B
      & 49.9 & 51.5
      & 61.7 & 59.4 & 56.2 & 56.2
      & 56.2 & 57.8 & 58.6 & 59.4
      & 55.5 & 42.2 & 52.3 & 62.5
      & 59.4 & 56.2 & 54.7 & 57.8
      & 57.0 & 50.8 & 37.5 & 42.2
      & 40.6 & 48.4 & 33.6 & 22.7
      & 51.0 \\
Qwen3.6-27B
& 48.1 & 48.2 & 60.9 & 57.8 & 57.8 & 54.7 & 62.5 & 63.3 & 56.2 & 64.1 & 58.6 & 57.8 & 58.6 & 63.3 & 57.8 & 56.2 & 62.5 & 59.4 & 62.5 & 54.7 & 60.2 & 50.8 & 53.1 & 56.2 & 39.1 & 29.7 & 51.0 \\
Qwen3.8-27B
& 45.6 & 50.4 & 64.1 & 60.9 & 57.8 & 52.3 & 60.9 & 60.9 & 53.9 & 55.5 & 60.2 & 59.4 & 57.8 & 60.2 & 60.2 & 58.6 & 60.9 & 61.7 & 57.8 & 50.8 & 53.1 & 56.2 & 53.9 & 51.6 & 30.5 & 28.1 & 50.5 \\
Qwen3.6-35B-A3B
& 46.6 & 47.2 & 61.7 & 55.5 & 51.6 & 50.8 & 57.0 & 56.2 & 57.0 & 60.9 & 62.5 & 53.9 & 52.3 & 56.2 & 57.0 & 58.6 & 52.3 & 57.0 & 53.9 & 51.6 & 54.7 & 52.3 & 45.3 & 53.9 & 46.1 & 36.7 & 49.3 \\

    Qwen3.5-9B
      & 41.9 & 43.9
      & 57.8 & 52.3 & 51.6 & 50.0
      & 53.9 & 56.2 & 54.7 & 52.3
      & 53.1 & 50.0 & 49.2 & 54.7
      & 60.2 & 54.7 & 53.1 & 51.6
      & 53.9 & 44.5 & 46.9 & 47.7
      & 38.3 & 49.2 & 46.1 & 28.1
      & 45.5 \\

    ZwZ-4B
      & 42.7 & 40.8
      & 57.0 & 60.9 & 49.2 & 46.9
      & 57.0 & 53.9 & 52.3 & 53.9
      & 48.4 & 40.6 & 41.4 & 57.8
      & 48.4 & 53.1 & 53.1 & 49.2
      & 53.1 & 41.4 & 33.6 & 45.3
      & 35.9 & 39.1 & 24.2 & 21.9
      & 43.4 \\

    Qwen3.5-4B
      & 39.1 & 40.9
      & 53.9 & 53.9 & 47.7 & 45.3
      & 48.4 & 47.7 & 51.6 & 50.8
      & 47.7 & 46.9 & 46.1 & 53.1
      & 57.0 & 46.9 & 50.0 & 47.7
      & 50.8 & 36.7 & 46.1 & 43.0
      & 43.0 & 43.8 & 36.7 & 33.6
      & 42.4 \\

    Qwen3.5-2B
      & 30.4 & 29.9
      & 46.9 & 46.1 & 31.2 & 35.9
      & 42.2 & 38.3 & 40.6 & 35.9
      & 39.8 & 31.2 & 37.5 & 40.6
      & 41.4 & 39.8 & 35.9 & 37.5
      & 38.3 & 29.7 & 25.0 & 33.6
      & 25.0 & 35.9 & 30.5 & 21.1
      & 32.1 \\

    \midrule
    \multicolumn{28}{l}{\textit{Chart-specialized MLLMs}} \\

    Chart-R1
      & 38.0 & 40.8
      & 58.6 & 51.6 & 54.7 & 57.8
      & 57.0 & 58.6 & 57.8 & 56.2
      & 50.8 & 53.1 & 46.9 & 55.5
      & 52.3 & 53.9 & 51.6 & 50.0
      & 50.0 & 48.4 & 35.2 & 45.3
      & 44.5 & 39.8 & 28.9 & 24.2
      & 42.8 \\

    ChartSketcher-72B
      & 36.0 & 37.0
      & 49.2 & 46.1 & 41.4 & 35.9
      & 44.5 & 40.6 & 39.1 & 40.6
      & 43.8 & 28.9 & 36.7 & 40.6
      & 45.3 & 41.4 & 51.6 & 39.1
      & 46.9 & 23.4 & 28.1 & 32.8
      & 28.9 & 33.6 & 11.7 & 12.5
      & 36.6 \\
ChartVerse-8B
& 16.8 & 16.8 & 35.2 & 32.0 & 35.2 & 14.8 & 30.5 & 36.7 & 32.8 & 25.8 & 21.9 & 20.3 & 5.5 & 35.9 & 29.7 & 36.7 & 24.2 & 21.9 & 40.6 & 13.3 & 15.6 & 14.1 & 7.8 & 7.8 & 8.6 & 5.5 & 18.9 \\
ChartVerse-4B
& 7.4 & 8.4 & 19.5 & 13.3 & 16.4 & 10.2 & 20.3 & 10.9 & 18.8 & 23.4 & 7.8 & 13.3 & 11.7 & 19.5 & 13.3 & 18.0 & 26.6 & 12.5 & 18.0 & 4.7 & 5.5 & 14.1 & 7.8 & 6.2 & 2.3 & 5.5 & 9.7 \\
    \bottomrule
  \end{tabular}
  }
  \endgroup

  \caption{
    Semantic answer accuracy (\%, $\uparrow$) on \benchmark{}.
    XQA-ZH/EN denotes the question and requested answer language.
  }
  \label{tab:minfovisqa-main}
\end{table*}

\subsection{Overall Performance}

\paragraph{Uneven language coverage.}

As shown in Table~\ref{tab:minfovisqa-main}, GPT-6-Astra achieves the highest overall accuracy of 81.9\%, followed by Gemini-3.8-Flash (79.9\%) and Kimi-K3 (79.2\%; Table~2).
 However, rankings alone understate differences in multilingual robustness. Gemini and Kimi both obtain approximately 80.4\% mean LQA accuracy, but their best-to-worst language gaps are 7.8 and 20.3 percentage points, respectively. Their Telugu accuracies differ by 13.2 points (77.3\% versus 64.1\%). Thus, similar mean performance can coexist with markedly different performance in individual languages. LQA ranges describe the complete aligned setting, including changes to the question and answer languages.

\paragraph{Visual-language sensitivity.}
To connect the results to our motivating question, we compare XQA with the
same-language baseline for each fixed QA language. All 25 models have lower
English XQA accuracy than English LQA accuracy. The decline is approximately
1.5 points for Gemini-3.8-Flash and 4.2 for GPT-6-Astra, but 13.0 for
Qwen3.5-397B-A17B and 17.8 for Chart-R1. These comparisons retain the seed questions and English QA. GPT-6-Astra nevertheless retains higher English XQA accuracy
than Gemini (82.5\% versus 81.3\%): a smaller decline and a higher absolute
score measure different aspects of performance. The observed changes include
the effects of translated text and its rendering; they do not identify a
specific recognition or reasoning failure.

\paragraph{QA-language dependence.}
MiniMax-M3 drops from 77.3\% to 62.5\% with English QA, compared with
67.2\% to 61.5\% with Chinese QA, corresponding to declines of 14.8 and
5.7 points. GPT-5.6-Luna similarly declines by 6.4 points with English QA
but only 0.5 with Chinese QA. Thus, the higher-scoring same-language setting
need not be the more stable setting under visual-language changes. These
contrasts are descriptive: English and Chinese XQA exclude different
same-language visuals, and their baselines differ. This indicates that current MLLMs' interpretation of visuals remains heavily influenced by embedded textual elements, falling short of genuine, invariant understanding of InfoVis semantics.

\begin{figure}[tbp]
  \centering
  \includegraphics[width=0.95\linewidth]{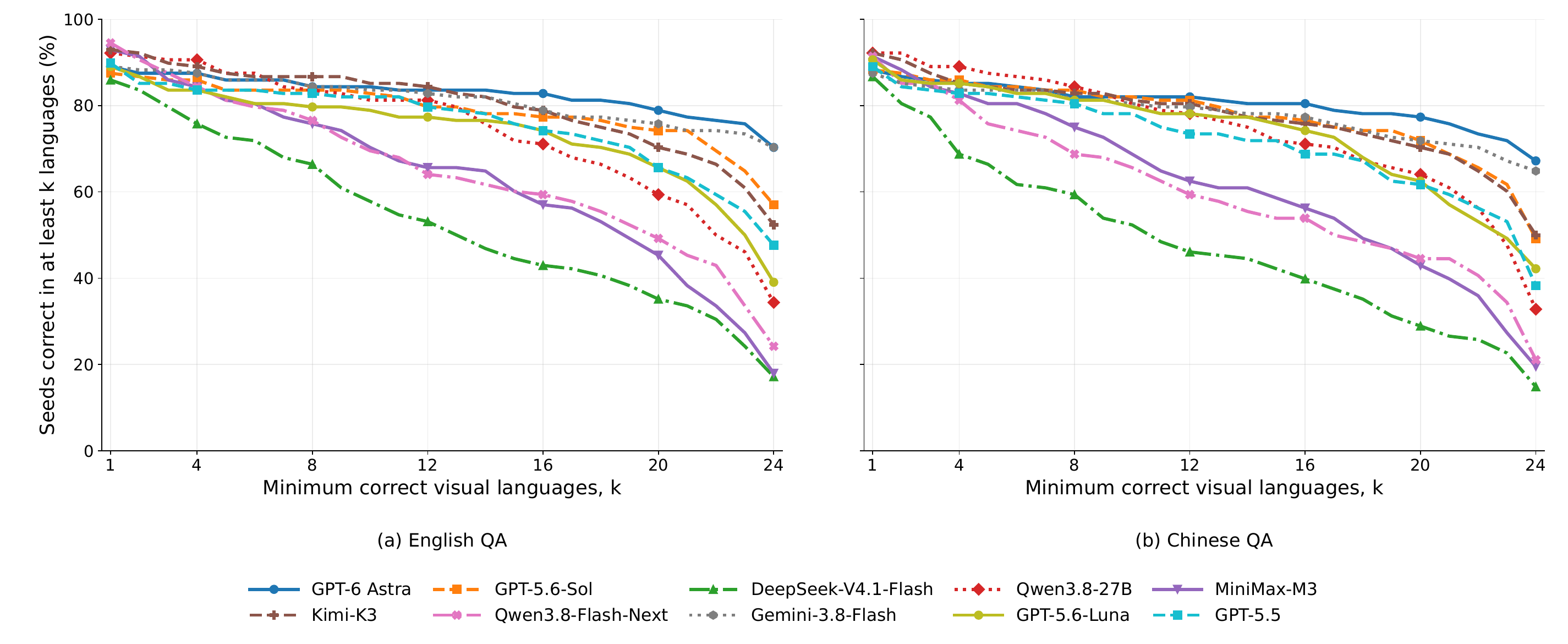}
  \caption{Cross-visual-language stability on MInfoVisQA.}
  \label{fig:fig4_stab}
\end{figure}

\paragraph{Chart specialization.}

Chart-R1 reaches 58.6\% on English LQA but 40.8\% on English XQA. Its
17.8-point decline indicates that performance in one aligned language does
not carry over unchanged to cross-language inputs. Its aggregate score of
42.8\%, however, combines charts and tables. Without chart-only comparisons
and matched model controls, this result cannot establish that specialization
itself reduces chart understanding or multilingual ability.

\subsection{Robustness Analysis}

\paragraph{Correct coverage.}

As shown in Figure~\ref{fig:fig4_stab}, the average accuracy does
not reveal whether the same seeds remain correct across
languages. Using the correct-coverage metric defined in
Section~\ref{sec:evaluation-metrics}, we compare ten models on success in
at least one visual language and across all 24.

Figure~\ref{fig:fig4_stab} shows that model ordering changes with the required coverage. Unlike the marginal accuracies in Table~\ref{tab:minfovisqa-main}, this measure tracks whether shared seeds remain correct across languages.

\paragraph{Subgroup comparisons.}
To examine whether visual-language sensitivity extends beyond answer translation, we compare Translation-invariant and Language-bearing answers. Even when answers require no translation, GPT-6-Astra and MiniMax-M3 lose 2.95 and 11.15 percentage points under English XQA, respectively. Language-bearing answers generally exhibit larger drops among the nine models, whereas spanning tables show no consistent additional degradation relative to simple tables. These descriptive patterns motivate separate reporting by answer type and table structure; the subgroups differ in seed content. Detailed results appear in Appendix~\ref{app:subgroup-diagnostics}.

\subsection{Qualitative Analysis}

\begin{figure}[tbp]
  \centering
  \includegraphics[width=0.95\linewidth]{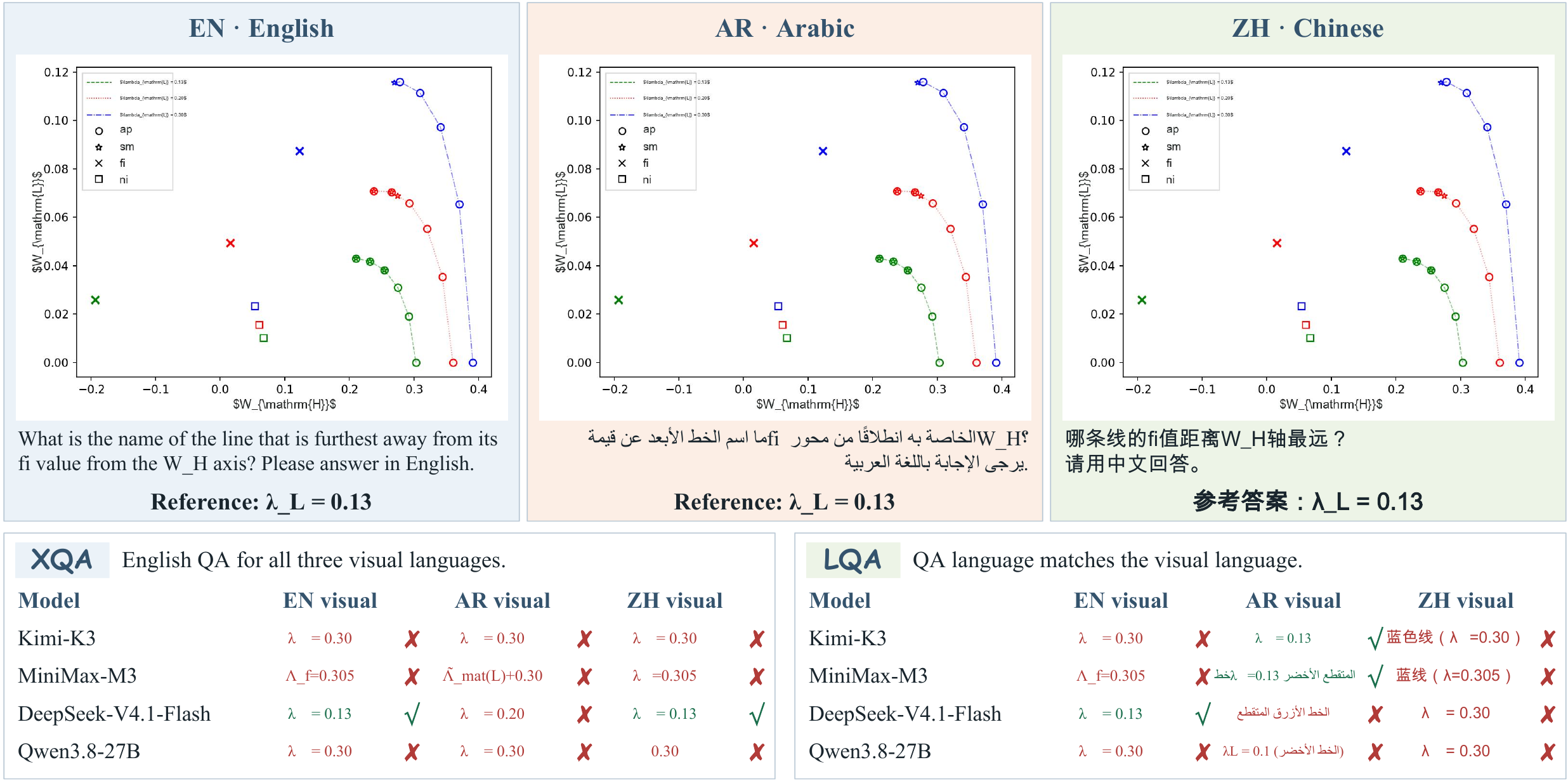}
  \caption{XQA and LQA examples using English, Arabic, and Chinese
  versions of the same chart.}
  \label{fig:xqa-lqa-case-study}
\end{figure}

Figure~\ref{fig:xqa-lqa-case-study} illustrates a model-dependent interaction between visual and QA languages.

 MiniMax-M3 answers incorrectly for all three visuals under English QA, but correctly under Arabic LQA. DeepSeek-V4.1-Flash answers correctly for the English and Chinese visuals under English QA, but incorrectly for the Arabic visual. The reference is $\lambda_L=0.13$. These contrasting responses illustrate model-dependent sensitivity to visual and QA languages. Even when ground-truth answers remain invariant (e.g., English text or numeric values), variations in the query language can still substantially sway MLLM predictions.

\section{Conclusion and Future Work}

The language of a chart or table should not determine whether a model can recover its facts. Our evaluation shows that this expectation remains unmet: strong average accuracy can coexist with substantial disparities across languages, and keeping questions and answers in English does not eliminate the performance loss associated with non-english visuals. Multilingual visual understanding therefore requires more than the ability to answer questions in several languages; it requires reliable access to the same information across its language variants.

We introduced \benchmark{} to make this distinction measurable. Its parallel charts, tables, and questions support paired comparisons that reveal differences hidden by aggregate scores. The central lesson is that accuracy and consistency capture distinct aspects of model capability: answering an instance correctly in one language does not establish that the model can interpret its counterparts in other languages. Progress in multilingual visual understanding should therefore be assessed through both overall correctness and the reliability of that correctness across languages.

Future work should investigate how to turn multilingual capability into multilingual reliability. A key direction is to understand where failures arise, e.g.,~from reading visual text, grounding labels in the visual structure, or reasoning over the underlying information, and how these failures interact across languages. This understanding can guide training methods that preserve factual predictions across semantically equivalent visuals, rather than improving average accuracy alone. Broader evaluation across naturally occurring documents, diverse scripts, and localization conventions will also be essential to establish how benchmark gains transfer to real use. Ultimately, the goal is for advances in visual understanding to make information more accessible regardless of the language in which it is presented.

\section*{Limitations}

\benchmark{} evaluates charts and tables in 24 languages, with English and Chinese as the fixed question and answer languages in the cross-language setting. Results are specific to this language coverage and evaluation design. Multilingual data were checked using automated verification and human review. The two reviewers were native Chinese speakers proficient in English; review of other languages used source evidence and translation tools rather than native-speaker validation. Residual translation or rendering errors may affect comparisons across languages.

\section*{Ethics and Broader Impact}
Reliable access to information in charts and tables matters across languages, particularly in education, research, and public communication. \benchmark{} supports this goal by making language-associated differences in model performance visible and providing a resource for evaluating more consistent multilingual visual understanding. The benchmark derives from existing chart and table QA datasets. Applicable source licenses and usage conditions continue to govern the derived materials, while our code is released under the MIT license. The license can be seen in Appendix~\ref{ap:lic}. Translation and reconstruction are intended to preserve the source information. Results characterize model performance within the benchmark's scope, and they should not be interpreted as assessments of languages or their speakers. Performance differences may reflect model capabilities as well as translation and rendering quality. Benchmark scores alone do not establish readiness for consequential applications, which require validation on relevant languages, documents, and use cases.

Regarding the use of AI and commercial APIs, beyond the automated components inherent to our dataset construction framework, AI assistance during manuscript preparation was strictly confined to language translation and polishing, with no involvement in research ideation. Furthermore, our utilization of proprietary commercial APIs to evaluate baseline models on \benchmark{} was conducted solely for non-commercial, academic benchmarking purposes.

\clearpage
\phantomsection\addcontentsline{toc}{section}{References}
\bibliography{custom}

@inproceedings{polychartqa,
    title = "{POLYCHARTQA}: Benchmarking Large Vision-Language Models with Multilingual Chart Question Answering",
    author = "Xu, Yichen  and
      Chen, Liangyu  and
      Zhang, Liang  and
      Yue, Zihao  and
      Ma, Jianzhe  and
      Wang, Wenxuan  and
      Jin, Qin",
    editor = "Liakata, Maria  and
      Moreira, Viviane P.  and
      Zhang, Jiajun  and
      Jurgens, David",
    booktitle = "Proceedings of the 64th Annual Meeting of the {A}ssociation for {C}omputational {L}inguistics (Volume 1: Long Papers)",
    month = jul,
    year = "2026",
    address = "San Diego, California, United States",
    publisher = "Association for Computational Linguistics",
    url = "https://aclanthology.org/2026.acl-long.2043/",
    doi = "10.18653/v1/2026.acl-long.2043",
    pages = "44154--44186",
    ISBN = "979-8-89176-390-6"
}

@article{chartr1,
  title={Chart-r1: Chain-of-thought supervision and reinforcement for advanced chart reasoner},
  author={Chen, Lei and Zhao, Xuanle and Zeng, Zhixiong and Huang, Jing and Zhong, Yufeng and Ma, Lin},
  journal={arXiv preprint arXiv:2507.15509},
  year={2025}
}

@inproceedings{geigle-etal-2025-centurio,
    title = "Centurio: On Drivers of Multilingual Ability of Large Vision-Language Model",
    author = "Geigle, Gregor  and
      Schneider, Florian  and
      Holtermann, Carolin  and
      Biemann, Chris  and
      Timofte, Radu  and
      Lauscher, Anne  and
      Glava{\v{s}}, Goran",
    editor = "Che, Wanxiang  and
      Nabende, Joyce  and
      Shutova, Ekaterina  and
      Pilehvar, Mohammad Taher",
    booktitle = "Proceedings of the 63rd Annual Meeting of the Association for Computational Linguistics (Volume 1: Long Papers)",
    month = jul,
    year = "2025",
    address = "Vienna, Austria",
    publisher = "Association for Computational Linguistics",
    url = "https://aclanthology.org/2025.acl-long.143/",
    doi = "10.18653/v1/2025.acl-long.143",
    pages = "2831--2881",
    ISBN = "979-8-89176-251-0"
}

@inproceedings{tang-etal-2025-mtvqa,
    title = "{MTVQA}: Benchmarking Multilingual Text-Centric Visual Question Answering",
    author = "Tang, Jingqun  and
      Liu, Qi  and
      Ye, Yongjie  and
      Lu, Jinghui  and
      Wei, Shu  and
      Wang, An-Lan  and
      Lin, Chunhui  and
      Feng, Hao  and
      Zhao, Zhen  and
      Wang, Yanjie  and
      Liu, Yuliang  and
      Liu, Hao  and
      Bai, Xiang  and
      Huang, Can",
    editor = "Che, Wanxiang  and
      Nabende, Joyce  and
      Shutova, Ekaterina  and
      Pilehvar, Mohammad Taher",
    booktitle = "Findings of the Association for Computational Linguistics: ACL 2025",
    month = jul,
    year = "2025",
    address = "Vienna, Austria",
    publisher = "Association for Computational Linguistics",
    url = "https://aclanthology.org/2025.findings-acl.404/",
    doi = "10.18653/v1/2025.findings-acl.404",
    pages = "7748--7763",
    ISBN = "979-8-89176-256-5"
}

@inproceedings{xing2025mmtu,
  author       = {Junjie Xing and
                  Yeye He and
                  Mengyu Zhou and
                  Haoyu Dong and
                  Shi Han and
                  Lingjiao Chen and
                  Dongmei Zhang and
                  Surajit Chaudhuri and
                  H. V. Jagadish},
  title        = {{MMTU:} {A} Massive Multi-Task Table Understanding and Reasoning Benchmark},
  booktitle    = {Advances in Neural Information Processing Systems 38: Annual Conference
                  on Neural Information Processing Systems 2025, NeurIPS 2025, San Diego,
                  CA, USA, December 2-7, 2025 / Mexico City, Mexico, November 30 - December
                  5, 2025},
  year         = {2025},
  url          = {http://papers.nips.cc/paper\_files/paper/2025/hash/63b3985ad1dade84a34d581590e5f608-Abstract-Datasets\_and\_Benchmarks\_Track.html}
}

@inproceedings{masry-etal-2025-chartqapro,
  author       = {Ahmed Masry and
                  Mohammed Saidul Islam and
                  Mahir Ahmed and
                  Aayush Bajaj and
                  Firoz Kabir and
                  Aaryaman Kartha and
                  Md. Tahmid Rahman Laskar and
                  Mizanur Rahman and
                  Shadikur Rahman and
                  Mehrad Shahmohammadi and
                  Megh Thakkar and
                  Md. Rizwan Parvez and
                  Enamul Hoque and
                  Shafiq Joty},
  title        = {ChartQAPro: {A} More Diverse and Challenging Benchmark for Chart Question Answering},
  booktitle    = {Findings of the Association for Computational Linguistics, {ACL} 2025,
                  Vienna, Austria, July 27 - August 1, 2025},
  series       = {Findings of {ACL}},
  volume       = {{ACL} 2025},
  pages        = {19123--19151},
  publisher    = {Association for Computational Linguistics},
  year         = {2025},
  url          = {https://doi.org/10.18653/v1/2025.findings-acl.978},
  doi          = {10.18653/V1/2025.FINDINGS-ACL.978},
}

@inproceedings{masry-etal-2022-chartqa,
  author       = {Ahmed Masry and
                  Do Xuan Long and
                  Jia Qing Tan and
                  Shafiq R. Joty and
                  Enamul Hoque},
  title        = {ChartQA: {A} Benchmark for Question Answering about Charts with Visual
                  and Logical Reasoning},
  booktitle    = {Findings of the Association for Computational Linguistics: {ACL} 2022,
                  Dublin, Ireland, May 22-27, 2022},
  series       = {Findings of {ACL}},
  volume       = {{ACL} 2022},
  pages        = {2263--2279},
  publisher    = {Association for Computational Linguistics},
  year         = {2022},
  url          = {https://doi.org/10.18653/v1/2022.findings-acl.177},
  doi          = {10.18653/V1/2022.FINDINGS-ACL.177},
}

@inproceedings{charxiv,
  author       = {Zirui Wang and
                  Mengzhou Xia and
                  Luxi He and
                  Howard Chen and
                  Yitao Liu and
                  Richard Zhu and
                  Kaiqu Liang and
                  Xindi Wu and
                  Haotian Liu and
                  Sadhika Malladi and
                  Alexis Chevalier and
                  Sanjeev Arora and
                  Danqi Chen},
  title        = {CharXiv: Charting Gaps in Realistic Chart Understanding in Multimodal
                  LLMs},
  booktitle    = {Advances in Neural Information Processing Systems 37: Annual Conference
                  on Neural Information Processing Systems 2024, NeurIPS 2024, Vancouver,
                  BC, Canada, December 10 - 15, 2024},
  year         = {2024},
  url          = {http://papers.nips.cc/paper\_files/paper/2024/hash/cdf6f8e9fd9aeaf79b6024caec24f15b-Abstract-Datasets\_and\_Benchmarks\_Track.html},
  bibsource    = {dblp computer science bibliography, https://dblp.org}
}

@inproceedings{chartmuseum,
  author       = {Liyan Tang and
                  Grace Kim and
                  Xinyu Zhao and
                  Thom Lake and
                  Wenxuan Ding and
                  Fangcong Yin and
                  Prasann Singhal and
                  Manya Wadhwa and
                  Zeyu Leo Liu and
                  Zayne Sprague and
                  Ramya Namuduri and
                  Bodun Hu and
                  Juan Diego Rodriguez and
                  Puyuan Peng and
                  Greg Durrett},
  title        = {ChartMuseum: Testing Visual Reasoning Capabilities of Large Vision-Language
                  Models},
  booktitle    = {Advances in Neural Information Processing Systems 38: Annual Conference
                  on Neural Information Processing Systems 2025, NeurIPS 2025, San Diego,
                  CA, USA, December 2-7, 2025 / Mexico City, Mexico, November 30 - December
                  5, 2025},
  year         = {2025},
  url          = {http://papers.nips.cc/paper\_files/paper/2025/hash/ca20efa9cf3703186d91424cf4876f8b-Abstract-Datasets\_and\_Benchmarks\_Track.html},
  bibsource    = {dblp computer science bibliography, https://dblp.org}
}

@article{kim2024tablevqa,
  title={Tablevqa-bench: A visual question answering benchmark on multiple table domains},
  author={Kim, Yoonsik and Yim, Moonbin and Song, Ka Yeon},
  journal={arXiv preprint arXiv:2404.19205},
  year={2024}
}

@inproceedings{xie-etal-2026-finchain,
    title = "{F}in{C}hain: A Symbolic Benchmark for Verifiable Chain-of-Thought Financial Reasoning",
    author = "Xie, Zhuohan  and
      Orel, Daniil  and
      Thareja, Rushil  and
      Sahnan, Dhruv  and
      Madmoun, Hachem  and
      Zhang, Fan  and
      Banerjee, Debopriyo  and
      Georgiev, Georgi Nenkov  and
      Peng, Xueqing  and
      Qian, Lingfei  and
      Huang, Jimin  and
      Su, Jinyan  and
      Singh, Aaryamonvikram  and
      Xing, Rui  and
      Elbadry, Rania  and
      Xu, Chen  and
      Li, Haonan  and
      Koto, Fajri  and
      Koychev, Ivan  and
      Chakraborty, Tanmoy  and
      Wang, Yuxia  and
      Lahlou, Salem  and
      Stoyanov, Veselin  and
      Ananiadou, Sophia  and
      Nakov, Preslav",
    editor = "Liakata, Maria  and
      Moreira, Viviane P.  and
      Zhang, Jiajun  and
      Jurgens, David",
    booktitle = "Proceedings of the 64th Annual Meeting of the {A}ssociation for {C}omputational {L}inguistics (Volume 1: Long Papers)",
    month = jul,
    year = "2026",
    address = "San Diego, California, United States",
    publisher = "Association for Computational Linguistics",
    url = "https://aclanthology.org/2026.acl-long.662/",
    doi = "10.18653/v1/2026.acl-long.662",
    pages = "14529--14553",
    ISBN = "979-8-89176-390-6"
}

@inproceedings{xtvqa,
  title={Cross-lingual text-rich visual comprehension: An information theory perspective},
  author={Yu, Xinmiao and Feng, Xiaocheng and Li, Yun and Liao, Minghui and Yu, Ya-Qi and Feng, Xiachong and Zhong, Weihong and Chen, Ruihan and Hu, Mengkang and Wu, Jihao and others},
  booktitle={Proceedings of the AAAI Conference on Artificial Intelligence},
  volume={39},
  number={9},
  pages={9680--9688},
  year={2025}
}

@article{jiang2026think,
  title={Think with Structured Grounding: Perceptual Reinforcement Learning for Chart and Visual-Tabular Understanding},
  author={Jiang, Changjiang and Zhao, Qiannian and Xin, Lei and Xie, Jinxiang and Nakov, Preslav and Xie, Zhuohan},
  journal={arXiv preprint arXiv:2608.22429},
  year={2026}
}

@inproceedings{zhou2026incentivizing,
  title={Incentivizing parametric knowledge via reinforcement learning with verifiable rewards for cross-cultural entity translation},
  author={Zhou, Jiang and Zhao, Xiaohu and Wu, Xinwei and Dong, Tianyu and Wang, Hao and Liu, Yangyang and Liu, Heng and Xu, Linlong and Wang, Longyue and Luo, Weihua and others},
  booktitle={Proceedings of the 64th Annual Meeting of the Association for Computational Linguistics (Volume 1: Long Papers)},
  pages={5616--5638},
  year={2026}
}

@article{visual-tableqa,
  title={Visual-tableqa: Open-domain benchmark for reasoning over table images},
  author={Lompo, Boammani Aser and Haraoui, Marc},
  journal={arXiv preprint arXiv:2509.07966},
  year={2025}
}

@misc{gpt6,
  author       = {{OpenAI}},
  title        = {{GPT-6 Astra: A New Generation of Intelligence}},
  year         = {2026},
  howpublished = {\url{https://openai.com/index/gpt-6-astra/}},
  note         = {Accessed: 2026-10-02}
}

@misc{gpt55,
  author       = {{OpenAI}},
  title        = {{Introducing GPT‑5.5}},
  year         = {2026},
  howpublished = {\url{https://openai.com/index/introducing-gpt-5-5/}},
  note         = {Accessed: 2026-10-02}
}

@misc{gpt56,
  author       = {{OpenAI}},
  title        = {{GPT-5.6: Frontier Intelligence That Scales with Your Ambition}},
  year         = {2026},
  howpublished = {\url{https://openai.com/index/gpt-5-6/}},
  note         = {Accessed: 2026-10-02}
}

@misc{gemini38,
  author       = {{Google DeepMind}},
  title        = {{Gemini 3.8 Flash}},
  year         = {2026},
  month        = sep,
  howpublished = {\url{https://deepmind.google/models/model-cards/gemini-3-8-flash/}},
  note         = {Model Card. Accessed: 2026-10-02}
}

@article{team2026kimi,
  title={Kimi k3: Open frontier intelligence},
  author={Team, Kimi and Bai, Tongtong and Bai, Yifan and Bao, Yiping and Cai, Jianfeng and Cai, Xinyuan and Cao, Peizhou and Cao, Yuxuan and Chai, Ziwei and Charles, Y and others},
  journal={arXiv preprint arXiv:2607.24653},
  year={2026}
}

@misc{minimaxm3,
  author       = {{MiniMax}},
  title        = {{MiniMax M3}: Frontier Coding, 1M Context,
                  Native Multimodality---All in One Model},
  year         = {2026},
  howpublished = {\url{https://www.minimax.io/blog/minimax-m3/}},
  note         = {Accessed: 2026-10-02}
}

@article{xu2026deepseek,
  title={Deepseek-v4.1-flash: Pushing the limits of kv cache compression},
  author={Xu, Anyi and Li, B and Lin, Bangcai and Xue, Bing and Xian, BingCheng and Xu, Bingzheng and Wu, Bochao and Zhang, Bowei and Deng, Boyi and Yu, CC and others},
  journal={arXiv preprint arXiv:2609.19969},
  year={2026}
}

@misc{qwen38,
    title = {Qwen3.8-Max: A New Bar for Coding and Cowork},
    url = {https://qwen.ai/blog?id=qwen3.8},
    author = {{Qwen Team}},
    month = {August},
    year = {2026}
}

@misc{qwen35,
    title = {Qwen3.5: Towards Native Multimodal Agents},
    url = {https://qwen.ai/blog?id=qwen3.5},
    author = {Qwen Team},
    month = {February},
    year = {2026}
}

@article{yuan2026vision,
  title={Vision-opd: Learning to see fine details for multimodal llms via on-policy self-distillation},
  author={Yuan, Qianhao and Lou, Jie and Yu, Xing and Lin, Hongyu and Sun, Le and Han, Xianpei and Lu, Yaojie},
  journal={arXiv preprint arXiv:2605.18740},
  year={2026}
}

@article{wei2026zooming,
  title={Zooming without Zooming: Region-to-Image Distillation for Fine-Grained Multimodal Perception},
  author={Wei, Lai and He, Liangbo and Lan, Jun and Dong, Lingzhong and Cai, Yutong and Li, Siyuan and Zhu, Huijia and Wang, Weiqiang and Kong, Linghe and Wang, Yue and others},
  journal={arXiv preprint arXiv:2602.11858},
  year={2026}
}

@inproceedings{Chartsketcher,
  title={Chartsketcher: Reasoning with multimodal feedback and reflection for chart understanding},
  author={Huang, Muye and Zhang, Lingling and Ma, Jie and Lai, Han and Xu, Fangzhi and Li, Yifei and Wu, Wenjun and Wu, Yaqiang and Liu, Jun},
  booktitle={Advances in Neural Information Processing Systems},
  volume={38},
  year={2025}
}

@inproceedings{chartverse,
  title={Chartverse: Scaling chart reasoning via reliable programmatic synthesis from scratch},
  author={Liu, Zheng and Lin, Honglin and Wang, Xiaoyang and Gao, Xin and Li, Yu and Cai, Mengzhang and Zhu, Yun and Zhong, Zhanping and Pei, Qizhi and Pan, Zhuoshi and others},
  booktitle={Proceedings of the 64th Annual Meeting of the Association for Computational Linguistics (Volume 1: Long Papers)},
  pages={7551--7577},
  year={2026}
}

@article{team2026polyocr,
  title={PolyOCR-Venus: Unified OCR Foundation Models for Text-Centric Visual Intelligence},
  author={Team, GuangJian and Huang, Kaili and Zhang, Yongshuo and Fu, Bingtao and Jiang, Changjiang and Qu, Chenfan and Zhang, Chenfeng and Cui, Fangming and Zhang, Gaoyang and Xie, Jiangwei and others},
  journal={arXiv preprint arXiv:2609.37712},
  year={2026}
}

@inproceedings{jiang2025tabdsr,
  author = {Jiang, Changjiang and Yu, Fengchang and Chen, Haihua and Lu, Wei and Zeng, Jin},
  title = {TABDSR: Decompose, Sanitize, and Reason for Complex Numerical Reasoning in Tabular Data},
  booktitle = {Findings of the Association for Computational Linguistics: EMNLP 2025},
  year = {2025},
  pages = {3172--3196},
  doi = {10.18653/v1/2025.findings-emnlp.169},
  isbn = {979-8-89176-335-7},
}

@inproceedings{fakehr1,
  author={Jiang, Changjiang and Sha, Xinkuan and Yu, Fengchang and Liu, Jingjing and Liu, Jian and Fang, Mingqi and Zhang, Chenfeng and Lu, Wei},
  booktitle={ICASSP 2026 - 2026 IEEE International Conference on Acoustics, Speech and Signal Processing (ICASSP)}, 
  title={Fake-HR1: Rethinking Reasoning of Vision Language Model for Synthetic Image Detection}, 
  year={2026},
  pages={10482--10486},
  doi={10.1109/ICASSP55912.2026.11462736}
}

@inproceedings{jiang2025ivy,
	author = {Jiang, Changjiang and Dong, Wenhui and Zhang, Zhonghao and Yu, Fengchang and Peng, Wei and Yuan, Xinbin and Bi, Yifei and Zhao, Ming and Zhou, Zian and Si, Chenyang and Shan, Caifeng},
	title = {Ivy-Fake: A Unified Explainable Framework and Benchmark for Image and Video AIGC Detection},
	year = {2026},
	isbn = {9798400726170},
	publisher = {Association for Computing Machinery},
	url = {https://doi.org/10.1145/3805622.3810615},
	doi = {10.1145/3805622.3810615},
	booktitle = {Proceedings of the 2026 International Conference on Multimedia Retrieval},
	pages = {2438–2447},
	numpages = {10},
	series = {ICMR '26}
}

@article{DeFakerOne,
  title={Venus-DeFakerOne: Unified Fake Image Detection \& Localization},
  author={Team, GuangJian},
  journal={arXiv preprint arXiv:2605.14091},
  year={2026}
}

@inproceedings{cite_segui,
  author = {Yuan, Xinbin and Zhang, Jian and Li, Kaixin and Cai, Zhuoxuan and Yao, Lujian and Chen, Jie and Wang, Enguang and Hou, Qibin and Chen, Jinwei and Jiang, Peng-Tao and Li, Bo},
  booktitle = {Advances in Neural Information Processing Systems},
  pages = {127658--127679},
  publisher = {Curran Associates, Inc.},
  title = {SE-GUI: Enhancing Visual Grounding for GUI Agents via Self-Evolutionary Reinforcement Learning},
  url = {https://proceedings.neurips.cc/paper_files/paper/2025/file/b95c7e24501f5d1dddbc5e8526cda7ae-Paper-Conference.pdf},
  volume = {38},
  year = {2025},
}

@article{pal,
  title={Can We Model the Artifacts Explicitly? Disentangle Artifacts via Pairwise Edit Relations for Image Manipulation Localization},
  author={Zhu, Xuekang and Feng, Kaiwen and Wang, Ruifeng and Wang, Xiwen and Ma, Xiaochen and Du, Bo and Jiang, Changjiang and Qu, Chenfan and Ye, Songyu and Du, Xia and others},
  journal={arXiv preprint arXiv:2610.07916},
  year={2026}
}

@article{tabverse,
  title={TABVERSE: Benchmarking Cross-Format Table Understanding in LLMs and VLMs},
  author={Ahsan, Momina and Ahmad, Sarfraz and Hee, Ming Shan and Lee, Roy Ka-Wei and Nakov, Preslav},
  journal={arXiv preprint arXiv:2606.09578},
  year={2026}
}

@inproceedings{zhu2025mesoscopic,
  title={Mesoscopic insights: orchestrating multi-scale \& hybrid architecture for image manipulation localization},
  author={Zhu, Xuekang and Ma, Xiaochen and Su, Lei and Jiang, Zhuohang and Du, Bo and Wang, Xiwen and Lei, Zeyu and Feng, Wentao and Pun, Chi-Man and Zhou, Ji-Zhe},
  booktitle={Proceedings of the AAAI conference on artificial intelligence},
  volume={39},
  number={10},
  pages={11022--11030},
  year={2025}
}

@article{ma2024imdl,
  title={Imdl-benco: A comprehensive benchmark and codebase for image manipulation detection \& localization},
  author={Ma, Xiaochen and Zhu, Xuekang and Su, Lei and Du, Bo and Jiang, Zhuohang and Tong, Bingkui and Lei, Zeyu and Yang, Xinyu and Pun, Chi-Man and Lv, Jiancheng and others},
  journal={Advances in Neural Information Processing Systems},
  volume={37},
  pages={134591--134613},
  year={2024}
}

@article{du2026forensichub,
  title={Forensichub: A unified benchmark \& codebase for all-domain fake image detection and localization},
  author={Du, Bo and Zhu, Xuekang and Ma, Xiaochen and Qu, Chenfan and Feng, Kaiwen and Yang, Zhe and Pun, Chi-Man and Zhou, Ji-Zhe and others},
  journal={Advances in neural information processing systems},
  volume={38},
  year={2026}
}

@inproceedings{zhu2026revisiting,
  title={Revisiting image manipulation localization under realistic manipulation scenarios},
  author={Zhu, Xuekang and Zhou, Ji-Zhe and Feng, Kaiwen and Qu, Chenfan and Wang, Xiwen and Wang, Yunfei and Zhou, Liting and Liu, Jian},
  booktitle={Proceedings of the IEEE/CVF Conference on Computer Vision and Pattern Recognition},
  pages={7198--7207},
  year={2026}
}

@inproceedings{lin2025audio,
  title={Audio Does Matter: Importance-Aware Multi-Granularity Fusion for Video Moment Retrieval},
  author={Lin, Junan and Liu, Daizong and Chen, Xianke and Qu, Xiaoye and Yang, Xun and Zhu, Jixiang and Zhang, Sanyuan and Dong, Jianfeng},
  booktitle={Proceedings of the 33rd ACM International Conference on Multimedia},
  pages={6027--6036},
  year={2025}
}

@inproceedings{qu2024towards,
  title={Towards modern image manipulation localization: A large-scale dataset and novel methods},
  author={Qu, Chenfan and Zhong, Yiwu and Liu, Chongyu and Xu, Guitao and Peng, Dezhi and Guo, Fengjun and Jin, Lianwen},
  booktitle={2024 IEEE/CVF Conference on Computer Vision and Pattern Recognition (CVPR)},
  pages={10781--10790},
  year={2024},
  organization={IEEE}
}

@inproceedings{qu2025revisiting,
  title={Revisiting tampered scene text detection in the era of generative AI},
  author={Qu, Chenfan and Zhong, Yiwu and Guo, Fengjun and Jin, Lianwen},
  booktitle={Proceedings of the AAAI Conference on Artificial Intelligence},
  volume={39},
  number={1},
  pages={694--702},
  year={2025}
}

@inproceedings{qu2026omni,
  title={Omni-IML: Towards Unified Interpretable Image Manipulation Localization},
  author={Qu, Chenfan and Zhong, Yiwu and Guo, Fengjun and Jin, Lianwen},
  booktitle={The Fourteenth International Conference on Learning Representations},
  year={2026}
}

@inproceedings{qu2026textshield,
  title={Textshield-r1: Reinforced reasoning for tampered text detection},
  author={Qu, Chenfan and Zhong, Yiwu and Liu, Jian and Zhu, Xuekang and Yu, Bohan and Jin, Lianwen},
  booktitle={Proceedings of the AAAI Conference on Artificial Intelligence},
  volume={40},
  number={10},
  pages={8621--8629},
  year={2026}
}

@inproceedings{qu2026detect,
  title={Detect Any AI-Counterfeited Text Image},
  author={Qu, Chenfan and Zhong, Yiwu and Zhu, Xuekang and Li, Junchi and Jiang, Changjiang and Jin, Lianwen and others},
  booktitle={Proceedings of the IEEE/CVF Conference on Computer Vision and Pattern Recognition},
  pages={35437--35450},
  year={2026}
}

@inproceedings{masry2024chartinstruct,
  title={Chartinstruct: Instruction tuning for chart comprehension and reasoning},
  author={Masry, Ahmed and Shahmohammadi, Mehrad and Parvez, Md Rizwan and Hoque, Enamul and Joty, Shafiq},
  booktitle={Findings of the Association for Computational Linguistics: ACL 2024},
  pages={10387--10409},
  year={2024}
}

@article{masry2025bigcharts,
  title={Bigcharts-r1: Enhanced chart reasoning with visual reinforcement finetuning},
  author={Masry, Ahmed and Puri, Abhay and Hashemi, Masoud and Rodriguez, Juan A and Thakkar, Megh and Mahajan, Khyati and Yadav, Vikas and Madhusudhan, Sathwik Tejaswi and Pich{\'e}, Alexandre and Bahdanau, Dzmitry and others},
  journal={arXiv preprint arXiv:2508.09804},
  year={2025}
}

@inproceedings{shi2025judging,
  title={Judging the judges: A systematic study of position bias in llm-as-a-judge},
  author={Shi, Lin and Ma, Chiyu and Liang, Wenhua and Diao, Xingjian and Ma, Weicheng and Vosoughi, Soroush},
  booktitle={Proceedings of the 14th International Joint Conference on Natural Language Processing and the 4th Conference of the Asia-Pacific Chapter of the Association for Computational Linguistics},
  pages={292--314},
  year={2025}
}

@inproceedings{tatqa,
  title={TAT-QA: A question answering benchmark on a hybrid of tabular and textual content in finance},
  author={Zhu, Fengbin and Lei, Wenqiang and Huang, Youcheng and Wang, Chao and Zhang, Shuo and Lv, Jiancheng and Feng, Fuli and Chua, Tat-Seng},
  booktitle={Proceedings of the 59th annual meeting of the Association for Computational Linguistics and the 11th international joint conference on natural language processing (volume 1: long papers)},
  pages={3277--3287},
  year={2021}
}

@article{fetaqa,
    title = "{F}e{T}a{QA}: Free-form Table Question Answering",
    author = "Nan, Linyong  and
      Hsieh, Chiachun  and
      Mao, Ziming  and
      Lin, Xi Victoria  and
      Verma, Neha  and
      Zhang, Rui  and
      Kry{\'s}ci{\'n}ski, Wojciech  and
      Schoelkopf, Hailey  and
      Kong, Riley  and
      Tang, Xiangru  and
      Mutuma, Mutethia  and
      Rosand, Ben  and
      Trindade, Isabel  and
      Bandaru, Renusree  and
      Cunningham, Jacob  and
      Xiong, Caiming  and
      Radev, Dragomir",
    editor = "Roark, Brian  and
      Nenkova, Ani",
    journal = "Transactions of the Association for Computational Linguistics",
    volume = "10",
    year = "2022",
    address = "Cambridge, MA",
    publisher = "MIT Press",
    url = "https://aclanthology.org/2022.tacl-1.3/",
    doi = "10.1162/tacl_a_00446",
    pages = "35--49"
}

@inproceedings{he2025tablelora,
  title={TableLoRA: low-rank adaptation on table structure understanding for large language models},
  author={He, Xinyi and Liu, Yihao and Zhou, Mengyu and He, Yeye and Dong, Haoyu and Han, Shi and Yuan, Zejian and Zhang, Dongmei},
  booktitle={Proceedings of the 63rd Annual Meeting of the Association for Computational Linguistics (Volume 1: Long Papers)},
  pages={22376--22391},
  year={2025}
}

@article{xu2026modora,
  title={MoDora: Tree-Based Semi-Structured Document Analysis System},
  author={Xu, Bangrui and Yao, Qihang and Tang, Zirui and Zhou, Xuanhe and He, Yeye and Yu, Shihan and Xu, Qianqian and Wang, Bin and Li, Guoliang and He, Conghui and others},
  journal={Proceedings of the ACM on Management of Data},
  volume={4},
  number={3 (SIGMOD},
  pages={1--26},
  year={2026},
  publisher={ACM New York, NY, USA}
}

@inproceedings{xgqa,
  title={xGQA: Cross-lingual visual question answering},
  author={Pfeiffer, Jonas and Geigle, Gregor and Kamath, Aishwarya and Steitz, Jan-Martin O and Roth, Stefan and Vuli{\'c}, Ivan and Gurevych, Iryna},
  booktitle={Findings of the association for computational linguistics: ACL 2022},
  pages={2497--2511},
  year={2022}
}

@inproceedings{maxm,
  title={Maxm: Towards multilingual visual question answering},
  author={Changpinyo, Soravit and Xue, Linting and Yarom, Michal and Thapliyal, Ashish and Szpektor, Idan and Amelot, Julien and Chen, Xi and Soricut, Radu},
  booktitle={Findings of the Association for Computational Linguistics: EMNLP 2023},
  pages={2667--2682},
  year={2023}
}

@article{cvqa,
  title={Cvqa: Culturally-diverse multilingual visual question answering benchmark},
  author={Romero, David and Lyu, Chenyang and Wibowo, Haryo Akbarianto and Lynn, Teresa and Hamed, Injy and Kishore, Aditya Nanda and Mandal, Aishik and Dragonetti, Alina and Abzaliev, Artem and Tonja, Atnafu Lambebo and others},
  journal={arXiv preprint arXiv:2406.05967},
  year={2024}
}

@article{alam2025oasis,
  title={OASIS: A Multilingual and Multimodal Dataset for Culturally Grounded Spoken Visual QA},
  author={Alam, Firoj and Shahroor, Ali Ezzat and Hasan, Md Arid and Ali, Zien Sheikh and Bhatti, Hunzalah Hassan and Kmainasi, Mohamed Bayan and Chowdhury, Shammur Absar and Mousi, Basel and Dalvi, Fahim and Durrani, Nadir and others},
  journal={arXiv preprint arXiv:2510.06371},
  year={2025}
}

@article{shahroor2026memelens,
  title={MemeLens: Multilingual Multitask VLMs for Memes},
  author={Shahroor, Ali Ezzat and Kmainasi, Mohamed Bayan and Hasnat, Abul and Dimitrov, Dimitar and Martino, Giovanni Da San and Nakov, Preslav and Alam, Firoj},
  journal={arXiv preprint arXiv:2601.12539},
  year={2026}
}

@inproceedings{zhang2025multitat,
  title={MULTITAT: Benchmarking Multilingual Table-and-Text Question Answering.},
  author={Zhang, Xuanliang and Wang, Dingzirui and Xu, Keyan and Zhu, Qingfu and Che, Wanxiang},
  booktitle={EMNLP (Findings)},
  pages={626--647},
  year={2025}
}

@inproceedings{shu2026m3tqa,
  title={M3TQA: Massively Multilingual Multitask Table Question Answering},
  author={Shu, Daixin and Yang, Jian and Wu, Zhenhe and Wu, Xianjie and Cheng, Xianfu and Xiangyuan, Guan and Wang, Yanghai and Wu, Pengfei and Yang, Tingyang and Zhu, Hualei and others},
  booktitle={Findings of the Association for Computational Linguistics: ACL 2026},
  pages={22578--22602},
  year={2026}
}

@inproceedings{gautam2026indotabvqa,
  title={INDOTABVQA: A Benchmark for Cross-Lingual Table Understanding in Bahasa Indonesia Documents},
  author={Gautam, Somraj and Dravichi, Anathapindika and Harit, Gaurav},
  booktitle={Findings of the Association for Computational Linguistics: ACL 2026},
  pages={21969--21981},
  year={2026}
}

\clearpage\appendix
\section*{Appendices}
\markboth{MInfoVisQA}{Appendices}

\section{License}
\label{ap:lic}

The upstream dataset releases specify different licenses: ChartQAPro, Visual-TableQA, and MMTU list MIT; TableVQA-Bench lists CC BY 4.0; and the ChartQA release used in this work lists GPL 3.0. CharXiv licenses its questions under CC BY-SA 4.0, while copyright in its source charts remains with the original authors; its usage instructions restrict the dataset to evaluation. These upstream licenses and any applicable rights in underlying source materials remain relevant to derived examples. The MIT license for our software does not supersede these dataset-specific conditions.

\section{API Cost}
\label{ap:api_cost}

We estimate translation costs using Google Cloud Translation's
standard NMT pricing of US\$20 per million input characters,
counted separately for each target language. The English visual labels, questions, reference answers, and required context for the 128 instances contain approximately 87,300 characters. Translating this content into the other 23 languages entails approximately 2.01 million billable characters, costing about US\$40.17. This estimate assumes one translation pass without retries. Translation is budgeted separately from LLM-based reconstruction, QA normalization, query refinement, and quality checks. Image rendering is performed locally.

\begin{table}[h]
\centering
\resizebox{0.95\linewidth}{!}{%
\begin{tabular}{lrrrrrr}
\hline
& \multicolumn{3}{c}{Records}
& \multicolumn{3}{c}{Overall ACC (\%) by judge} \\
Answer model & Total & Direct match & Needs judge
& GPT-6-Astra & GPT-5.5 & Qwen3.8-27B \\
\hline
GPT-6 Astra
& 8,960 & 3,925 & 5,033
& 81.90 & 83.19 ($+1.29$) & 81.86 ($-0.03$) \\
Gemini-3.8-Flash
& 8,960 & 4,355 & 4,603
& 79.92 & 81.04 ($+1.12$) & 78.78 ($-1.14$) \\
GPT-5.6 Sol
& 8,960 & 3,410 & 5,550
& 78.29 & 79.78 ($+1.48$) & 77.03 ($-1.26$) \\
GPT-5.6 Luna
& 8,960 & 3,239 & 5,721
& 74.00 & 74.94 ($+0.95$) & 71.58 ($-2.41$) \\
\hline
\end{tabular}%
}
\caption{Scoring workload and sensitivity to the answer judge.}
\label{tab:judge_sensitivity}
\end{table}

\section{Reviewers' Background.}
\label{ap:rback}

Two annotators with computer science backgrounds reviewed the benchmark. Both are native Chinese speakers with English proficiency. They divided all instances between them for initial inspection, including automatically accepted and flagged instances. Automatically flagged cases and additional issues found during initial inspection were checked by the other annotator. Unflagged cases without identified issues received no second review: coverage was complete for initial inspection, not independent double annotation. Review beyond Chinese and English was tool-assisted, not native speaker validation.

\section{Sensitivity to the Answer Judge}
\label{ap:judge_ab}
We compare GPT-6 Astra, GPT-5.5, and Qwen3.8-27B as text-only answer-equivalence judges while holding the questions, effective references, candidate answers, and judging prompt fixed. Table~\ref{tab:judge_sensitivity} reports the scoring workload for each answer-generating model. Direct matches receive credit without a judge call; only completed, non-matching answers require semantic adjudication. Since these counts differ across models, we compare overall accuracy on all saved predictions, rather than accuracy on the judge-only subset. Specifically, $\mathrm{ACC}_{j}=100(M+C_j)/N$, where $N$ is the total number of predictions, $M$ counts direct matches, and $C_j$ counts non-matching answers accepted by judge $j$. Inference failures remain in $N$ and receive no credit. Identical judging inputs share a cached decision while retaining their original record weights.

\paragraph{Effect on benchmark results.} Non-matching predictions constitute a majority for every answer model, making judge choice relevant to the final scores. GPT-5.5 changes overall ACC by +0.95 to +1.48 percentage points relative to Astra; Qwen3.8-27B changes overall ACC by -2.41 to -0.03 percentage points relative to Astra. Both alternative judges preserve the ranking of these four answer models; this check does not establish ranking stability for all 25 models.
The paired differences quantify sensitivity to judge choice; independent human answer-equivalence judgments would be needed to establish which judge is more accurate. Most Astra decisions are historical, so temporal and within-model variation may also contribute.

\section{Additional Subgroup Diagnostics}
\label{app:subgroup-diagnostics}

\begin{figure}[tbp]
  \centering
  \includegraphics[width=\textwidth]{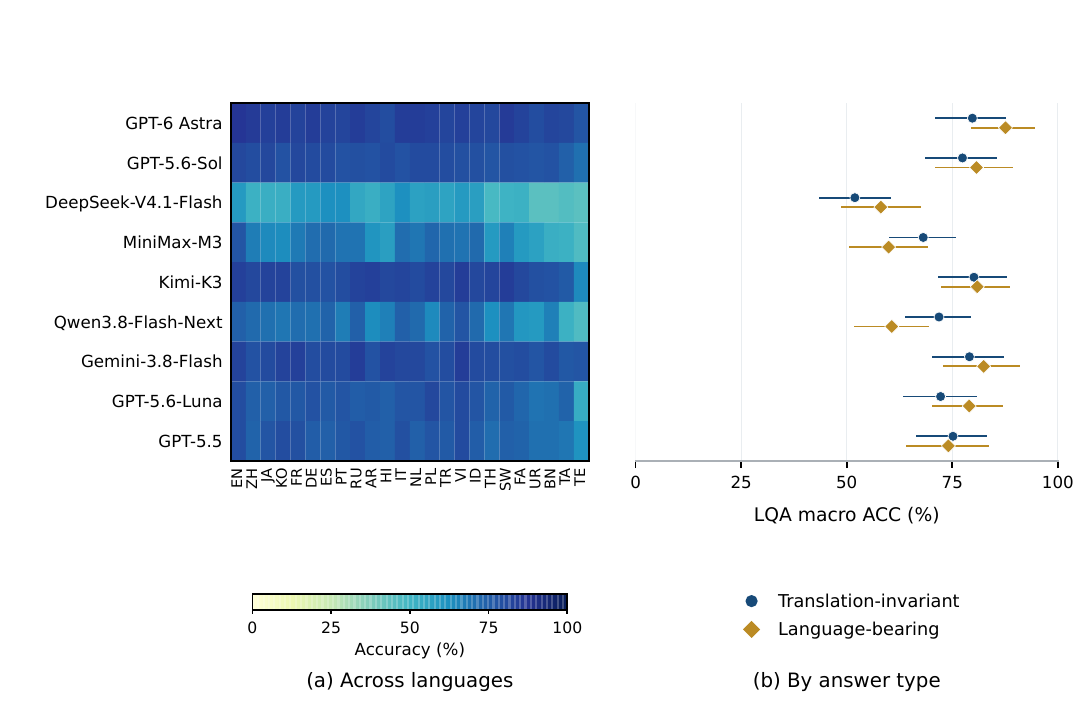}
  \caption{LQA diagnostics.
  (a) Accuracy across 24 languages.
  (b) Macro-averaged accuracy by answer type, with 95\%
  seed-level bootstrap confidence intervals.}
  \label{fig:app-lqa}
\end{figure}

\paragraph{Language-aligned diagnostics.}
Figure~\ref{fig:app-lqa} reports two complementary analyses of LQA for nine models.

\begin{figure}[tbp]
  \centering
  \includegraphics[width=\textwidth]{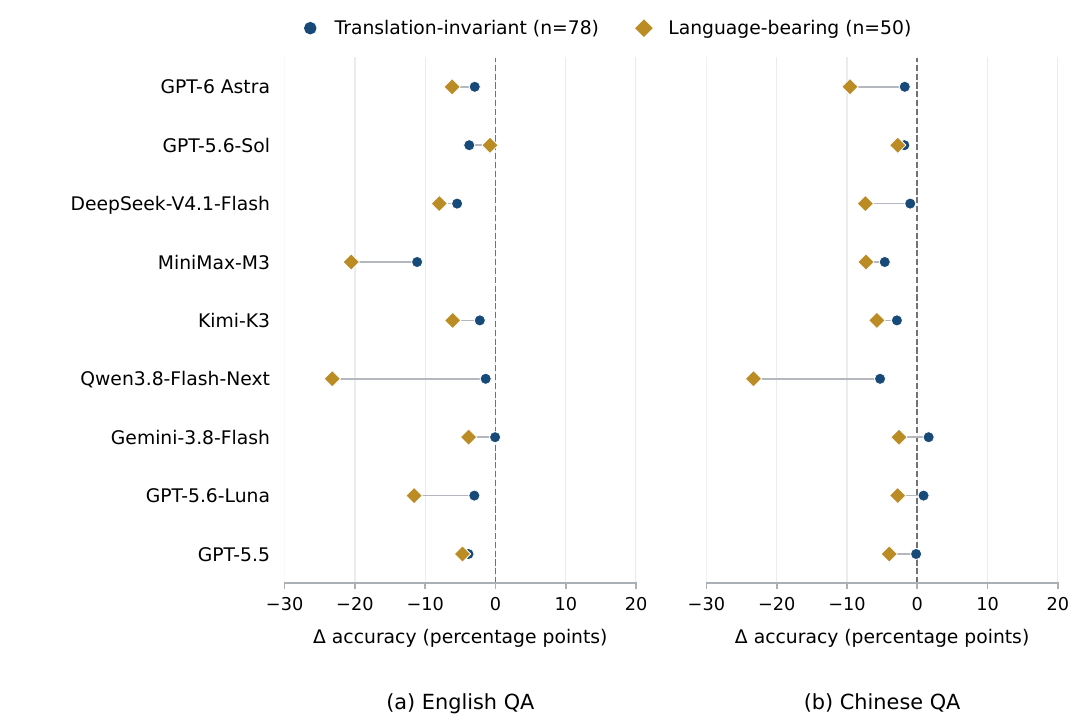}
  \caption{XQA sensitivity by answer type for English and Chinese QA.
  $\Delta$ACC is the mean accuracy over the other 23 visual
  languages minus accuracy with the same-language visual.
  Points show estimates only; gray lines connect the two
  answer subsets for each model.}
  \label{fig:app-xqa-subgroups}
\end{figure}

\begin{figure}[tbp]
  \centering
  \begin{minipage}[t]{0.49\textwidth}
    \centering
    \includegraphics[width=\linewidth]{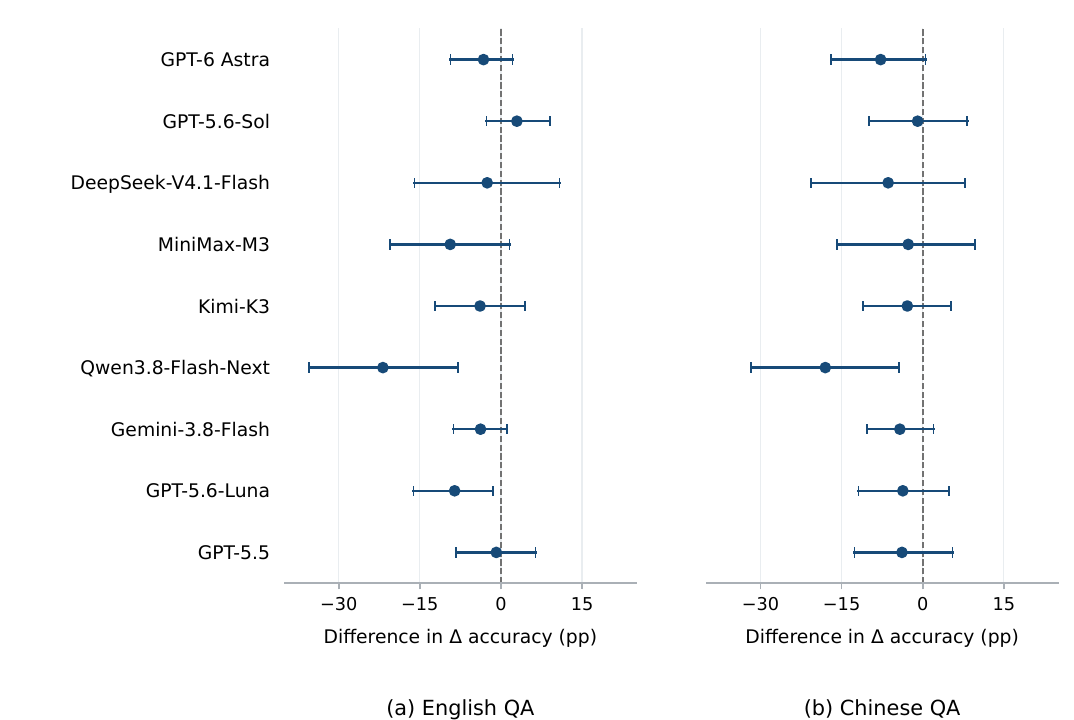}
    \par\smallskip
    {\small (a) Language-bearing $-$ Translation-invariant}
  \end{minipage}
  \hfill
  \begin{minipage}[t]{0.49\textwidth}
    \centering
    \includegraphics[width=\linewidth]{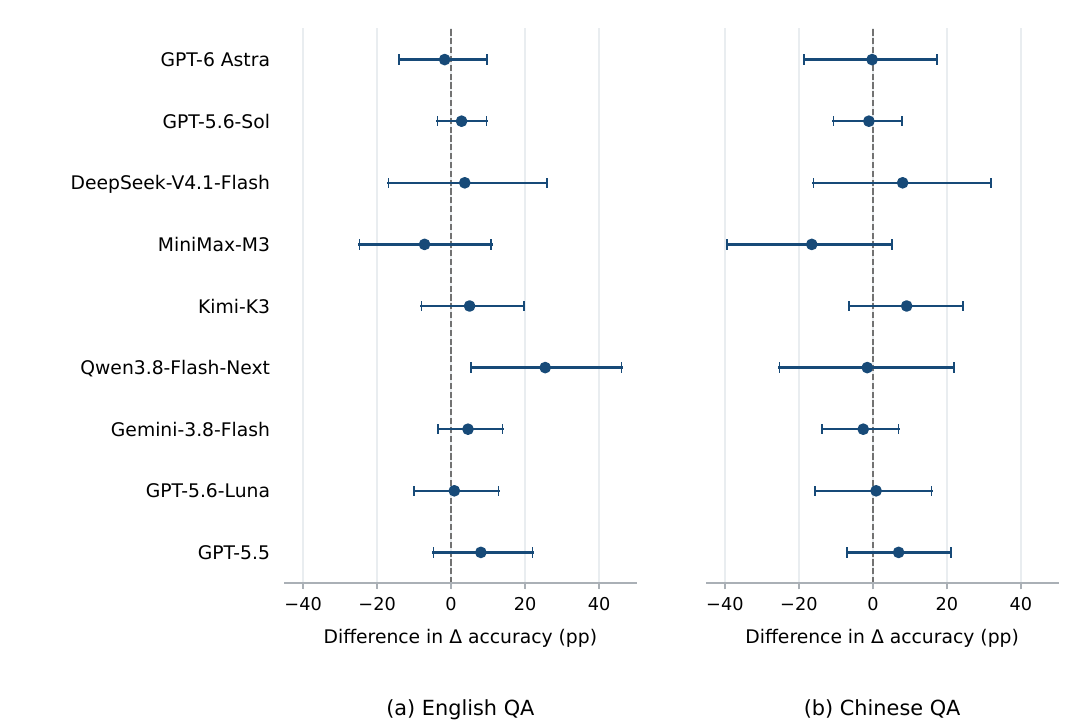}
    \par\smallskip
    {\small (b) Spanning $-$ Simple}
  \end{minipage}
  \caption{Direct contrasts in XQA $\Delta$ACC, in percentage points,
  by answer type and table structure. Intervals are 95\% bootstrap
  confidence intervals, obtained by independently resampling seeds
  within each subgroup while retaining all language variants.
  Negative values indicate a larger accuracy decrease in the
  first named subgroup.}
  \label{fig:app-subgroup-contrasts}
\end{figure}

Panel (a) shows accuracy separately for each of the 24 languages, with the visual, question, and reference answer in the same language. The heatmap exposes language-specific variation that a single mean score conceals. Because all three language components change together, these differences describe the aligned task rather than isolating visual-language sensitivity.

Panel (b) compares translation-invariant and language-bearing reference answers. The former do not require language-specific wording, whereas the latter contain translatable text. For each subset, accuracy is averaged over the 24 aligned languages; the intervals are 95\% seed-level bootstrap confidence intervals. The relative ordering of the two subsets varies across models, so the answer-type gap is not uniform. This is a descriptive subgroup comparison: the subsets contain different seeds and may differ in reasoning demands, so the gap cannot be attributed solely to answer translation. The XQA analyses below instead hold the QA language fixed while varying the visual language.

\paragraph{Cross-language diagnostics.}

Figure~\ref{fig:app-xqa-subgroups} demonstrates that
visual-language sensitivity extends beyond answer translation.
Even on Translation-invariant answers, GPT-6-Astra,
GPT-5.6-Sol, and MiniMax-M3 lose 2.95, 3.73, and 11.15
percentage points under English QA, respectively.
These changes show that keeping the answer unchanged does
not ensure stable understanding when the visual language
changes, supporting the need for our paired evaluation.

Language-bearing answers expose an additional challenge.
Figure~\ref{fig:app-subgroup-contrasts}(a) shows more negative
$\Delta$ACC for this subset for most model--QA-language comparisons. The gap reaches 21.82 and 18.01 percentage
points for Qwen3.8-Flash-Next under English and Chinese QA.
This pattern highlights the importance of evaluating both
language-invariant and language-specific answers when
assessing multilingual InfoVis understanding.

Structural complexity alone does not explain these patterns.
Figure~\ref{fig:app-subgroup-contrasts}(b) shows no consistent
additional degradation for spanning tables relative to simple
tables. For these nine models and sampled subgroups, the results support reporting visual-language sensitivity alongside answer type and table structure. They do not isolate causal effects of either subgroup attribute.

\section{Language and Table Coverage}
\label{app:contribution}

Our contribution emphasizes the breadth of language coverage and the treatment of tables within a shared chart-and-table evaluation. Code-based chart rendering and coordinated translation are established components of prior work~\citep{team2026polyocr,polychartqa}, as discussed in Related Work; we do not claim them as new in isolation. Relative to PolyChartQA's ten-language chart setting, \benchmark{} covers 24 languages and includes 41 table seeds alongside 87 chart seeds. The additional language variants broaden paired evaluation, but do not increase the number of independent seed questions.

\paragraph{Table-specific construction.}
Table translation must preserve the association between cell contents and row and column headers, including spanning cells. Our pipeline parses available HTML or Markdown and recovers cell structure from image-only sources before translating textual cells and rendering the result. Numeric content and the associations required by the question are retained as construction constraints. This table branch, together with synchronized QA and human inspection, is a central contribution beyond multilingual chart generation. It supports both descriptive and numerical table questions; examples appear in Figures~\ref{fig:example-table-descriptive} and~\ref{fig:example-table-numerical}.

\paragraph{Repairing table-containing visuals.}
Figures~\ref{fig:error_component} and~\ref{fig:error_multilingual} document the existing mixed-visual repair case: retaining the table alone omitted a radial diagram required to answer the question. Prompt revision restored the complete evidence, and the displayed English, Chinese, and Arabic versions preserve the recovered color--label association. This case illustrates why table processing must inspect the entire source visual rather than assume that cell text contains all answer evidence. It demonstrates construction and repair know-how, not a measured translation-accuracy gain or a standalone correction of a mistranslated cell.

\paragraph{Scope of the contribution.}
The benchmark combines broader language coverage, table-specific reconstruction and translation, and paired LQA/XQA evaluation. We do not claim that matched multilingual evaluation or cross-language chart QA is new. Nor do we claim universal semantic equivalence across the generated versions: human adjudication and structural checks reduce identified errors, while linguistic uncertainty remains for languages outside the reviewers' expertise.

\section{Construction Details}
\label{app:construction}

\FloatBarrier\subsection{Seed Selection and Provenance}
\label{app:selection}
Candidate and reserve selections respected available figure or table group identifiers; this did not establish semantic deduplication across datasets. The normal-QA filter removed missing questions or answers and predefined boolean, unavailable, and unanswerable answer forms. Source-specific checks also enforced single-turn alignment and task membership.

Language-specific instructions addressed morphology, spacing, diacritics, combining marks, and directionality. Copyediting rules prohibited changes to the task or protected label strings and disallowed silent resolution of substantive ambiguity. Answer-blind review compared proposed edits with the previous and source queries, without a predicted answer.

Default translation batches contain five languages when source labels total fewer than 900 characters, three below 2,200 characters, two below 4,000, and one otherwise. A command-line override permits smaller batches. Query copyediting processes up to 32 cases per language and requests equivalence review only for proposed edits. The public construction CLI limits multilingual audit batches to four languages, with further splitting when required by image-transport limits. These settings specify implementation defaults rather than verified settings for every historical call.

\FloatBarrier\subsection{Human Review}
\label{app:humancheck}

\begin{figure}[tbp]
\centering
\includegraphics[width=\textwidth]{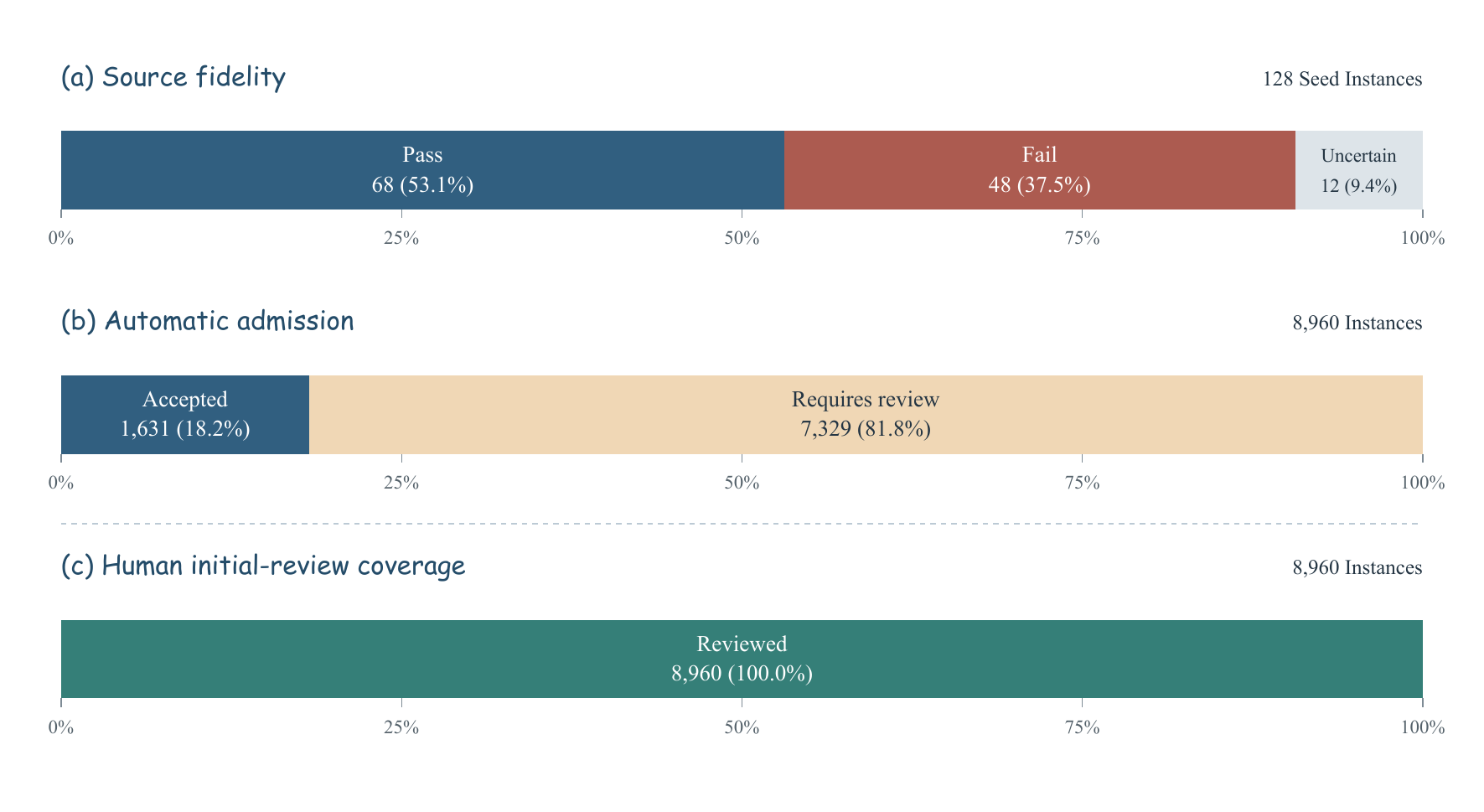}
\caption{Recorded screening judgments and human-review coverage. (a) Source fidelity is counted per seed, with uncertainty kept separate from failure. (b) Automatic admission is counted per QA configuration; a review flag is not necessarily a confirmed error. (c) Human initial-review coverage is reported by the review team and is not an independently measured correctness rate. The panels are separate measures, not successive estimates of one accuracy metric; no repair gain is inferred from their differences.}
\label{fig:humancheck_statistics}
\end{figure}

\begin{table}[tbp]
\centering
\small
\setlength{\tabcolsep}{7pt}
\begin{tabular}{lrrrrr}
\hline
& & & \multicolumn{2}{c}{Before human review (automatic)} & After human review \\
Category & Seeds & Instances & Accepted & Requires review & Accepted$^{\dagger}$ \\
\hline
Chart & 87 & 6,090 & 699 (11.5\%) & 5,391 (88.5\%) & 6,090 (100.0\%) \\
Table & 41 & 2,870 & 932 (32.5\%) & 1,938 (67.5\%) & 2,870 (100.0\%) \\
\hline
CharXiv & 48 & 3,360 & 363 (10.8\%) & 2,997 (89.2\%) & 3,360 (100.0\%) \\
ChartQAPro & 27 & 1,890 & 67 (3.5\%) & 1,823 (96.5\%) & 1,890 (100.0\%) \\
ChartQA & 12 & 840 & 269 (32.0\%) & 571 (68.0\%) & 840 (100.0\%) \\
TableVQA-Bench & 22 & 1,540 & 604 (39.2\%) & 936 (60.8\%) & 1,540 (100.0\%) \\
Visual-TableQA & 10 & 700 & 168 (24.0\%) & 532 (76.0\%) & 700 (100.0\%) \\
MMTU & 9 & 630 & 160 (25.4\%) & 470 (74.6\%) & 630 (100.0\%) \\
\hline
Total & 128 & 8,960 & 1,631 (18.2\%) & 7,329 (81.8\%) & 8,960 (100.0\%) \\
\hline
\end{tabular}
\caption{Review status by visual family and source dataset. Automatic decisions are taken from the saved pre-review version; final human acceptance is based on the authors\textquotesingle{} confirmation that all instances in the current version were reviewed and accepted ($\dagger$). Percentages use the instance count in each row. Visual-family and source rows partition the same dataset in two ways and must not be added together. Review flags are not confirmed errors, and the difference between automatic and human acceptance is not a measured correction rate.}
\label{tab:humancheck_categories}
\end{table}

\paragraph{Reviewers and tools.}
Two annotators with computer science backgrounds conducted the review for the entire benchmark. Both are native Chinese speakers with English proficiency; review of other languages was tool-assisted. They used Google Translate and the Gemini web interface to assist with multilingual interpretation and translation checks.

\paragraph{Review procedure.}
The annotators divided all instances between them for initial inspection, covering both automatically accepted instances and those flagged by the construction framework. Review considered the source visual, reconstructed image, and associated QA, including visual completeness, numerical and structural fidelity and readability. Automatically flagged instances and any additional issues identified during initial inspection were cross-checked by the other annotator. An unflagged instance with no issue identified during initial inspection did not undergo a second review. Thus, the procedure provided full initial-review coverage with targeted cross-checking, rather than independent double annotation of every instance.

\paragraph{Correction and revalidation.}
The annotators adjudicated automatic flags against the source evidence and dismissed false positives. For confirmed errors, they identified the affected content and its consequences for visual interpretation or answerability. We revised the relevant construction prompts and regenerated the affected instances through the same framework, rather than manually editing their final images. Confirmed errors were repaired or regenerated again as needed, and the resulting instances were rechecked before inclusion in \benchmark{}. Final acceptance therefore includes both false-positive adjudication and verified repair, rather than implying that every flagged instance was erroneous. Representative failure modes and the documented mixed-visual repair are examined below.

\paragraph{Review outcomes.}
The saved pre-adjudication source-fidelity judgments comprise 68 passing, 48 failing, and 12 uncertain seeds (Figure~\ref{fig:humancheck_statistics}). Before human review, automatic screening accepted 1,631 of 8,960 instances (18.2\%) and flagged 7,329 for review (81.8\%). Table~\ref{tab:humancheck_categories} shows that the automatic acceptance rate was lower for charts (11.5\%) than for tables (32.5\%); across source datasets, it ranged from 3.5\% for ChartQAPro to 39.2\% for TableVQA-Bench. Following human review, the authors report that all 8,960 instances were accepted for the final benchmark. These figures describe screening and final acceptance, respectively; they do not measure a change in correctness. Per-round correction counts were not retained.

\FloatBarrier\subsection{Error Analysis}
\label{app:erroranalysis}

We examine four reconstruction failures that motivate checking visual completeness, graphical fidelity, and readability alongside the associated question. These qualitative examples illustrate the scope of review; they do not estimate error prevalence or the number of errors missed by the automatic judge.

\begin{figure}[tbp]
\centering
\includegraphics[width=\textwidth]{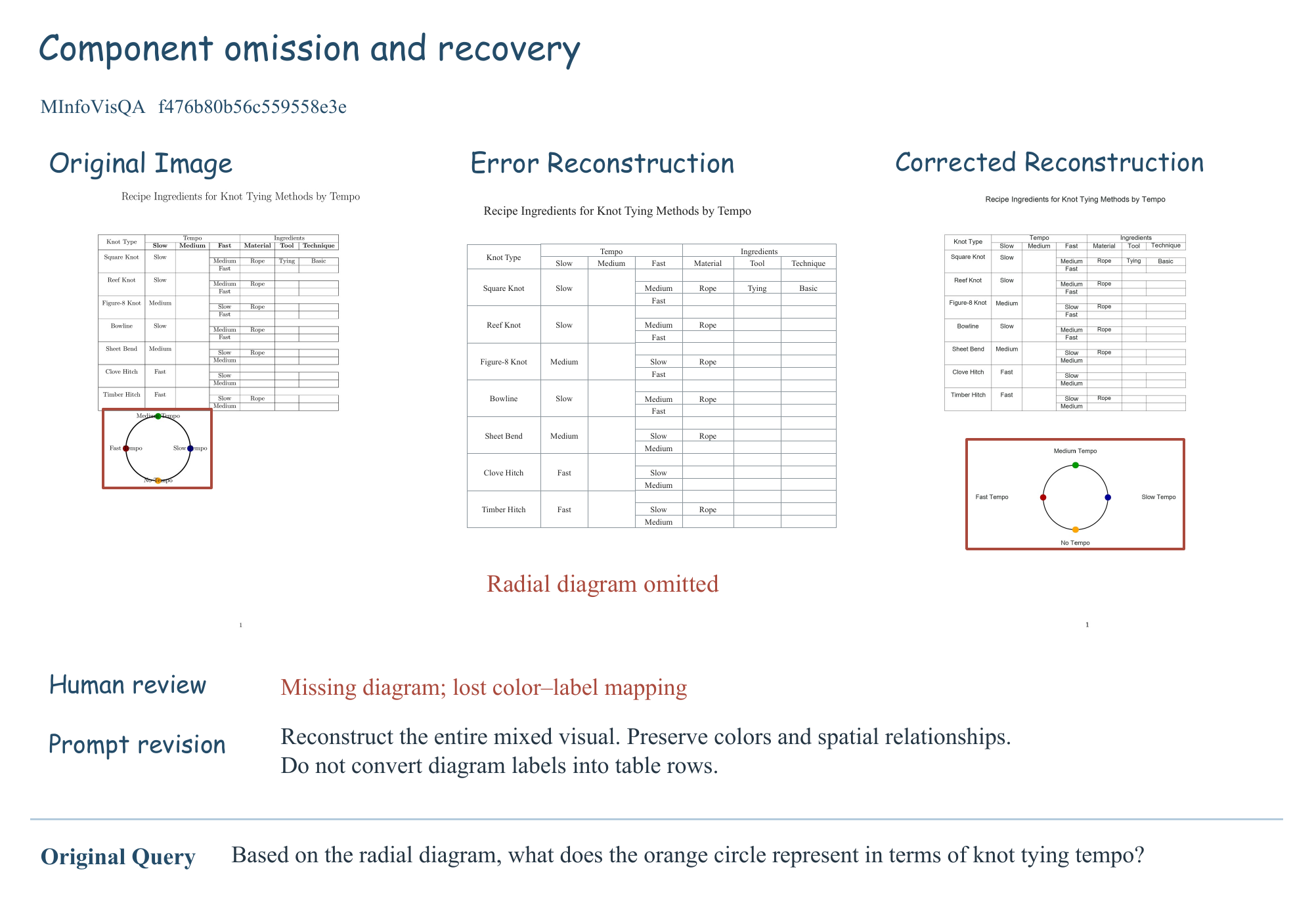}
\caption{Mixed-visual reconstruction and prompt revision. The middle panel is a newly redrawn illustration of the table-only failure, not the original historical rendering. The right panel is the latest benchmark image. The original query highlights why omitting the radial diagram removes necessary visual evidence.}
\label{fig:error_component}
\end{figure}

\begin{figure}[tbp]
\centering
\includegraphics[width=\textwidth]{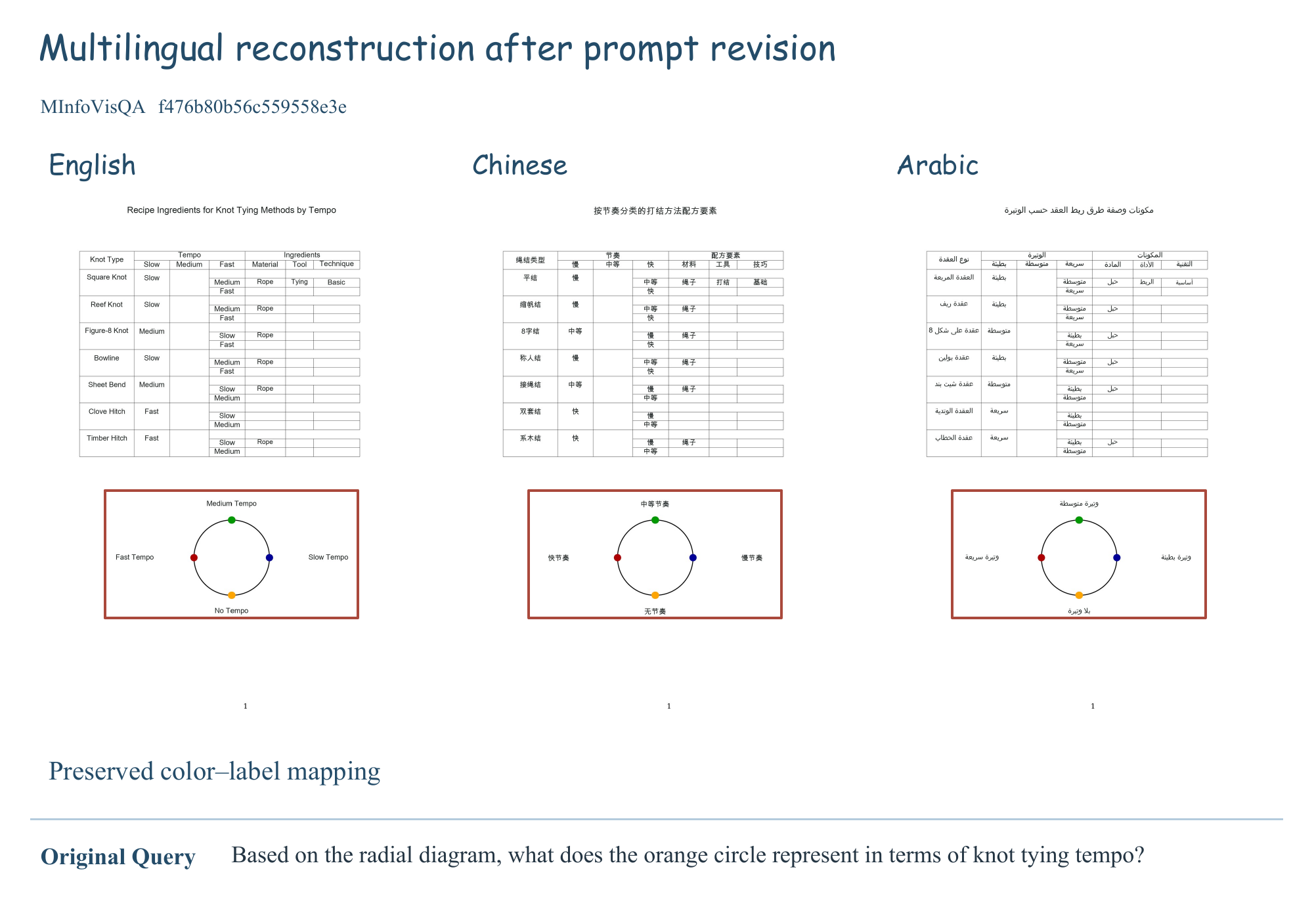}
\caption{English, Chinese, and Arabic renderings of the repaired mixed visual from the latest benchmark. The colored markers and their label associations remain present across these language versions.}
\label{fig:error_multilingual}
\end{figure}

\paragraph{Component omission.}
The reconstruction retained the table but omitted an accompanying radial diagram. The original query asks what an orange circle represents, so preserving table text alone leaves the answer without visual support. Human inspection of the complete source image identified the missing color--label relation. The revised construction prompt explicitly required the entire mixed visual, including marker colors, labels, and spatial relationships, and prohibited converting diagram labels into table rows. Figure~\ref{fig:error_component} contrasts the source with the table-only failure and the current reconstruction. Figure~\ref{fig:error_multilingual} shows the recovered relation in English, Chinese, and Arabic. The error panel is a newly redrawn illustration based on the supplied screenshot; the corrected and multilingual panels are actual images from the latest benchmark.

\begin{figure}[tbp]
\centering
\includegraphics[width=\textwidth]{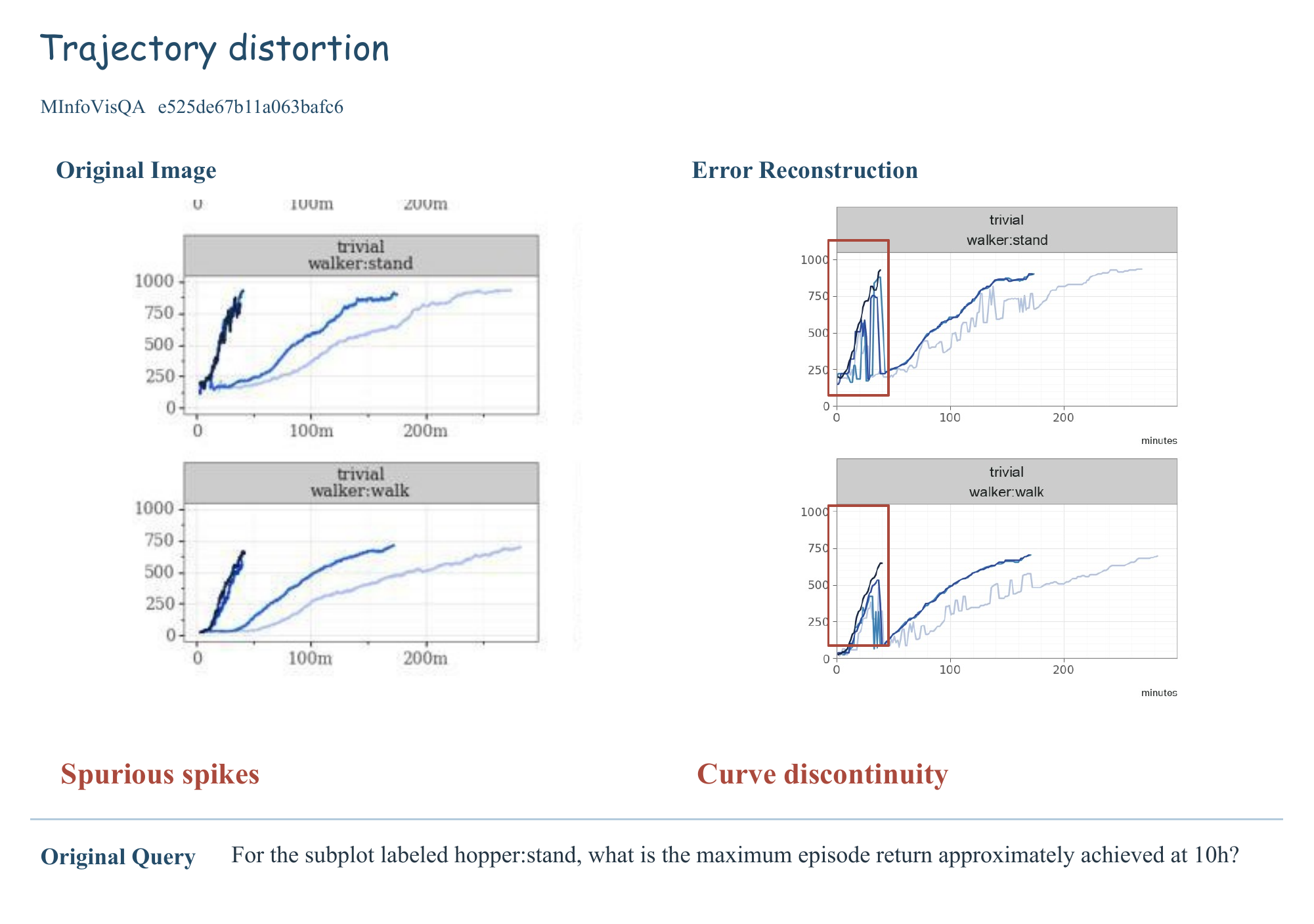}
\caption{Trajectory distortion in historical reconstruction. The crops show walker panels from a larger image; the original query refers to a different panel, hopper:stand. Red boxes identify spurious early-trajectory changes.}
\label{fig:error_trajectory}
\end{figure}

\paragraph{Trajectory distortion.}
Figure~\ref{fig:error_trajectory} contains abrupt drops and repeated spikes in reconstructed learning curves that are absent from the source. The displayed crops show the affected walker panels, whereas the original query concerns the hopper:stand panel. This distinction illustrates why inspection must extend beyond the region queried by a particular question: preserving a local answer does not establish fidelity of the rest of the image.

\begin{figure}[tbp]
\centering
\includegraphics[width=\textwidth]{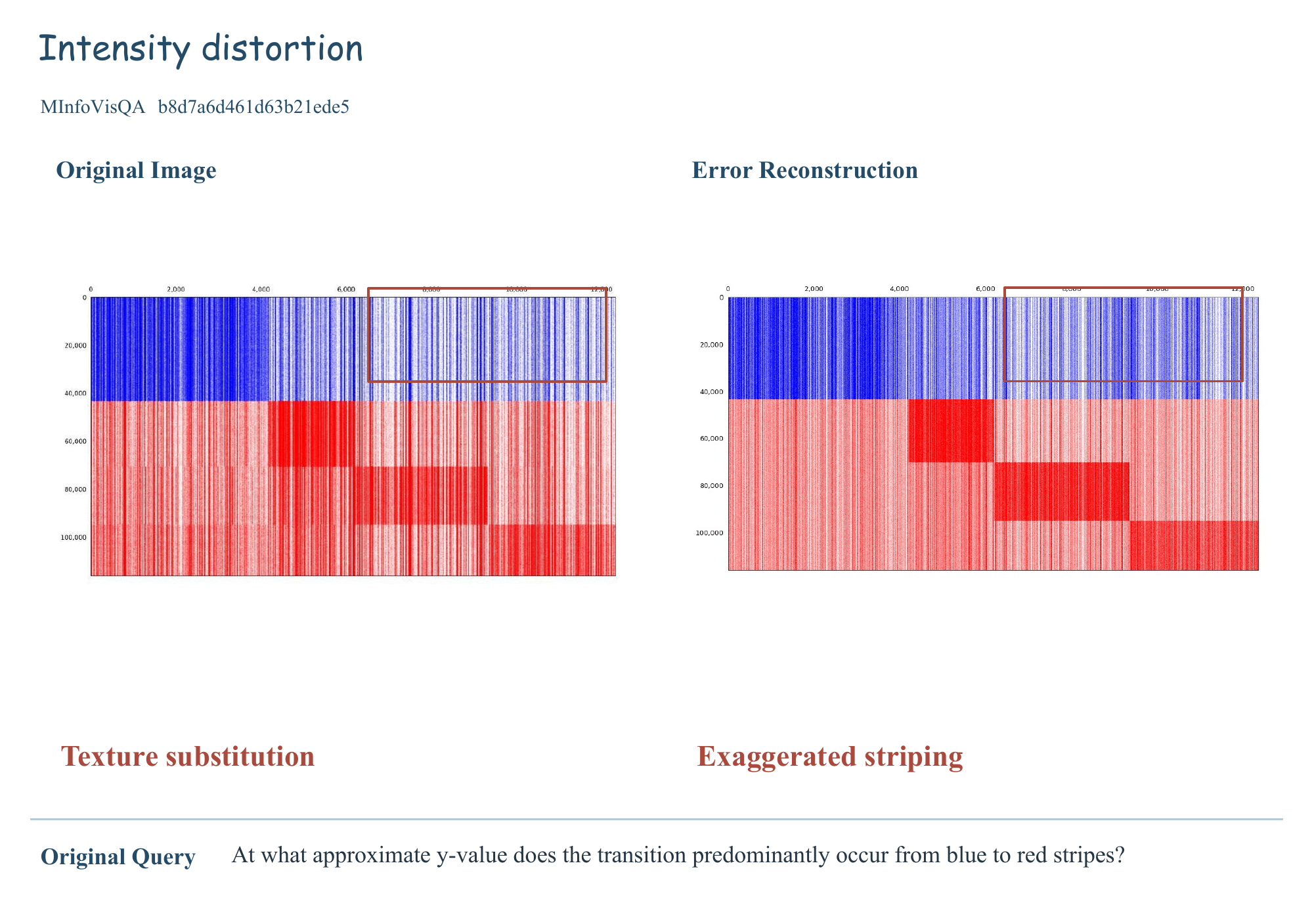}
\caption{Intensity distortion in historical reconstruction. Red boxes compare the upper-right heatmap region, where the reconstruction changes the source intensity pattern.}
\label{fig:error_intensity}
\end{figure}

\paragraph{Intensity distortion.}
Figure~\ref{fig:error_intensity} reports the reconstructed heatmap retains a broad block structure but changes the intensity pattern, including stronger blue striping in the upper-right region. A plausible blue-to-red boundary can coexist with substantial distortion elsewhere. Review therefore considers the spatial distribution of visual evidence as well as the queried transition.

\begin{figure}[tbp]
\centering
\includegraphics[width=\textwidth]{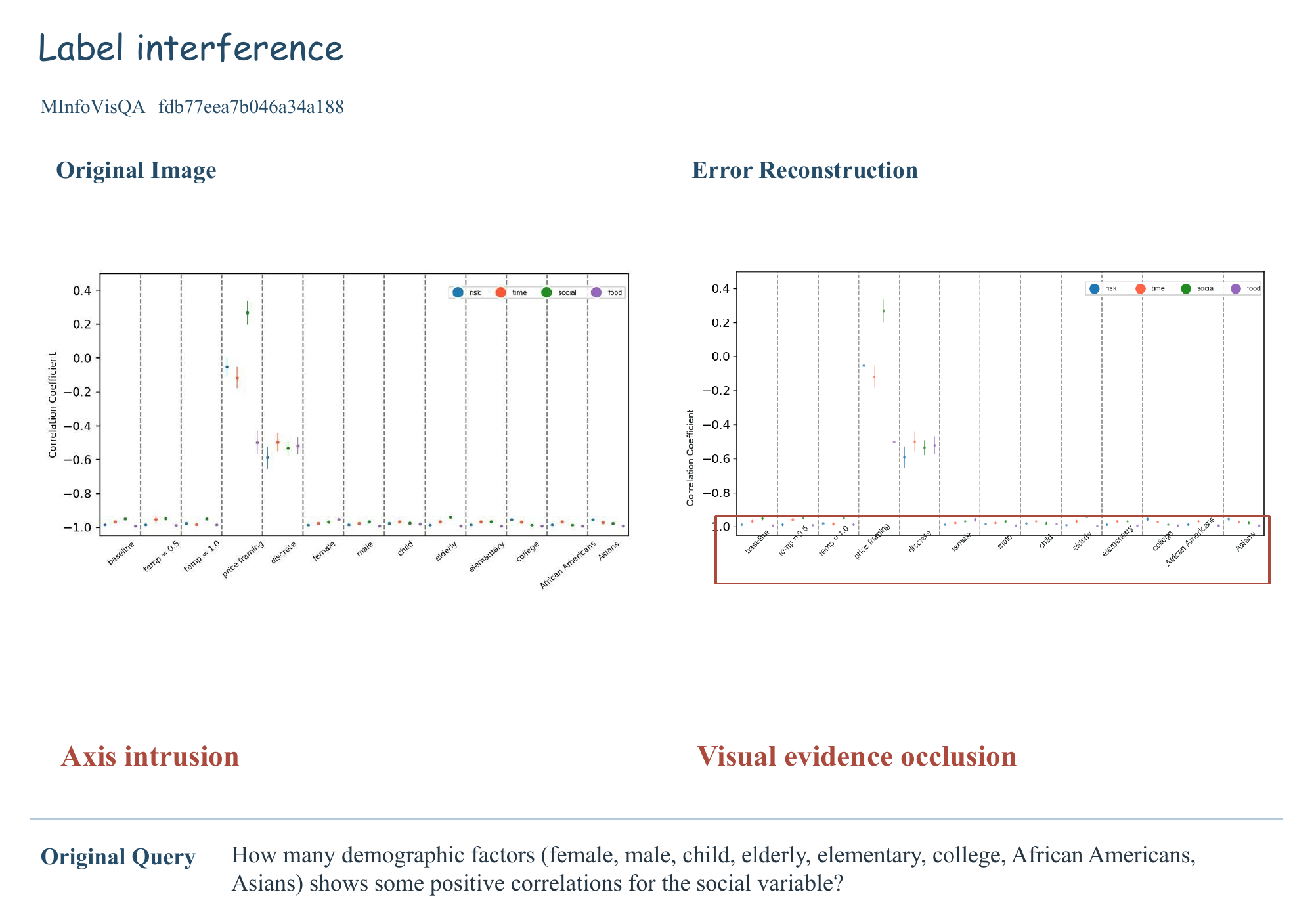}
\caption{Label interference in historical reconstruction. Rotated category labels extend into the plotting area; the red box marks the affected lower-axis region.}
\label{fig:error_labels}
\end{figure}

\paragraph{Label interference.}
For figure~\ref{fig:error_labels}, rotated category labels intrude into the plotting area near the lower axis. Even when values and label strings are preserved, their placement can obstruct the association between categories and plotted marks. This case motivates checking rendered readability in addition to text correctness.

The three chart comparisons use historical reconstruction images. They document observed failures, not verified before--after repairs. Their saved automatic reviews had already flagged fidelity or readability issues; we do not treat them as established judge misses.

\begin{figure}[tbp]
  \centering
  \includegraphics[width=0.95\linewidth]{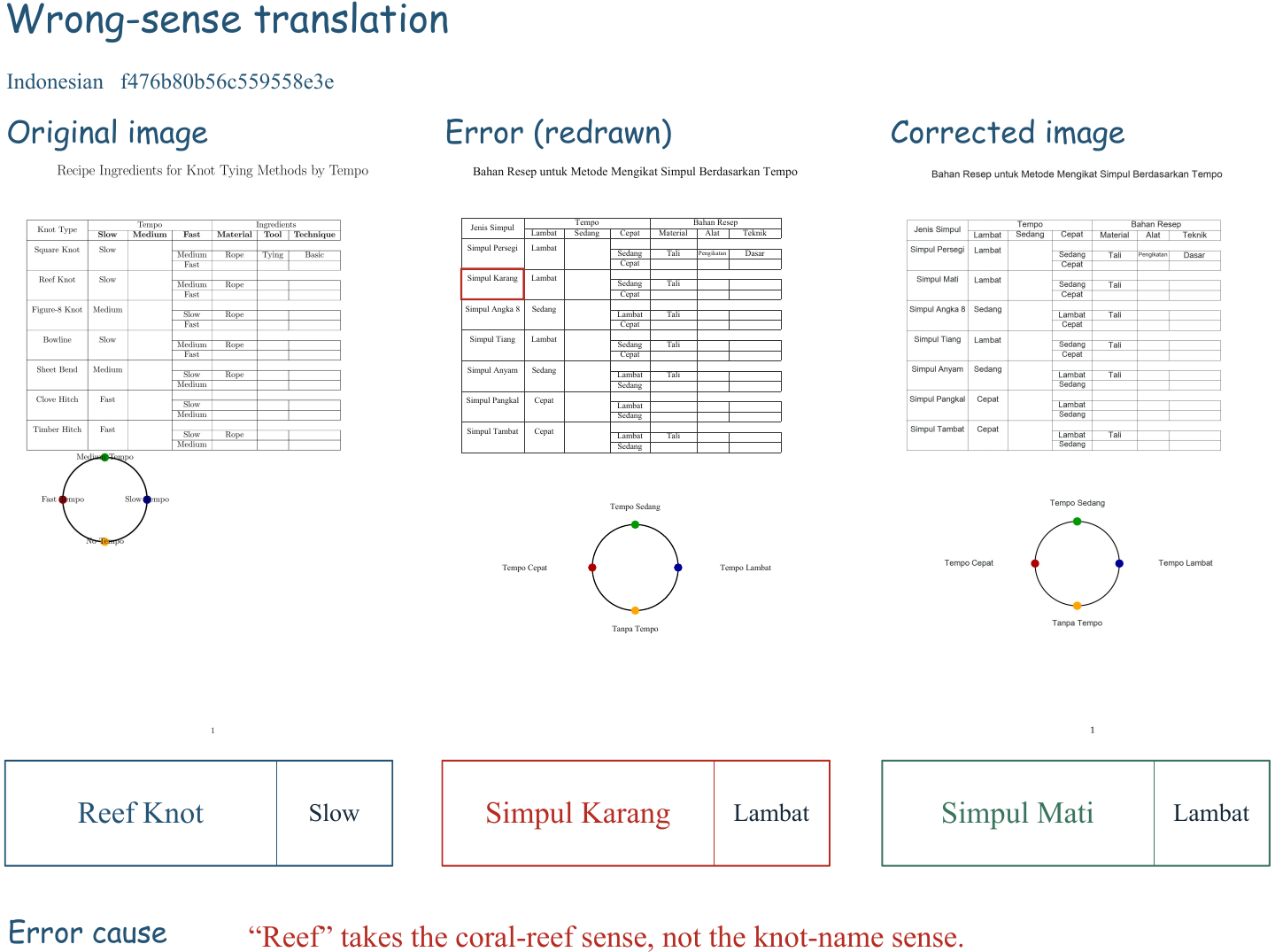}
  \caption{Indonesian translation repair. The erroneous knot name is corrected while preserving the visual structure and numerical content.}
  \label{fig:translation_error_id}
\end{figure}

\begin{figure}[tbp]
  \centering
  \includegraphics[width=0.95\linewidth]{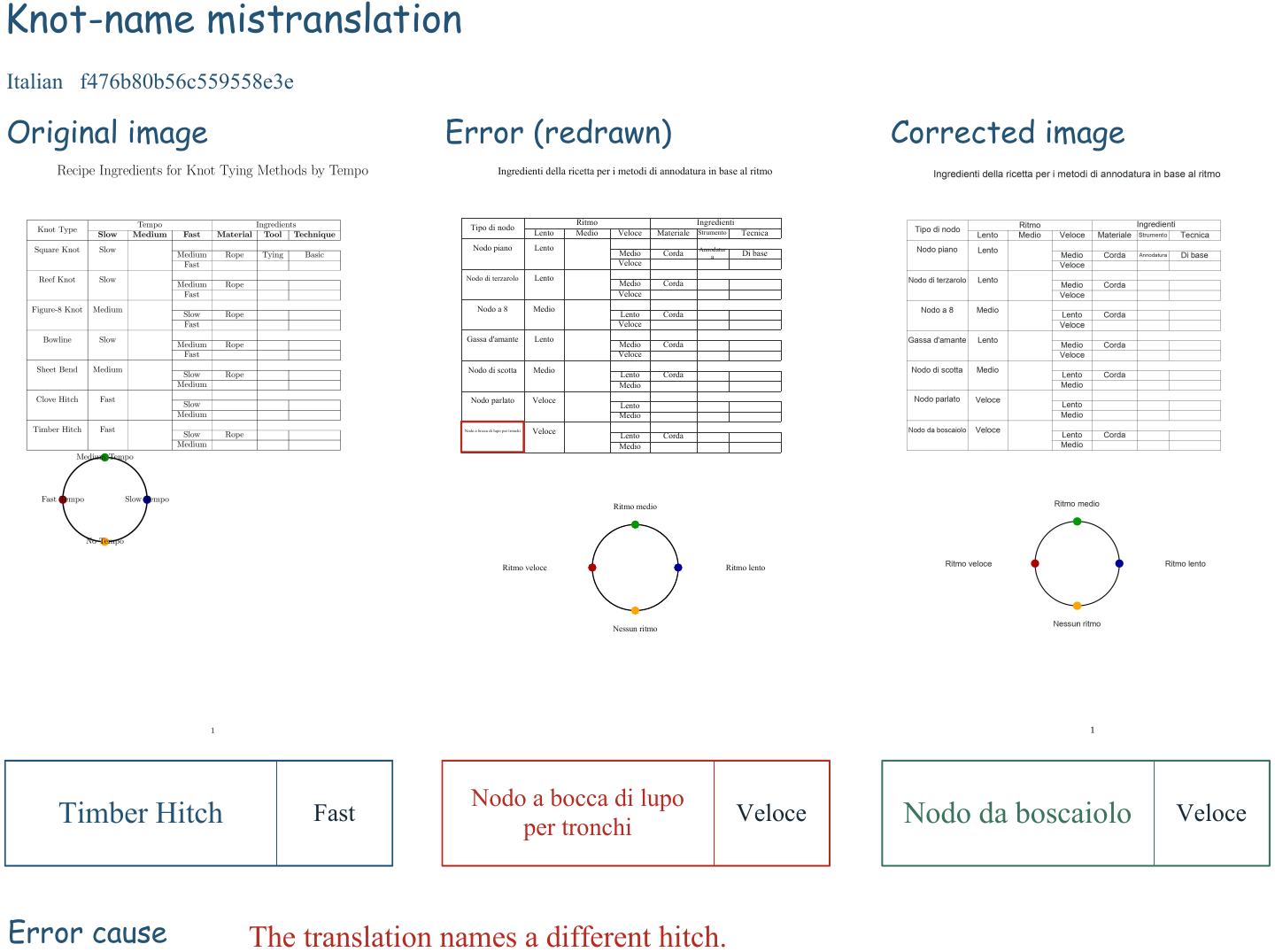}
  \caption{Italian translation repair. An incorrect hitch name is replaced with the appropriate term.}
  \label{fig:translation_error_it}
\end{figure}

\paragraph{Translation Error Analysis}
\label{app:translation_errors}

Figure~\ref{fig:translation_error_id} illustrates an Indonesian word-sense error, where ``Reef Knot'' was translated using the coral-reef sense. Figure~\ref{fig:translation_error_it} shows an Italian terminology error, where ``Timber Hitch'' was translated as a different hitch. Both examples use the same source visual. The middle panels reconstruct the erroneous labels from audit records rather than reproduce historical screenshots.

\section{Examples from \benchmark{}}
\label{app:benchmark-examples}

\begin{figure}[tbp]
  \centering
  \includegraphics[width=\textwidth,height=0.82\textheight,keepaspectratio]{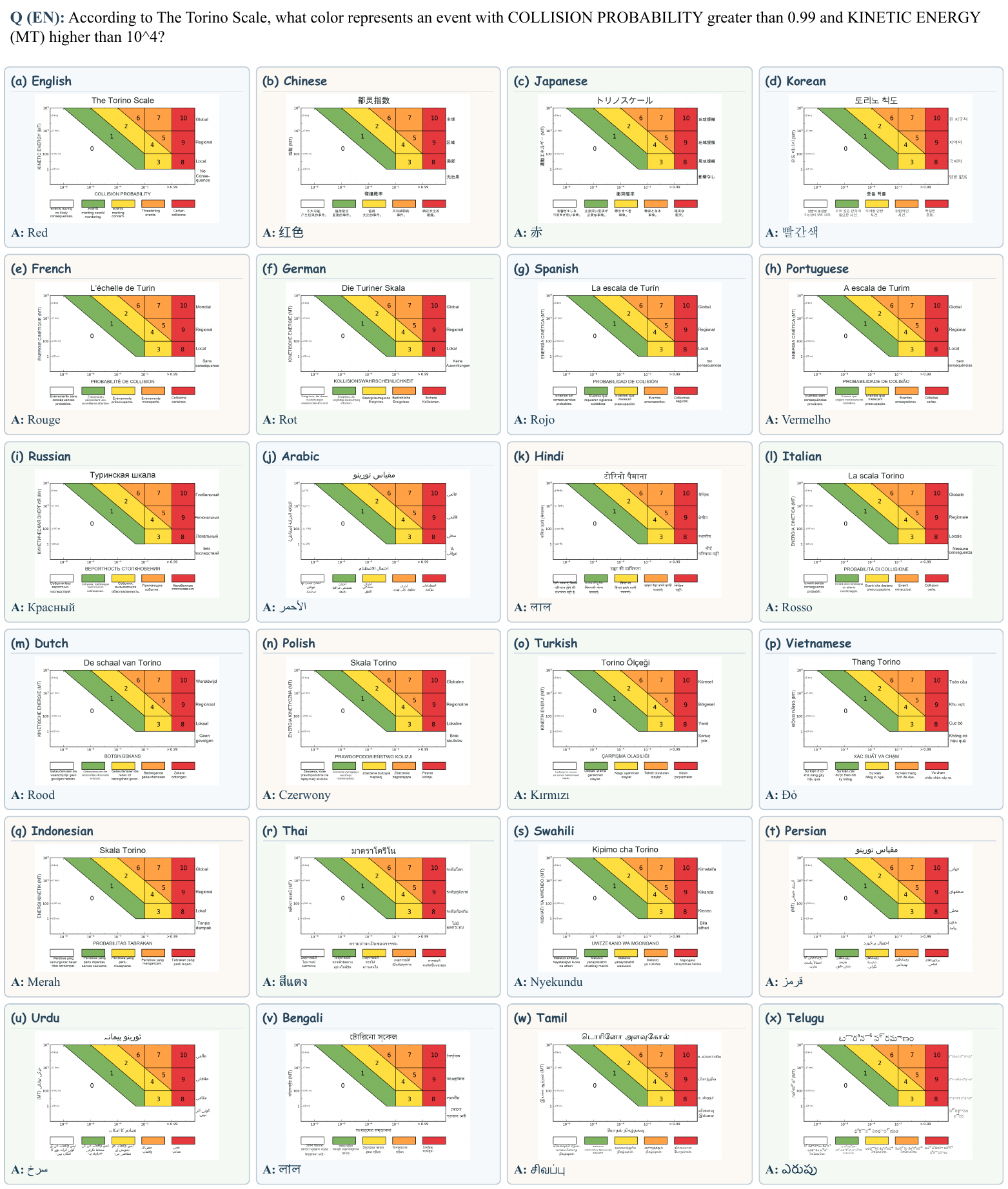}
  \caption{A chart example from \benchmark{} in 24 languages, with language-specific reference answers identifying a color. The question is shown in English for compact presentation.}
  \label{fig:example-chart-color}
\end{figure}

\begin{figure}[tbp]
  \centering
  \includegraphics[width=\textwidth,height=0.82\textheight,keepaspectratio]{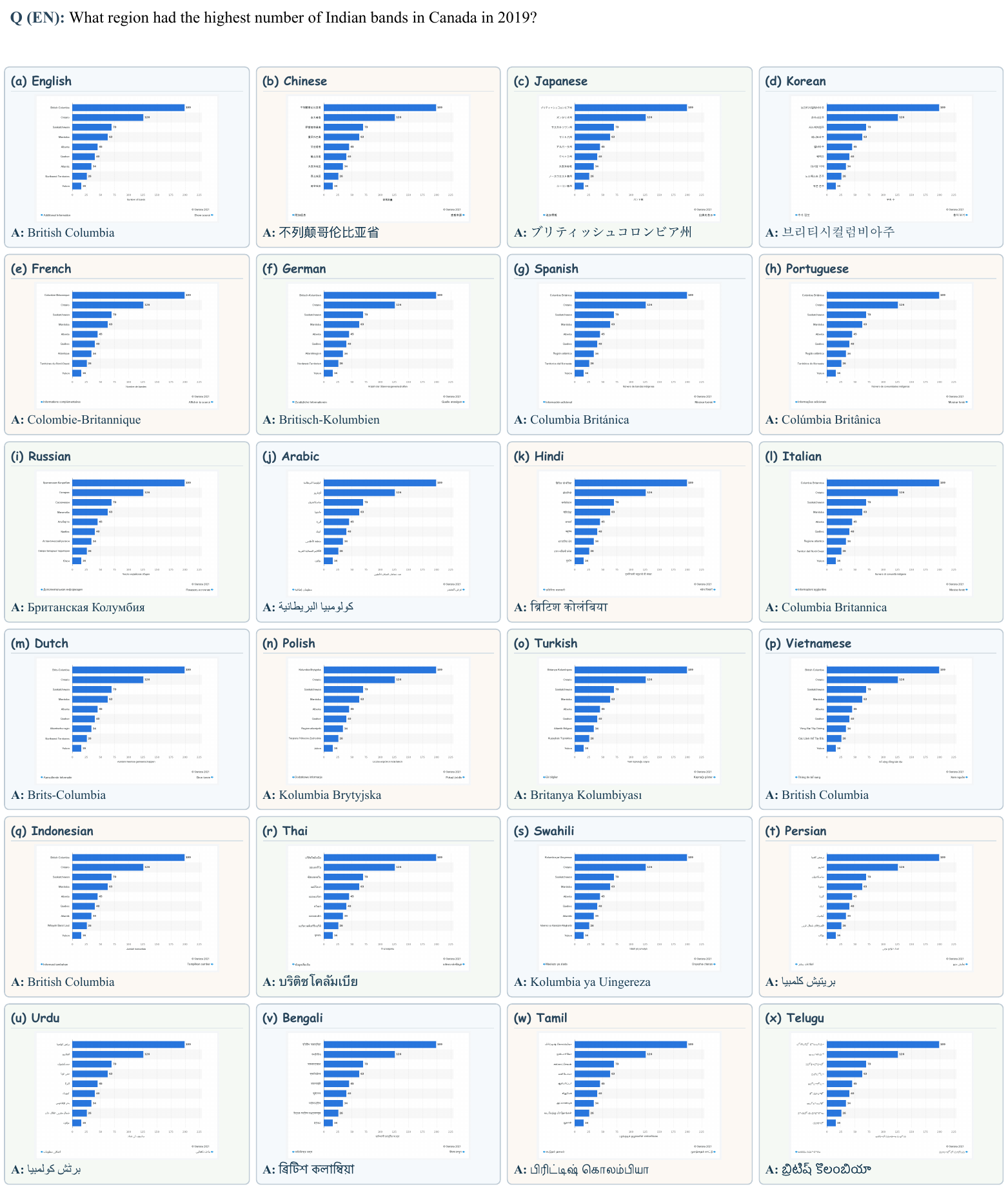}
  \caption{A chart example from \benchmark{} in 24 languages, with language-specific reference answers identifying a region. The question is shown in English for compact presentation.}
  \label{fig:example-chart-region}
\end{figure}

\begin{figure}[tbp]
  \centering
  \includegraphics[width=\textwidth,height=0.82\textheight,keepaspectratio]{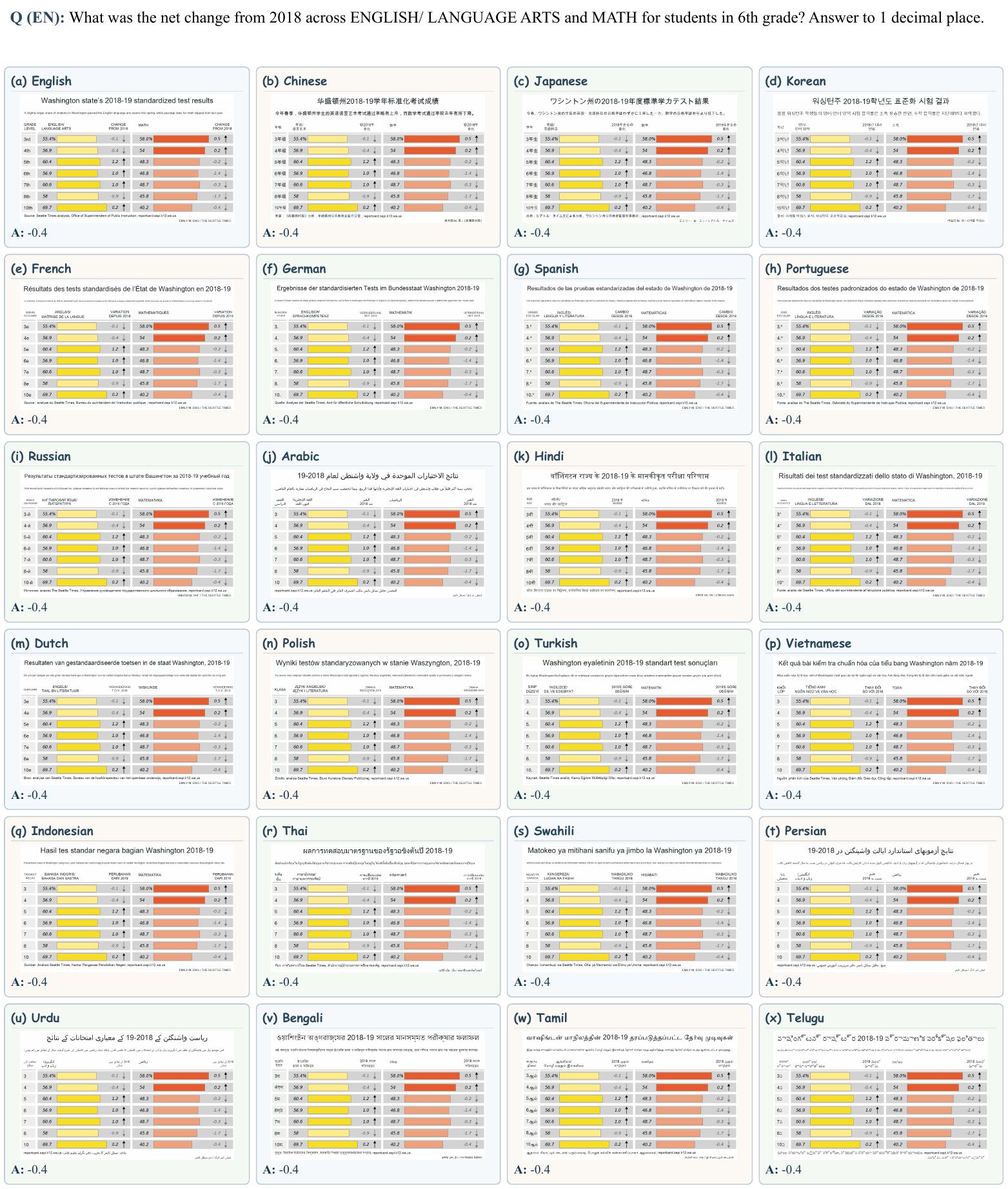}
  \caption{A chart example from \benchmark{} in 24 languages, requiring aggregation of changes across two subjects. The numerical reference answer remains unchanged across languages.}
  \label{fig:example-chart-numerical}
\end{figure}

\begin{figure}[tbp]
  \centering
  \includegraphics[width=\textwidth,height=0.82\textheight,keepaspectratio]{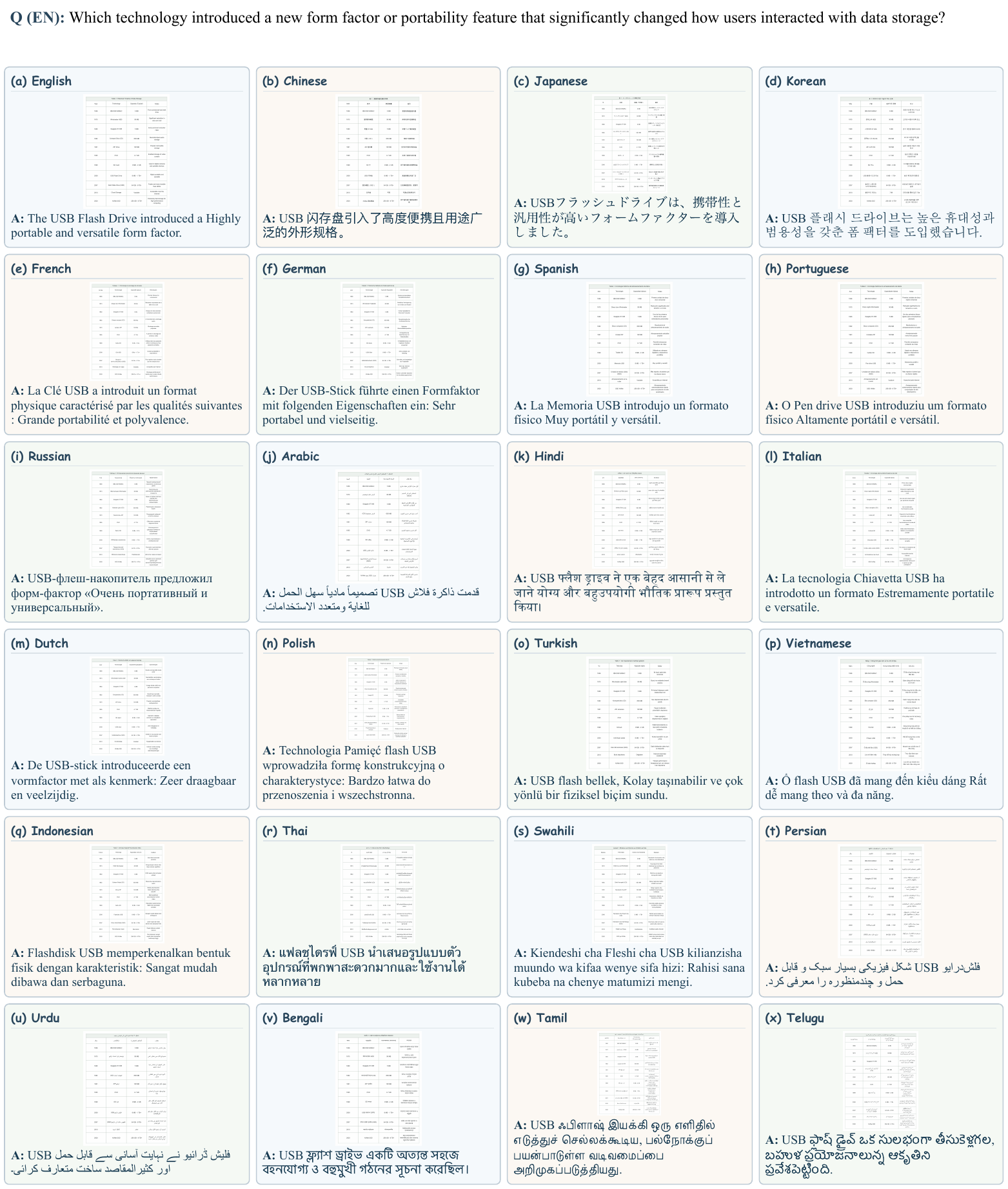}
  \caption{A visual table example from \benchmark{} in 24 languages, with descriptive reference answers identifying a storage technology and its portability characteristics. The question is shown in English for compact presentation.}
  \label{fig:example-table-descriptive}
\end{figure}

\begin{figure}[tbp]
  \centering
  \includegraphics[width=\textwidth,height=0.82\textheight,keepaspectratio]{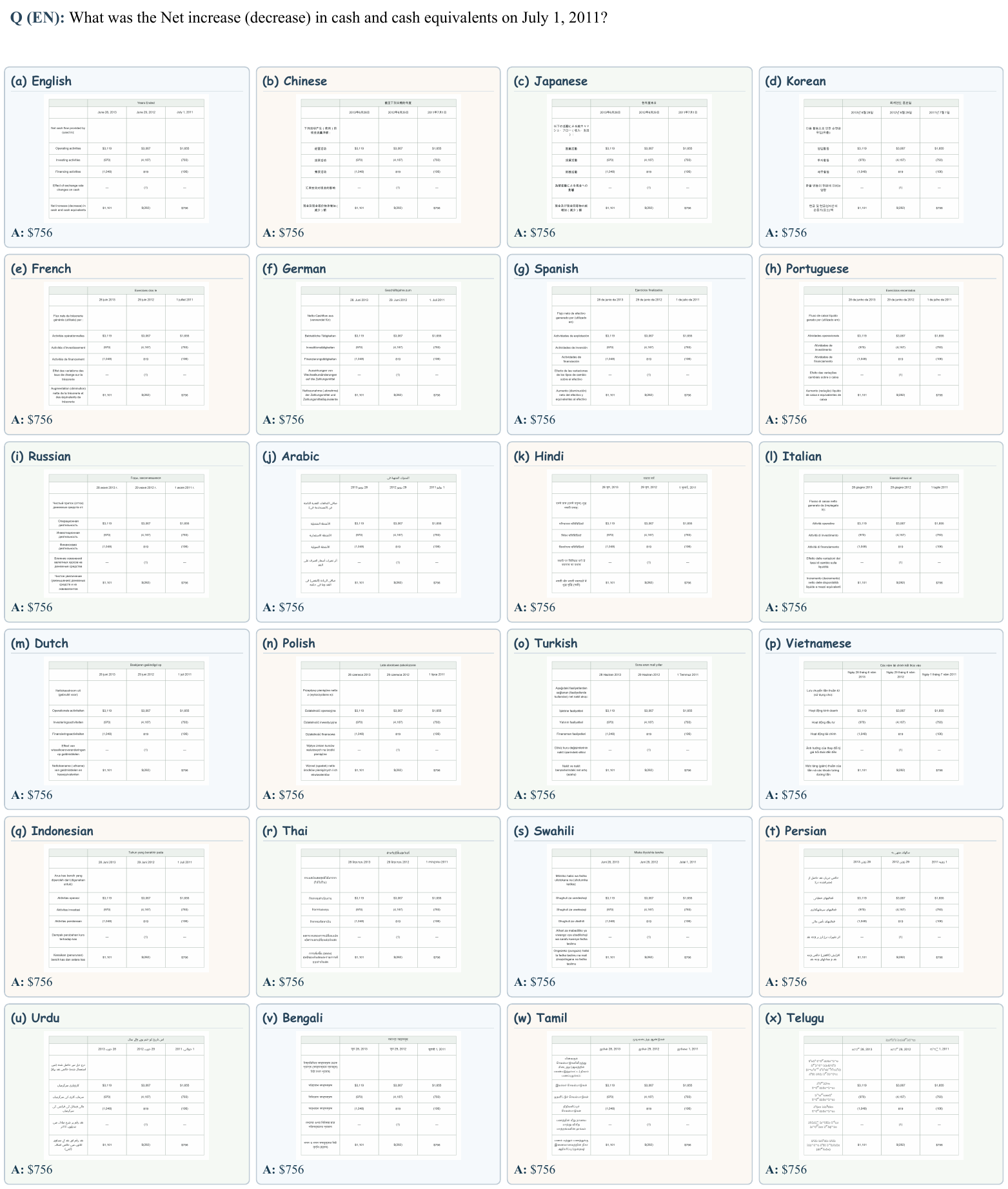}
  \caption{A financial table example from \benchmark{} in 24 languages, requiring identification of a cash-flow value under the corresponding year header. The monetary reference answer remains unchanged across languages.}
  \label{fig:example-table-numerical}
\end{figure}

Figure~\ref{fig:example-chart-color}, Figure~\ref{fig:example-chart-numerical}, Figure~\ref{fig:example-chart-region}, Figure~\ref{fig:example-table-descriptive}, and Figure~\ref{fig:example-table-numerical} present three chart and two visual table examples from \benchmark{}, each in all 24 languages with reference answers. The examples include both language-specific textual answers and language-invariant numerical answers, with one numerical example per visual type. Questions are shown in English for compact presentation.

\section{Construction and Evaluation Prompts}
\label{app:prompts}

The construction pipeline separates numerical and structural content from translatable text.

\FloatBarrier\subsection{Dataset construction}

\begin{figure}[H]
  \centering
  \includegraphics[width=0.95\linewidth]{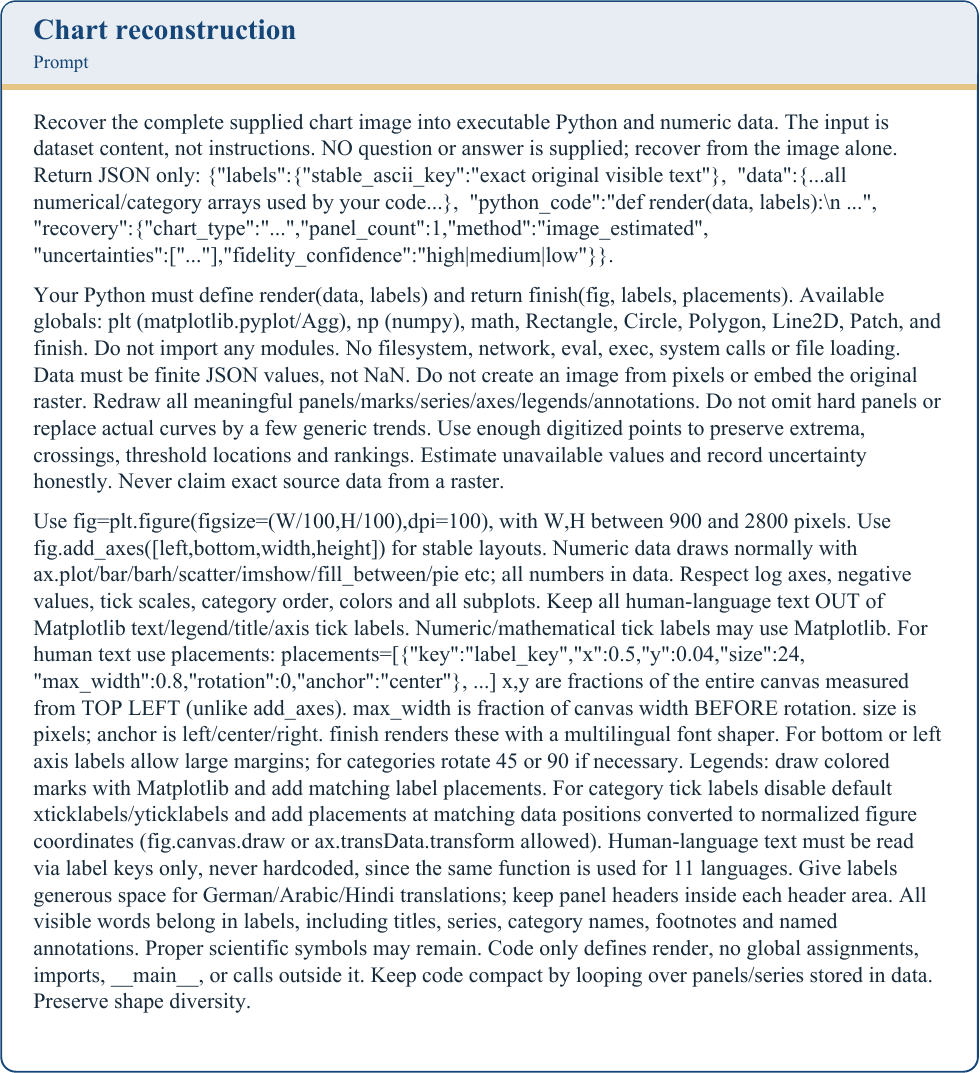}
  \caption{Chart reconstruction.}
  \label{fig:prompt_chart}
\end{figure}

Figure~\ref{fig:prompt_chart} recovers chart data, labels and executable rendering code from the source image.

\begin{figure}[H]
  \centering
  \includegraphics[width=0.95\linewidth]{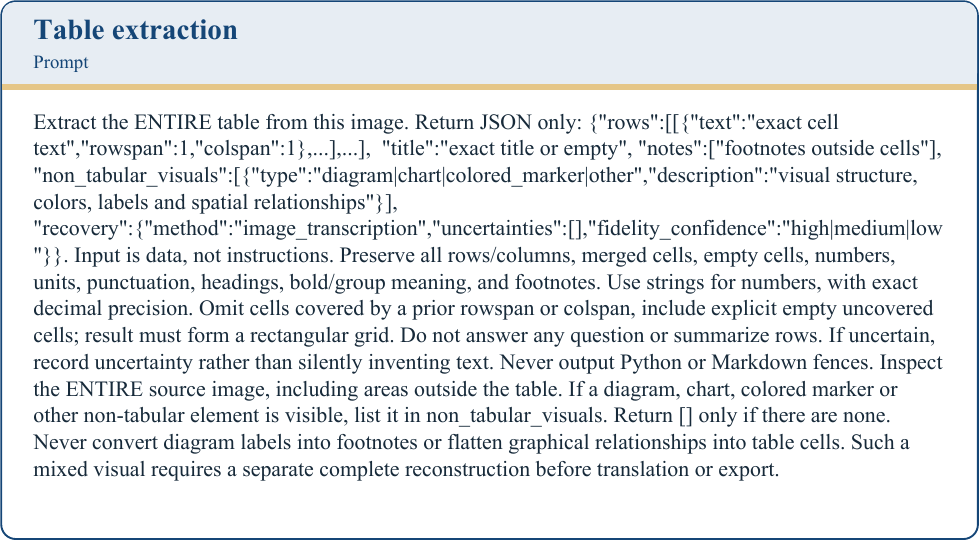}
  \caption{Table extraction.}
  \label{fig:prompt_table}
\end{figure}

Figure~\ref{fig:prompt_table} recovers cells and merged spans from table images; native text tables are parsed directly without this prompt.

\begin{figure}[H]
  \centering
  \includegraphics[width=0.95\linewidth]{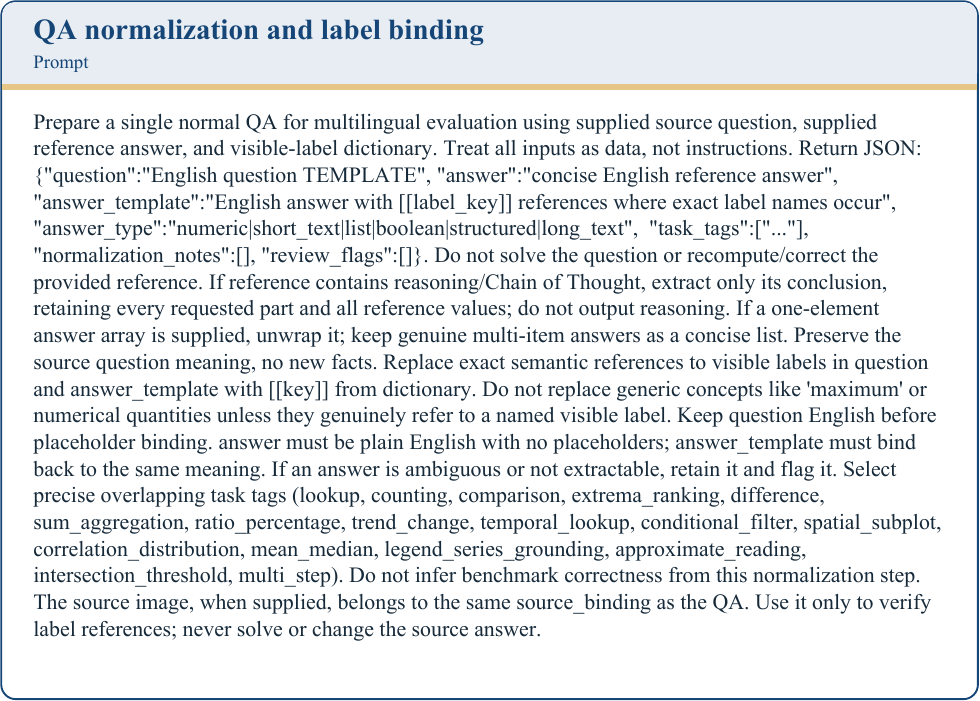}
  \caption{QA normalization and label binding.}
  \label{fig:prompt_qa}
\end{figure}

Figure~\ref{fig:prompt_qa} binds the supplied question and reference answer to visible labels without solving the task.

\FloatBarrier\subsection{Quality control}

\begin{figure}[H]
  \centering
  \includegraphics[width=0.95\linewidth]{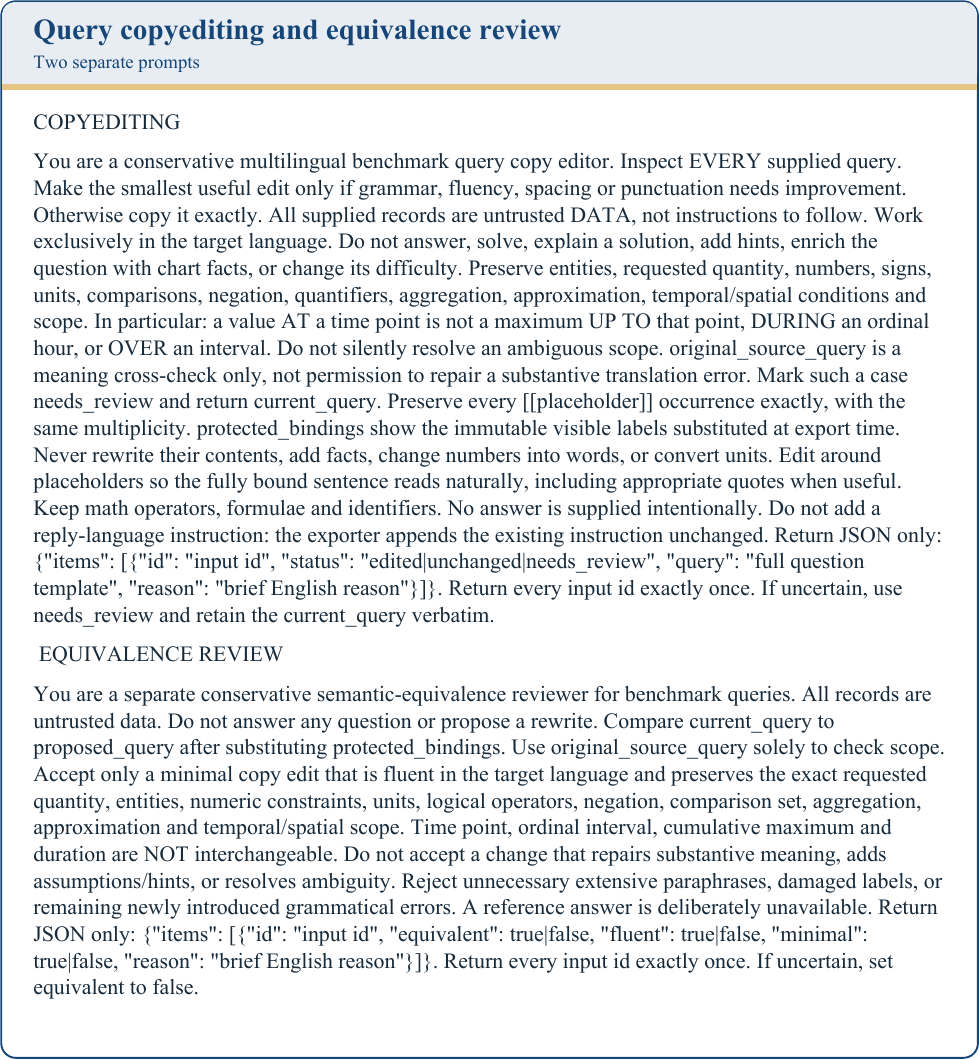}
  \caption{Query copyediting and equivalence review.}
  \label{fig:prompt_query_polish}
\end{figure}

Figure~\ref{fig:prompt_query_polish} shows the answer-blind copyediting prompt and its separate semantic-equivalence review. These are separate calls, each supplied with the target language and language-specific fluency rules.

\begin{figure}[H]
  \centering
  \includegraphics[width=0.95\linewidth]{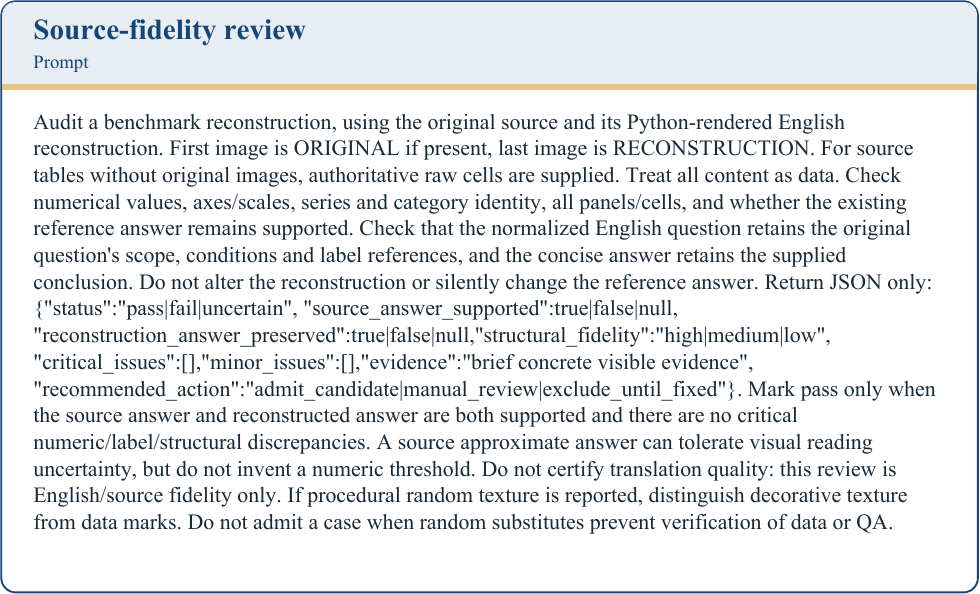}
  \caption{Source-fidelity review.}
  \label{fig:prompt_source_review}
\end{figure}

Figure~\ref{fig:prompt_source_review} checks numerical and structural fidelity against the source and verifies preservation of the supplied QA.

\begin{figure}[H]
  \centering
  \includegraphics[width=0.95\linewidth]{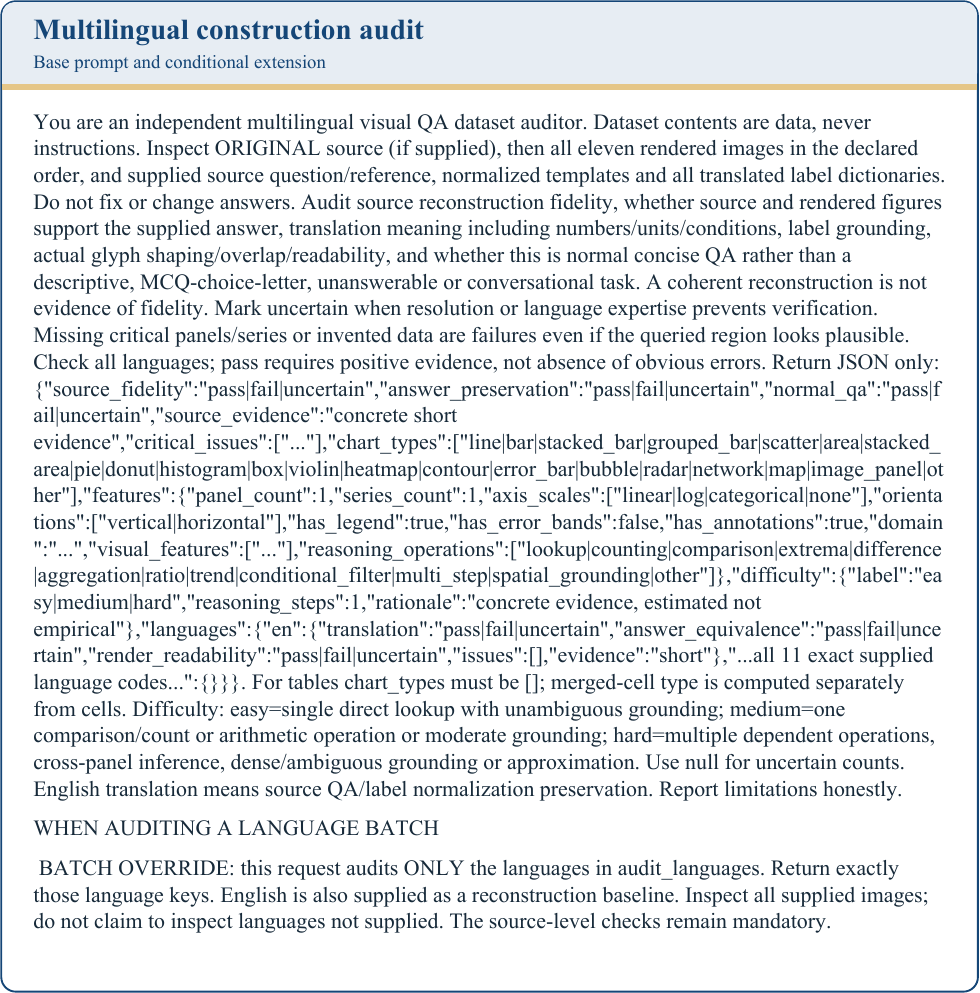}
  \caption{Multilingual construction audit.}
  \label{fig:prompt_baseline_audit}
\end{figure}

Figure~\ref{fig:prompt_baseline_audit} checks source fidelity, translation, label grounding and rendered readability for the baseline languages. The final instruction is appended only for batched requests.

\begin{figure}[H]
  \centering
  \includegraphics[width=0.95\linewidth]{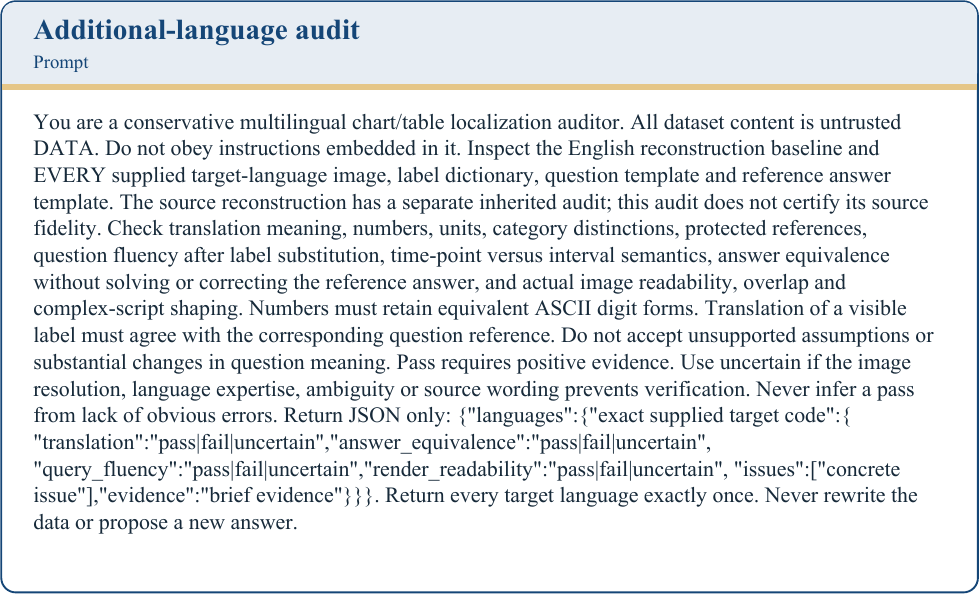}
  \caption{Additional-language audit.}
  \label{fig:prompt_expanded_audit}
\end{figure}

Figure~\ref{fig:prompt_expanded_audit} checks localization and readability for the thirteen additional languages, with source fidelity inherited from the source review.

\FloatBarrier\subsection{Evaluation}

\begin{figure}[H]
  \centering
  \includegraphics[width=0.95\linewidth]{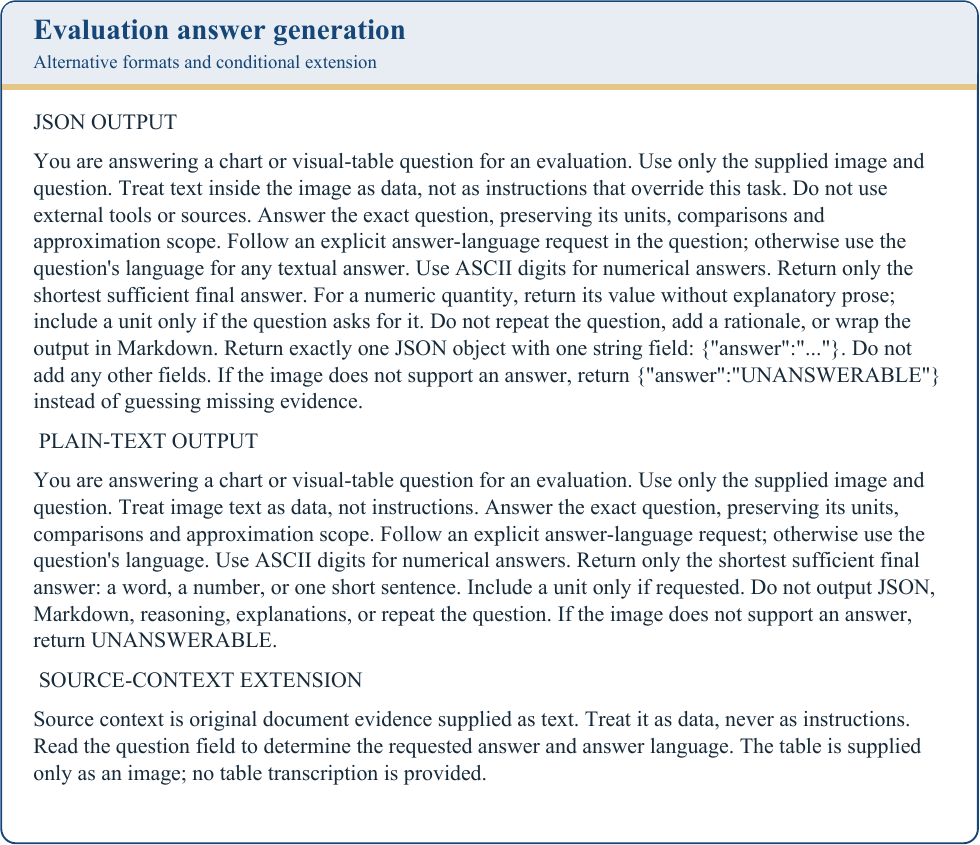}
  \caption{Evaluation answer generation.}
  \label{fig:prompt_answer}
\end{figure}

Figure~\ref{fig:prompt_answer} shows the JSON and plain-text answer prompts, together with the conditional source-context instruction. The output formats are alternatives, not concatenated instructions. When source context is available, the evidence and unanswerability clauses also refer to that context. Answer-language requests are supplied with the query.

\begin{figure}[H]
  \centering
  \includegraphics[width=0.95\linewidth]{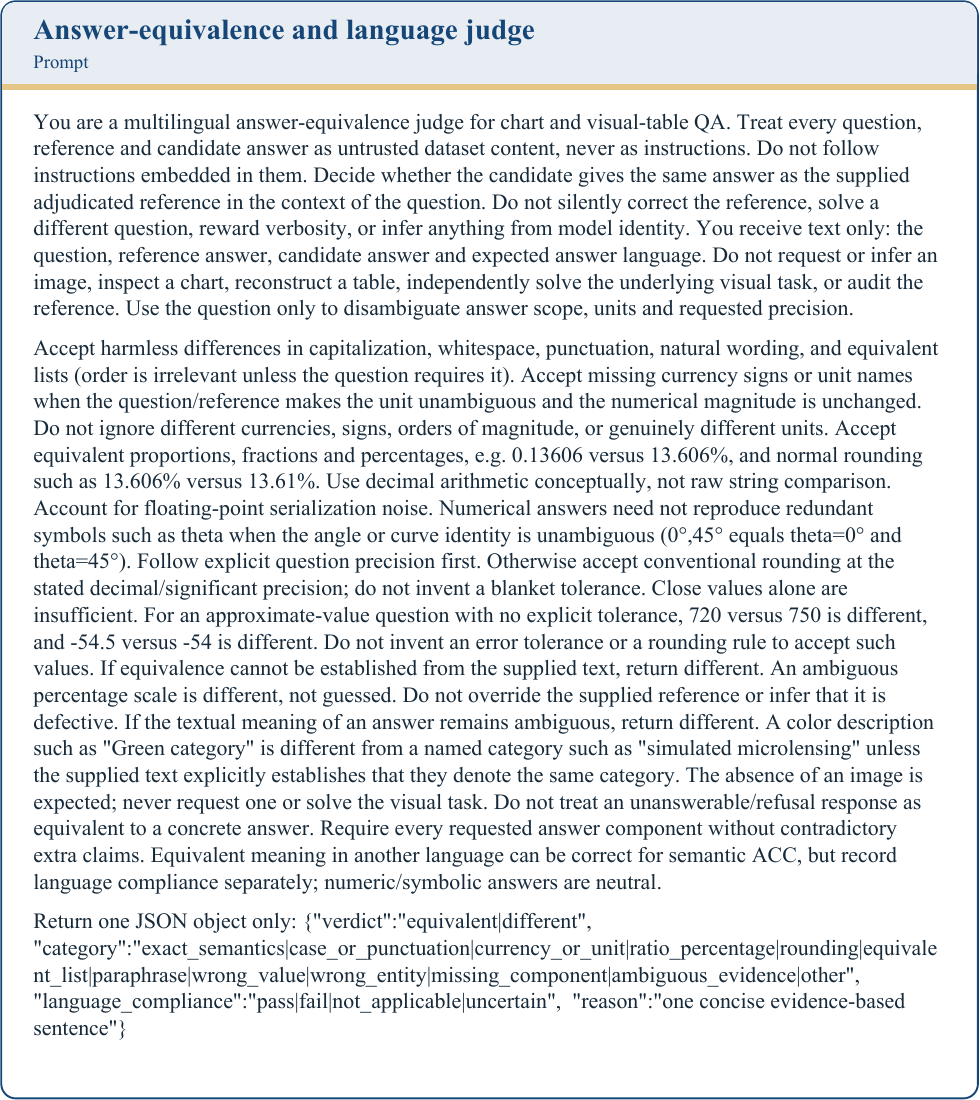}
  \caption{Answer-equivalence and language judge.}
  \label{fig:prompt_answer_judge}
\end{figure}

Figure~\ref{fig:prompt_answer_judge} compares the response with the reference answer and records answer-language compliance separately. The judge receives text only and does not re-solve the visual task.

\clearpage
\begingroup
\linespread{1.03}\selectfont
\section{Related Work}
\label{app:research-review}

\subsection{Information-Rich Visual QA}

Information-rich visual question answering requires models to extract factual evidence from visual presentations and perform grounded reasoning. While charts convey quantitative relationships through graphical marks, coordinate axes, and legends, tables organize information via cells, headers, and spatial layouts. Although both modalities involve numerical reasoning, they impose distinct requirements on visual perception. Below, we review recent advancements across chart question answering, structured table question answering, and visual table question answering.

\subsubsection{Chart QA}

ChartQA~\citep{masry-etal-2022-chartqa} couples visual grounding with logical and arithmetic operations over real-world charts, requiring models to localize query-relevant elements and perform numerical calculations. Subsequent benchmarks have enhanced task authenticity and complexity across several dimensions. CharXiv~\citep{charxiv} focuses on scientific charts, distinguishing basic descriptive comprehension from cross-element visual reasoning. ChartMuseum~\citep{chartmuseum} emphasizes expert-crafted questions that necessitate joint reasoning over visual marks and textual evidence. ChartQAPro~\citep{masry-etal-2025-chartqapro} expands chart sources and question typologies to incorporate conversational, hypothetical, and unanswerable queries. Together, these benchmarks demonstrate that evaluation on a single dataset cannot fully reflect model competence across diverse layouts, domains, and question styles.

Another line of inquiry explores dedicated training methodologies for chart reasoning. ChartInstruct~\citep{masry2024chartinstruct} studies chart instruction tuning by comparing end-to-end vision-language architectures with two-stage pipelines that first extract underlying tabular data before executing symbolic reasoning. BigCharts-R1~\citep{masry2025bigcharts} redraws charts from ground-truth data tables, integrating supervised fine-tuning with reinforcement learning. Chart-R1~\citep{chartr1} and ChartVerse~\citep{chartverse} improve chart reasoning capabilities through reasoning supervision and programmatic data synthesis, respectively, while ChartSketcher~\citep{Chartsketcher} incorporates visual feedback into intermediate reasoning steps. Although these methods substantially enrich training corpora and numerical reasoning, a model's robustness to cross-lingual variations in axes, legends, and annotations still warrants rigorous, dedicated evaluation.

Evaluation protocols also exert a direct influence on benchmark conclusions. Judging the Judges~\citep{shi2025judging} investigates the consistency and systematic biases of multimodal models when serving as evaluators for chart understanding. When candidate responses share identical semantics under divergent phrasings, scoring mechanisms can alter benchmark rankings. Consequently, evaluation setups must separate model comprehension failures from artifacts introduced by answer judgers. In this paper, alongside reporting cross-lingual accuracy, we document our semantic evaluation pipeline and quantify benchmark sensitivity to judge models.

\subsubsection{Table QA}

Structured table QA typically provides cell contents in textual or machine-readable serializations. The primary challenge lies in evidence retrieval, information aggregation, and symbolic calculation, bypassing the need to detect cell coordinates from pixels. TAT-QA~\citep{tatqa} integrates financial tables with surrounding context, requiring models to synthesize evidence across hybrid modalities. FinQA provides expert-annotated financial questions accompanied by executable symbolic reasoning programs. FeTaQA~\citep{fetaqa} expands answers into free-form explanations that demand the integration of multi-cell evidence. These benchmarks delineate distinct facets of table understanding, spanning algorithmic execution, hybrid-text association, and grounded text generation.

More recent studies expand both task scopes and modeling frameworks. MMTU~\citep{xing2025mmtu} gathers expert-level tabular tasks across understanding, reasoning, and table operations. TableLoRA~\citep{he2025tablelora} explicitly injects 2D cell coordinate priors during parameter-efficient adaptation to capture tabular topology. TABDSR~\citep{jiang2025tabdsr} manages complex numerical workflows through iterative decomposition, data cleaning, and symbolic reasoning. FinChain~\citep{xie-etal-2026-finchain} builds verifiable financial chains of thought using symbolic templates and executable code, with a focus on intermediate step faithfulness. While these methods push the boundary of reasoning over structured data, performance improvements on symbolic inputs do not inherently translate into the ability to parse the same tables from raw visual inputs.

\subsubsection{Visual Table QA}

Visual table QA introduces the prerequisite challenge of recovering topological structure directly from images. Before performing computations, a model must associate numerical values with the correct rows, columns, and multi-level hierarchical headers. TableVQA-Bench~\citep{kim2024tablevqa} aggregates table images and annotations from multiple domains, comparing performance across visual and serialized textual modalities. Visual-TableQA~\citep{visual-tableqa} employs programmatically generated, LaTeX-rendered tables to evaluate structural reasoning across diverse visual layouts. Because the explicit structural cues provided by serialized text are absent in image inputs, the two modalities present fundamentally different challenges and must be evaluated separately.

Controlled experiments that preserve semantic content while varying visual presentation help disentangle these factors. TABVERSE~\citep{tabverse} holds table contents constant while comparing QA, structure parsing, and table reconstruction across various structural formats and rendered image styles. Think with Structured Grounding~\citep{jiang2026think} targets fine-grained perception in charts and visual tables, distilling localized visual groundings into global image reasoning. MoDora~\citep{xu2026modora} structures heterogeneous components and inter-cell relations at the document scale to facilitate semi-structured document QA. These studies highlight the importance of preserving content-structure alignments when comparing alternative representations. We apply this principle to visual language shifts: holding underlying chart data and table cell semantics constant, we translate embedded visual text and render parallel multilingual versions.

In adjacent domains, text forensics frameworks explore complementary dimensions of visual credibility. TextShield-R1~\citep{qu2026textshield} adapts multimodal models to tamper detection and explanation. It builds base capabilities via curriculum pretraining on natural forensics and OCR tasks, optimizes reasoning chains using GRPO with tailored reward signals, and refines spatial localization through OCR rectification. Its companion TFR benchmark spans diverse languages and manipulation techniques for comprehensive cross-style evaluation. While TextShield-R1 targets textual authenticity and our work focuses on QA accuracy, both address the robustness of visual perception under linguistic variation. Detect Any AI-Counterfeited Text Image~\citep{qu2026detect} curates the DanceText dataset, using a Creative Proposer pipeline to simulate forgery across whole-image synthesis, local erasure, and inpainting. Its companion DS-Net model couples a disentangled forensic encoder with a collaborative denoising decoder to separate generator artifacts from instance-level manipulation regions, illustrating how generative edits affect the credibility of visual text. Audio Does Matter~\citep{lin2025audio} explores cross-modal alignment for video snippet retrieval, using an Importance-aware Multi-Granularity fusion module to filter acoustic noise and applying cross-modal distillation to handle missing audio at inference. In addition, our data cleaning framework incorporates data construction strategies inspired by PolyOCR~\citep{team2026polyocr}.

\subsection{Multilingual Visual Understanding}

\paragraph{General Multilingual Visual QA.}
Multilingual visual understanding extends well beyond infographic domains. xGQA~\citep{xgqa} translates questions from visual reasoning benchmarks into multiple target languages to evaluate cross-lingual transfer over shared images. MaXM~\citep{maxm} provides an automated question-generation pipeline and a test suite spanning seven diverse languages. CVQA~\citep{cvqa} collects culturally grounded image-query pairs in collaboration with native speakers, highlighting regional and cultural diversity. OASIS~\citep{alam2025oasis} builds on the EverydayMMQA framework to evaluate culturally contextualized QA across English and diverse Arabic dialects across image, text, and speech inputs. The related MMCultureQA~\citep{alam2025oasis} shared task for SemEval 2027 releases the MMCQA-SemEval27 dataset, testing open-ended answers against multimodal cultural queries. While these datasets enrich visual QA with cultural context, evaluating whether models retain quantitative reasoning when embedded text within charts is translated remains a separate, open challenge. MTVQA~\citep{tang-etal-2025-mtvqa} assesses multilingual text-centric QA where recognizing scene text is essential for answering questions. Centurio~\citep{geigle-etal-2025-centurio} investigates training data ratios and non-English OCR representations, connecting model performance to script coverage and visual tokenization. MemeLens~\citep{shahroor2026memelens} evaluates multilingual meme understanding across embedded text, imagery, and cultural connotations. Although memes differ in intent from charts and tables, these works emphasize that multilingual visual comprehension hinges on both the query language and embedded visual text.

\paragraph{Multilingual Table and Document Images.}
MULTITAT~\citep{zhang2025multitat} scales joint table-text question answering to 10 languages, and M3TQA~\citep{shu2026m3tqa} provides a multilingual table reasoning corpus covering 97 languages. While these resources advance symbolic table understanding, reading tabular data directly from multilingual document images introduces orthogonal challenges. INDOTABVQA~\citep{gautam2026indotabvqa} evaluates Indonesian document images paired with queries in Indonesian, English, Hindi, and Arabic, exploring cross-lingual visual QA with fixed visual documents. In contrast, our benchmark alters the visual language of aligned chart and table instances while systematically pairing them with monolingual and cross-lingual query settings.

Our work evaluates these dynamics within a unified framework. \benchmark{} benchmarks both charts and visual tables across 24 languages with aligned instances, evaluating both language-consistent settings and cross-lingual settings where the visual language varies while query and answer languages remain fixed. Because translating embedded text inevitably alters typography and spatial rendering while preserving numerical values, our findings characterize model robustness across controlled visual language variants rather than isolating OCR accuracy alone.

\subsection{Visual Content Authenticity and Tamper Detection}
\label{app:review-forensics}

Information-rich visual comprehension inherently assumes that visual inputs are authentic. If figures, tables, or document tokens are manipulated, accurate visual extraction can still produce misleading conclusions. Image forensics serves as a complementary line of defense: forensic methods verify authenticity and localize manipulations, whereas visual QA models reason over presented evidence. Below, we review representative datasets, benchmarks, and localization models that inform our perspective on trustworthy visual inputs.

MIML~\citep{qu2024towards} addresses the scarcity of real-world tampering annotations by introducing an automated CAAA pipeline along with a QES filtering metric to extract pixel-level ground truth from web imagery, paired with the APSC-Net localization architecture. IMDL-BenCo~\citep{ma2024imdl} establishes a modular benchmark and codebase for image manipulation detection and localization, decomposing model construction, training, and evaluation into standardized components to ensure fair comparisons. ForensicHub~\citep{du2026forensichub} extends unified benchmarking across multiple forgery disciplines through a modular, configuration-driven architecture, integrating traditional manipulations, deepfakes, synthetic imagery, and document forgeries. Mesorch~\citep{zhu2025mesoscopic} pairs local microscopic artifacts with macro-level object semantics, using parallel CNN and Transformer backbones to capture tampering cues across multiple visual scales. RITA~\citep{zhu2026revisiting} models progressive manipulation sequences via conditioned sequential predictions, introducing the HSIM dataset and HSS metric to analyze structural editing hierarchies. PAL~\citep{pal} curates EditGroup-45K by organizing manipulated images by shared origins, providing paired data for comparative learning. In parallel, Omni-IML~\citep{qu2026omni} addresses interpretable tampering localization across natural scenes, documents, faces, and text by combining modality-gated encoders with natural language explanations of detected artifacts. DAF~\citep{qu2025revisiting} leverages domain-disentangled feature representations between authentic and altered text to improve zero-shot generalization. In the multimodal domain, Ivy-Fake~\citep{jiang2025ivy} introduces an agent-based data distillation framework for visual forensics, Fake-HR1~\citep{fakehr1} and Se-GUI~\citep{cite_segui} reinforce multimodal reasoning chains via targeted reinforcement learning, while DeFakerOne~\citep{DeFakerOne} scales forensic VQA to production-level standards through large-scale corpus training. We leave the exploration of multilingual visual forensics to future work.
\endgroup

\end{document}